\documentclass{veriphy}
\usepackage{veriphy}

\DeclareRobustCommand{\VeriPhy}{{\fontfamily{lmtt}\selectfont VeriPhy}}
\title{\VeriPhy: Agentic Physical Reasoning for World Model Evaluation and Refinement}
\author[1\S]{Wenzhuo Xu}
\author[2]{Yuchen Zhu}
\author[4]{Chongjian Ge}
\author[3]{Xuan Shen}
\author[4]{Jing Shi}
\author[4]{Jason Kuen}
\author[2]{Yongxin Chen}
\author[2]{Molei Tao}
\author[1]{Christopher McComb}
\author[1]{Noelia Grande Guti\'errez}
\author[4\S\dagger]{Jiuxiang Gu}

\affiliation[1]{Carnegie Mellon University}
\affiliation[2]{Georgia Institute of Technology}
\affiliation[3]{Northeastern University}
\affiliation[4]{Adobe Research}

\contribution[\S]{Core Contribution}
\contribution[\dagger]{Project Lead}

\date{September 2026}
\hypersetup{
  pdftitle={\VeriPhy: Agentic Physical Reasoning for World Model Evaluation and Refinement},
  pdfauthor={Wenzhuo Xu, Yuchen Zhu, Chongjian Ge, Jing Shi, Jason Kuen, Jiuxiang Gu},
  pdfsubject={Agentic physical reasoning for video evaluation and refinement}
}

\abstract{
Visual fluency in generated video does not imply physical reliability, and a scalar quality
score alone is incapable of indicating the obligation a clip violates or the moment it fails. We present
\VeriPhy{}, an auditable physical-verification system in which a text-only planner compiles
the prompt into typed physical obligations and a statically validated execution plan before
any frame is observed. During execution, observations gate and scope only declared calls to
frozen low-level experts (e.g., segmentation and tracking, counting, eleven typed physical
measurements over the resulting tracks, depth, OCR, and audio-event detection). Each action
returns a provenance-carrying evidence record whose payload, when usable, is either a typed
measurement or an explicitly tagged learned state. Typed resolvers and fixed composition map usable records to
a three-valued state (\supported{},
\contradicted{}, or \unknownv{}, surfaced as \emph{plausible},
\emph{implausible}, or \abstain{}) with full provenance, so that every verdict is
traceable to the evidence that produced it. We anchor evaluation in a
1{,}500-clip corpus of human-annotated flaw records that localize real generation
failures in prompt reference, space, and time. On a $149$-clip core carrying $304$
such records, \VeriPhy{} accounts for $228$, against $164$ for a published
question-decomposition evaluator given the same clips and the same claims. Recall alone does
not separate it from prompting the same backbone monolithically, which reaches $222$; what
separates them is that each decision retains its evidence record and provenance, making the
traces auditable one verdict at a time and usable as the interface through which a critic
verdict could be written back into generation. The core is a development set, so these
figures characterize the system rather than its generalization.
}

\veriphydata[Project Page]{\href{https://veriphy-ai.github.io}{https://veriphy-ai.github.io}}

\newcommand{\code}[1]{\texttt{\small #1}}
\newcommand{\supported}{\textsf{supported}}
\newcommand{\contradicted}{\textsf{contradicted}}
\newcommand{\unknownv}{\textsf{unknown}}
\newcommand{\abstain}{\textsf{abstain}}

\graphicspath{{figs/}}
\usepackage{tikz}
\usetikzlibrary{arrows.meta,positioning,calc}
\usepackage{float}
\usepackage{algorithm}
\usepackage{algpseudocode}

\crefname{figure}{Figure}{Figures}
\Crefname{figure}{Figure}{Figures}
\crefname{table}{Table}{Tables}
\Crefname{table}{Table}{Tables}
\crefname{section}{Section}{Sections}
\Crefname{section}{Section}{Sections}
\crefname{subsection}{Section}{Sections}
\Crefname{subsection}{Section}{Sections}
\crefname{subsubsection}{Section}{Sections}
\Crefname{subsubsection}{Section}{Sections}
\crefname{appendix}{Appendix}{Appendices}
\Crefname{appendix}{Appendix}{Appendices}
\crefname{algorithm}{Algorithm}{Algorithms}
\Crefname{algorithm}{Algorithm}{Algorithms}
\crefname{equation}{Equation}{Equations}
\Crefname{equation}{Equation}{Equations}

\begin{document}
\maketitle

\section{Introduction}\label{sec:intro}

Contemporary video generation systems can synthesize visually compelling,
temporally coherent clips with synchronized audio. Perceptual plausibility,
however, is not evidence of physical validity: a failure may arise not only in
rendered appearance but in any observable consequence of a physical event.
Within the visual stream, entities may be omitted, trajectories may be
inconsistent with the forces implied by the scene, and interactions may violate
contact or causal constraints. Within the acoustic stream, an expected sound
may be absent, mistimed, or attributed to the wrong source; across modalities,
visual and acoustic observations may encode incompatible event timing or
causality. Thus an object may remain unsupported under gravity, a ball may pass
through a rigid wall, a visible collision may be silent, or the sound of an impact
precede contact. These failures become especially consequential as generated
video moves beyond media synthesis into the training and evaluation loops of
embodied systems. Recent video world models have been used to synthesize
robot-training trajectories, support policy learning and evaluation, and reduce
the sim-to-real gap; stronger world models have also been associated with better
downstream policy
performance~\cite{jang2025dreamgen,huang2025enerverse,shang2025roboscape}.
Recently, evaluation has expanded beyond generic video quality to embodied task
correctness and rule-level physical
diagnosis~\cite{li2025worldmodelbench,deng2026rethinking,videophy2_2025,lin2026phyground},
while complementary work has introduced temporal localization and the evaluation
of acoustic or cross-modal physical consistency~\cite{physioneval2026,chinchure2026spotlight,xie2025phyavbench,cui2026avphys}.

\begin{figure}[t]
  \centering
  \input{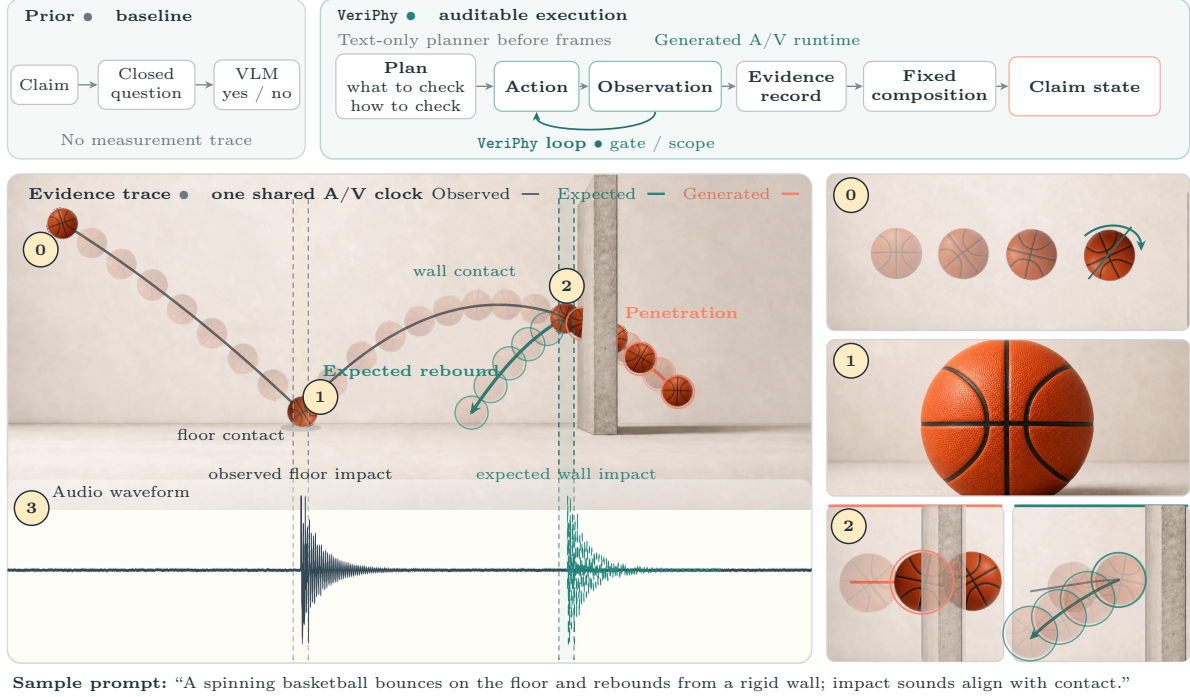}
  \caption{Overview of \VeriPhy{} relative to question decomposition. The
  numbered views magnify rotation, floor-contact deformation, generated wall
  penetration, and the counterfactual rigid-wall rebound; the waveform shares
  the visual event clock. The scene is illustrative rather than a recorded
  execution.}
  \label{fig:teaser}
\end{figure}
\FloatBarrier

Despite this progress, recent evaluators capture complementary components of
physical diagnosis rather than a single auditable evidence chain. They provide
rule-level judgments~\cite{videophy2_2025,lin2026phyground}, temporal
localization~\cite{physioneval2026,chinchure2026spotlight}, measurement-grounded
tests in controlled settings~\cite{wang2026gauge}, and audio-physical or
cross-modal consistency checks~\cite{xie2025phyavbench,zhou2026avgenbench}.
These capabilities are nevertheless typically organized as benchmark-specific
protocols built around curated datasets, predefined phenomena, or manually
authored rubrics. The recent AV-Phys Agent is a particularly close point of
comparison: it combines a reasoning--action loop with deterministic acoustic
tools, but its released protocol remains tied to prompt-specific human-authored
rubrics and multimodal-model binary judgments~\cite{cui2026avphys}. The
ball--wall schematic in \cref{fig:teaser}, which illustrates the interface
rather than a recorded execution, makes the resulting integration gap concrete:
a defensible verdict must establish that the ball and wall are present,
localize floor- and wall-contact windows, measure trajectory and relative
geometry, and determine whether any detected impact sound aligns with the
wall-contact window. Although audio events can be detected in parallel,
synchrony cannot be resolved until that visual window is available. Unusable
evidence leaves the dependent claim unknown; if no claim is contradicted, any
unknown makes the clip abstain, while a terminal infrastructure failure
remains a separate outcome. The gap targeted here is therefore not the absence
of any one component, but their integration within a single evaluator that
compiles the prompt into typed physical obligations and a statically validated
plan, lets observations gate and scope already-declared specialist calls, and
uses fixed rules to compose usable evidence records (typed measurements or
explicitly tagged learned states) into three-valued decisions with traceable
time windows, tool outputs, values, rules, and provenance.

\VeriPhy{} addresses these requirements as a single physical-verification
system comprised of a text-only planner, a video-aware semantic verifier, frozen
specialist operators, and fixed composition. For open-ended prompts that provide
no evaluation rubric, the planner operates before any frame is seen, compiling
exact prompt spans into typed physical obligations and a statically validated
execution plan. For evidence distributed across entities, events, time, and
modalities, the verifier reads timestamped frames densely on a shared A/V clock;
its observations activate, skip, or localize only calls already declared by the
plan, and those scoped calls dispatch SAM~3 grounding and
tracking~\cite{sam3_2025}, identity-based counting, eleven track-based physical
measurements, monocular depth, optical character recognition, and audio-event
detection~\cite{flexsed2025}. For heterogeneous or unusable outputs, each action
emits a claim-bound record containing either a typed measurement or an explicitly
tagged learned base state, together with its realized scope,
measured/abstained/errored status, artifact hashes, cost, and provenance. Typed
resolvers and fixed composition map usable records to supported, contradicted,
or unknown claim states and then to plausible, implausible, or abstain at clip
level; unavailable evidence remains unknown, while terminal infrastructure
failures remain separate. For opaque verdicts, the system returns an auditable
claim-level trace linking each decision to its prompt span, planned calls,
localized evidence, and composition rule. Together, these components turn an
open prompt and generated A/V clip into a dependency-aware evidence chain rather
than an unordered set of model answers. \Cref{fig:teaser} contrasts this design
with the evaluated question-decomposition baseline, which asks one closed
yes/no question per claim without a measurement trace.

To evaluate the critic at the granularity of its trace, we assemble a corpus of
$1{,}500$ generated clips with $2{,}582$ human-written, prompt-grounded flaw
records. Each record quotes the violated prompt span and provides a rationale,
severity, confidence, and, when available, a temporal span and object track;
these records are the only ground truth used here. A flaw-level protocol scores
each system finding as whole, partial, or missed against the corresponding
human record. On a $149$-clip recall analysis, \VeriPhy{} accounts for $228$ of
$304$ annotated defects, compared with $164$ for a published
question-decomposition evaluator given the same clips, claims, and served model.
A monolithic prompt to that same backbone accounts for $222$, so recall alone
does not isolate the value of the agentic organization; the principal
distinction is that every \VeriPhy{} decision retains its supporting evidence
and provenance. This analysis is recall-only, single-annotator, and not held
out, and therefore characterizes the current system rather than its
generalization (\cref{sec:eval}).

\VeriPhy{} casts physically grounded generation of, and a verdict on, a clip $\mathcal{V}$ for its
prompt $p$ as two maps we build and benchmark separately: a \emph{generation} map
$\hat{\mathcal{V}}\sim\operatorname{Gen}(p,C)$, the control-conditioned video diffusion of
\cref{sec:generation}, and an \emph{evaluation} map $Y=\operatorname{Crit}(\mathcal{V},p)$, the
auditable critic of \cref{sec:critic}. Both maps and the clip-level verdict are made precise there.

In summary, our contributions are threefold: (i) \VeriPhy{}, an integrated
physical-verification system that compiles prompts into typed obligations,
scopes specialist operators with observations, and returns three-valued
decisions with localized evidence and provenance; (ii) the $1{,}500$-clip corpus
and its flaw-level matching protocol; and (iii) a simulation-conditioned
generation testbed in which known MuJoCo geometry is rendered as depth or
silhouette control for a frozen Wan~2.2-VACE generator
~\cite{todorov2012mujoco,wan2025,vace2025}. The current report evaluates the
critic and generator separately. Together, these contributions provide an
auditable evaluation stack and a concrete interface for future physical
refinement.

\section{Related Work}\label{sec:related}

\paragraph{Physics-aware generation and refinement.}
Controllable video generation separates appearance synthesis from structural
guidance by conditioning diffusion models on depth, segmentation, reference
frames, masks, motion fields, and compositional spatiotemporal signals
~\cite{zhang2023controlnet,wang2023videocomposer,wan2025,vace2025}. Such
interfaces determine how geometric information enters a generator, but do not
by themselves establish that either the condition or the synthesized pixels
obey physical laws. A complementary line derives control from physical
reasoning or simulation: scene properties can be inferred and reconstructed for
simulation, simulated trajectories or flow can guide synthesis, textual
physical context can refine a prompt, and language can be compiled into a
coarse motion plan. More recent research introduces continuous controls over
physical properties or couples simulation and rendering directly with generation
~\cite{motioncraft2024,physgen2024,thinkbeforediffuse2025,phyt2v2025,vlipp2025,
phyco2026,psivg2026,moregen2026}. Evaluation feedback has also been used for
preference optimization, verifier-guided candidate search, and localized
regeneration~\cite{aifeedback2024,videot1_2025,videorepair2026}. These studies
pursue both pre-synthesis physical conditioning and evaluation-guided
refinement. The present work decouples these two stages: simulator-derived
metric depth or silhouette controls condition a frozen video generator, while
\VeriPhy{} independently evaluates the resulting clip. This design distinguishes
adherence to the control signal from physical validity in the synthesized
output; feeding diagnoses back into regeneration remains future work.

\paragraph{Model judgments, decomposition, and physical evaluation.}
Automated video evaluation has progressed from aggregate metrics toward
increasingly structured model judgments. General protocols factor perceptual
quality, temporal consistency, and prompt alignment, while learned VLM or MLLM
judges map a prompt and clip to multidimensional ratings, discrete states, or
natural-language explanations
~\cite{vbench2024,t2vcompbench2024,fetv2023,evalcrafter2024,
videoscore2024,videoscore2_2025}. Prompt-decomposition methods instead translate
a prompt into propositions, questions, or query chains and revisit the visual
input through smaller checks~\cite{tifa2023,dsg2024,vqascore2024,
han2025videobench}; the checks are explicit, although their supporting evidence
is generally still produced by the judging model. Physics-oriented benchmarks
further introduce
phenomenon-, rule-, trajectory-, and prompt-specific criteria, criterion-level
reasoning, failure localization, controlled measurements, and audiovisual
consistency~\cite{videophy2024,videophy2_2025,phygenbench2024,physicsiq2025,
phycobench2025,phyworldbench2026,lin2026phyground,physioneval2026,
chinchure2026spotlight,luo2026lovif,wang2026gauge,xie2025phyavbench,
zhou2026avgenbench,cui2026avphys}. Prior physical evaluation is therefore not
uniformly scalar, unlocalized, visual-only, or unmeasured; rather, these
capabilities remain distributed across evaluators with different scopes, output
contracts, and evidence representations.

\paragraph{Execution harnesses, context engineering, and persistent state.}
Execution harnesses separate orchestration from model judgment by making
planning, dependencies, specialist routing, observation scope, validation,
termination, and trace logging explicit
~\cite{toolformer2023,gpt4tools2023,hugginggpt2023,rewoo2023,llmcompiler2024}.
Modular visual and video methods extend this organization with specialized
perception, selective observation, and adaptive temporal sampling
~\cite{visprog2023,vipergpt2023,codevqa2023,proviq2023,morevqa2024,
videoagent_wang2024,videoagent_fan2024,videotree2025,caviar2025,lenswalk2026},
while video-evaluation pipelines already combine prompt structuring with
temporal tools, formal specifications and model checking, localized repair,
deterministic acoustic measurements, or prompt-specific rubrics
~\cite{videogeneval2025,neusv2025,neuse2025,cui2026avphys}. Across episodes, a
harness also constructs the context for each model call by storing, retrieving,
filtering, and formatting persistent state; prior work distills trajectories
into reflective lessons, reusable strategies, or structured knowledge without
weight updates, and separates recorded evidence from derived beliefs to reduce
context pollution~\cite{explicitmem2026,contexttraining2026,reasoningbank2026,
reflexion2023,plugmem2026,hindsight2026}. These components predate our work, and
controlled studies show that additional tools or scaffolding may increase cost
without improving accuracy~\cite{findeis2025external,kapoor2026hal}. Relative
to this literature, \VeriPhy{} contributes their integration under a common
verification contract: open prompts are compiled into typed obligations and a
validated plan, observations only gate or scope declared calls, measurements
remain distinct from learned states, and fixed rules produce initial
three-valued claims with any review override preserved separately in the trace.
Across episodes, it uses offline lessons from a disjoint experience pool and
separately evaluates an evolving in-context lesson state
(\cref{sec:selfevolve:measured,sec:leaderboard}). The reported results therefore
establish experience-conditioned harness execution, not autonomous
self-improvement or open-ended replanning; automatic changes to tools,
workflows, admission policies, and external knowledge remain future work.

\section{Physics-Guided Simulation Pipeline}\label{sec:generation}

\begin{figure}[t]
  \centering
  \input{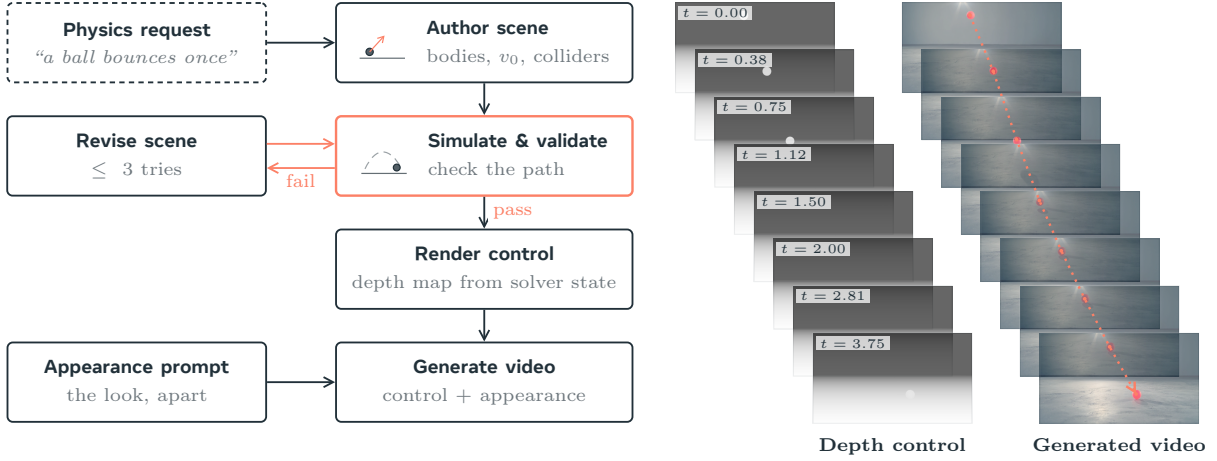}
  \caption{Physics-guided generation. A \emph{physics request} becomes a scene
  that is simulated and validated, revising on failure (\textcolor{veriphyred}{red}~loop);
  the validated motion is a depth control that, with an \emph{appearance prompt},
  conditions the generator. Right: depth control and generated clip at matched
  timestamps, tracing the ball's path.}
  \label{fig:genpipe}
\end{figure}

Modern generators render physically plausible motion, yet prompt-driven synthesis supplies no known
physical target: because physics and appearance are produced jointly, the intended trajectory,
contact, or counterfactual is neither fixed in advance nor recoverable afterward, and every physical
property stays confounded with appearance. An evaluation testbed instead needs a target factored out
of appearance---specified, known by construction, and auditable independently of generator
capability. We obtain one from a trusted physics solver and render it as a geometric control, so the
solver fixes the motion while the prompt supplies only appearance (\cref{fig:genpipe}). This is developed across two stages: \cref{sec:gen-control} turns the event description into a validated target rendered as a
registered control, and \cref{sec:gen-mechanism} conditions the generator on that control and an
appearance prompt. Whether the generated video realizes the target is the open question of
\cref{sec:gen-experiments}.

\subsection{Preliminaries}
\label{sec:gen-prelim}

The atomic unit is a \emph{clip--prompt pair}: a clip
$\mathcal{V}=\{(I_t,\tau_t)\}_{t\in\mathcal{T}_{\mathcal V}}$, a time-ordered sequence of frames $I_t$
at timestamps $\tau_t$ with duration $D_{\mathcal V}$ (\cref{sec:critic}), and a natural-language
prompt $p$. The prompt supplies appearance only (subject, material, setting), the geometric control
$C$ carrying the physical target---a solver-rendered video with per-frame per-pixel depth intensity
$C_t(u)$ (\cref{sec:gen-control}); $\hat{\mathcal{V}}$ denotes a generated sample, $\mathcal{V}$ any
clip fed to the critic. Generation and evaluation are thus two maps over a clip and its prompt, built
and benchmarked separately,
\begin{equation}
  \hat{\mathcal{V}}\sim\operatorname{Gen}(p,C)
  \qquad\qquad
  Y=\operatorname{Crit}(\mathcal{V},p) ,
  \label{eq:gen-crit-maps}
\end{equation}
where $\operatorname{Gen}$ samples a clip conditioned on $p$, $C$, a mask $M$ and a scalar strength
$s$---asserting no physical fidelity---and $\operatorname{Crit}$ is the auditable critic \VeriPhy{}
of \cref{sec:critic}, returning the clip-level verdict $Y$ (\cref{eq:claim-clip-rollup}).

\subsection{From specification to registered control}\label{sec:gen-control}

This stage produces the geometric control, validating the motion it carries before any control is
rendered, so the generator is conditioned on a verified trajectory rather than an unconstrained
request. It is a three-map chain from the event description $p_{\mathrm{ev}}$ to the control $C$,
\begin{equation}
p_{\mathrm{ev}}
\;\xrightarrow{\;\textsc{Author}\;}\; \mathcal{S}
\;\xrightarrow{\;\textsc{Sim}\;}\; \tau
\;\xrightarrow{\;\textsc{Render}(\,\cdot\,;\,\mathcal{S})\;}\; C,
\label{eq:control-chain}
\end{equation}
where $\textsc{Author}$ prompts a language model to compile the text into a structured scene specification
$\mathcal{S}$ (bodies, initial states, colliders, static surfaces, and a camera; \cref{fig:authoring}),
$\textsc{Sim}$ runs the physics solver to a trajectory $\tau$, and $\textsc{Render}$ converts the
solver state into the depth control $C$ of \cref{sec:gen-control}. The middle map is gated by a
validation loop (\cref{eq:scene-validate}) so only a $\tau$ meeting the event's physical predicates is
rendered; the three maps are developed in turn below.

A body is simulated from its initial state when the engine determines the motion (a ballistic arc and
its contacts), while prescribed waypoints are reserved for externally actuated or explicitly
counterfactual motion. Validation applies to the simulated trajectory, not the specification text,
through a conjunction of event-specific predicates $\Phi_{\mathrm{ev}}$: a ballistic segment must form
an arc, satisfy the required hit or miss relation, settle when requested, and remain visible. The
$\textsc{Sim}$ map is therefore a bounded revise-and-resimulate loop, $\operatorname{Solve}$ running
the solver to a trajectory and $\textsc{Summarize}$ reducing it to a text report of its realized
motion. Starting from $\mathcal{S}^{(0)}=\mathcal{S}$, for round $k=0,1,2$,
\begin{equation}
\begin{aligned}
\tau^{(k)} &= \operatorname{Solve}\!\big(\mathcal{S}^{(k)}\big),\\
\mathcal{S}^{(k+1)} &= \textsc{Author}\!\big(\mathcal{S}^{(k)},\, \textsc{Summarize}(\tau^{(k)})\big),
\end{aligned}
\label{eq:scene-validate}
\end{equation}
where the revision on the second line is taken only while $\neg\,\Phi_{\mathrm{ev}}(\tau^{(k)})$: a
failed predicate returns a summary of the realized motion to the authoring model, which revises the
specification, for at most three revisions (four simulations). The first $\tau^{(k)}$ with
$\Phi_{\mathrm{ev}}(\tau^{(k)})$ true is the validated trajectory $\tau$ of \eqref{eq:control-chain},
passed to $\textsc{Render}$; exhausting the budget yields an explicitly marked unvalidated fallback
rather than a validated scene.

The $\textsc{Render}$ map turns the validated trajectory $\tau$ and its scene $\mathcal{S}$ into the
control $C$. Rendered from solver state rather than inferred from a generated video, $C$ is registered
to the intended motion by construction, all streams sharing the scene's frame clock (solver, camera,
and playback settings deferred to \cref{sec:gen-setup}). For each of the clip's $T$ frames, rendering
places $\mathcal{S}$'s geometry at the per-frame pose, images it with the scene camera, and reads a
depth buffer $D_t(u)$---camera distance at pixel $u$ in frame $t$---normalized to a near-bright,
far-dark intensity $C_t(u)=1-\operatorname{clip}\big((D_t(u)-d_-)/(d_+-d_-),0,1\big)\in[0,1]$, where
$d_-,d_+$ are the $1$st and $97$th percentiles of the non-background depths pooled over the whole clip
and $\operatorname{clip}$ clamps to $[0,1]$; the control $C=(C_t)_{t=0}^{T-1}$ then conditions the
generator in \cref{sec:gen-mechanism}. Background ($D_t(u)$
beyond $d_+$) maps to $0$; pooling the percentiles once fixes the mapping so the same distance is the
same intensity throughout, the cut asymmetric because the far tail abuts the background and is
dominated by depth-edge noise, with a small Gaussian blur softening rasterized boundaries. When
$d_+>d_-$ fails (as a result of no non-background pixels, coincident percentiles, or a predeclared flat
scene) $\textsc{Render}$ falls back to a binary moving-body silhouette, depth otherwise preferred for
preserving both outline and distance ordering.

\subsection{Physics-conditioned video generation}\label{sec:gen-mechanism}

The generator is a latent video-diffusion transformer that samples a clip
by iteratively denoising a latent from noise. We use Wan~2.2~\cite{wan2025}: a DiT-style denoiser run
along a flow-matching schedule of $S$ UniPC multistep iterations, indexed by $i$ from high noise to
low as $\sigma_i$ decreases (\cref{eq:flow-sigma}). On its own it is a text-to-video generator
conditioned only on an appearance prompt $p$ (subject, material, setting), with no input through which
a specified motion can be imposed.

To make the motion controllable we adopt
VACE~\cite{vace2025}, which augments a frozen backbone with a \emph{control branch}: a control video
and a generation mask are encoded into control features added into the backbone's transformer blocks,
so an external spatiotemporal signal steers generation without retraining the base
weights~\cite{zhang2023controlnet,wang2023videocomposer}. The composite Wan~2.2-VACE takes four
inputs: the appearance prompt $p$, a geometric control video $C$, a generation mask $M$, and a scalar
conditioning strength $s$.

We drive VACE's control branch with the registered depth
control of \cref{sec:gen-control}: $C$ is the depth video $(C_t)_{t=0}^{T-1}$ carrying the validated
motion, and the mask decides which pixels VACE may synthesize. We set $M$ all-white, so every output
pixel is generated and no pixel of the simulator render is copied in---the control acts only through
the branch, never by pasting appearance. The branch then injects, at each transformer layer $\ell$ and
step $i$, an additive update to the hidden state $h_i^{(\ell)}$; with $z^{\mathrm{ctrl}}$ the encoded
control and $W^{(\ell)}$ the branch's projection at layer $\ell$,
\begin{equation}
h_i^{(\ell)} \;\leftarrow\; h_i^{(\ell)} \;+\; s\,W^{(\ell)}z^{\mathrm{ctrl}},
\label{eq:control-release}
\end{equation}
so the control enters as a strength-$s$ residual rather than by overwriting pixels. The same $s$ is
applied at every control layer and stays active throughout sampling in the deployed generator, lowered
when the control covers little of the frame. Conditioned on $(p,C,M,s)$ with a fixed seed and a set of
excluded terms, the backbone induces a distribution over clips from which
$\hat{\mathcal{V}}\sim\operatorname{Gen}(p,C)$ is drawn. It should be noted that this is just a sample, asserting no physical fidelity.

\subsection{Control withdrawal as a mechanism probe}\label{sec:gen-withdrawal}

\emph{When} can the control be released? The injection of \eqref{eq:control-release} runs throughout
sampling, but if the trajectory settles by some intermediate noise level the control is redundant
afterward. Switching it off partway is awkward for VACE's additive residual, so we study the question
on a \emph{separate} Wan2.2-derived backbone whose control path is easy to interrupt---a mechanism
probe, not the deployed generator---and read the commitment point back as a property of the Wan~2.2
diffusion prior both share. Run at $40$ UniPC steps with flow shift $5.0$, this backbone has no
trained control branch: its only conditioning path encodes a rendered source video into context
tokens concatenated with the text and image tokens, so withdrawal is directly available---the source
tokens are removed from the attended sequence at a chosen step, distinct from the additive-residual
injection of \eqref{eq:control-release}: over $S$ steps the source tokens are present at step $i$ iff
$i<i^{*}$, where $i^{*}=\operatorname{clip}(\operatorname{round}(fS),0,S-1)$ and $f\in[0,1]$ is the
withdrawal fraction. Steps before $i^{*}$ attend over the source tokens at the scale
$\omega$ the deployed recipe uses; from $i^{*}$ onward the sequence is identical to that of an
unconditioned generation.

Because $i^{*}$ is an index, its noise level depends on the schedule, so the withdrawal point is
reported as $\sigma_{\mathrm{rel}}=\sigma_{i^{*}}$, the level the flow-matching schedule assigns step
$i^{*}$ (the shift-dependent map is \eqref{eq:flow-sigma} in \cref{sec:app-schedule}); two runs at the
same step index thus hit different noise levels whenever $S$ or shift $\kappa$ differs, which is why
\cref{sec:gen-release} indexes by $\sigma_{\mathrm{rel}}$. Finally, the multistep solver forms each
update from derivatives cached at previous steps, which after a withdrawal still carry the removed
tokens' influence; the measurement compares one variant that clears that history at $i^{*}$ against
one that keeps it, otherwise identical.

\section{\VeriPhy{}}\label{sec:critic}\label{sec:compiler}

This section presents \VeriPhy{}, an auditable multi-tool critic that adjudicates
each generated clip against its prompt, returning a three-valued verdict whose every
decision traces back to the evidence that produced it. We score the critic against
the human-annotated flaw benchmark of \cref{sec:dataset} under the matching protocol
of \cref{sec:eval}.

\subsection{System overview}
\label{sec:critic-react}

\VeriPhy{} is a constrained \emph{reason--act--observe} critic with four roles. A \emph{text-only planner} compiles the prompt into typed physical claims and an executable measurement plan before seeing any frame (\emph{reason}). Each \emph{act} step dispatches one of two backends: \emph{specialist instruments} returning targeted measurements---counts, tracks, masks, depths, text, and sound intervals---or a \emph{video-aware semantic verifier} reading timestamped frames densely for predicates like existence and event occurrence. Each \emph{observe} step consumes the returned measurement to gate, scope, and---via a \emph{deterministic aggregator}---compose the plan's declared physical relations into a three-valued verdict. The governing invariant: tool outputs steer execution, but only within obligations declared before any pixel is read. Backends are listed in \cref{tab:specialists}; the planner and semantic verifier may share a served checkpoint but are separate calls with different inputs and responsibilities. We situate this design in the ReAct and tool-using-agent lineage in \cref{sec:framework:genealogy}.

Two properties make the critic auditable. First, reasoning sits entirely up front: the planner compiles and \emph{statically validates} the typed measurement plan before observing the video, following plan-before-act agents~\cite{rewoo2023,llmcompiler2024} and code-as-plan systems that emit an executable program of visual primitives~\cite{vipergpt2023,visprog2023}. Planning is thus \emph{global with respect to the prompt}---the naming pass sees the whole prompt and later claim groups share its vocabulary---while execution is \emph{local in time}: observations activate, skip, or localize already-declared nodes. With the trajectory fixed before execution, every measurement traces to the obligation that requested it. Second, evidence production and adjudication stay separate: specialist operators return typed measurements, learned semantic base states are tagged in their evidence records, and deterministic composition over the usable records yields the verdict. 
These properties give a traceable evidence chain behind every three-valued decision: \supported{}, \contradicted{}, or \unknownv{}. These states are surfaced as \emph{plausible}, \emph{implausible}, or \abstain{}, respectively, and formalized in \cref{sec:semantics}.

We now fix notation for the objects these roles operate on. Recall the clip--prompt
pair of \cref{sec:gen-prelim}; the critic addresses frames and, when present, an audio
track on the clip's own container clock, writing the clip as
$\mathcal{V}=\{(I_t,\tau_t)\}_{t\in\mathcal{T}_{\mathcal V}}$ over frame indices
$\mathcal{T}_{\mathcal V}=\{0,\ldots,T_{\mathcal V}-1\}$ with ordered timestamps
$0\leq\tau_0<\cdots<\tau_{T_{\mathcal V}-1}\leq D_{\mathcal V}$, where $I_t$ is frame $t$,
$\tau_t$ its timestamp, and $D_{\mathcal V}$ the clip duration; separating index from time
permits nonuniform decode timestamps. From the
prompt the planner produces exact prompt-grounded claims and a surface plan,
\begin{equation}
\mathcal{C}(p)
  = \{c_i=(s_i,\phi_i)\}_{i=1}^{N_p},
\qquad s_i\sqsubseteq p,
\qquad
(\mathcal{C},\widetilde{\Pi})=\Pi_{\theta}(p),
\label{eq:claims-and-plan}
\end{equation}
where $\Pi_{\theta}$ is the frozen planner, $\widetilde{\Pi}$ the surface plan, $N_p$ the
claim count, $s_i$ a verbatim span of $p$, and $\phi_i$ its checkable assertion; the relation
$s_i\sqsubseteq p$ is enforced, not inferred after execution, and types belong to the plan's
operations and checks rather than to a free-form claim label. \Cref{fig:criticpipe} summarizes the
dataflow and \cref{alg:critic} gives the same one-clip procedure as executable pseudocode.

\subsection{Plan synthesis and static validation}
\label{sec:plan-construction}
This subsection expands the planner map $\Pi_\theta$ of \eqref{eq:claims-and-plan} into two stages producing the claim set $\mathcal{C}$ and surface plan $\widetilde{\Pi}$. A first pass over the whole prompt fixes a shared namespace: a stable identifier for every distinct entity and every separately asserted event occurrence. Claims are then partitioned into groups of at most four, each planned independently against the full prompt and the shared names, with a group's local identifiers mechanically renamed into the shared namespace before the groups are concatenated. This shared-namespace planning with rename-on-assembly prevents two groups from silently referring to the same entity by different names, or to different occurrences by one name.

The surface language follows code-as-plan visual programming
~\cite{vipergpt2023,visprog2023}, but is deliberately restricted to the
operations in \cref{tab:planops}. For claim $c_i$, the planner writes
\code{check(ci, <expr>)} with one operation or a conjunction of operations. At
this stage the planner has no video input: the line states what evidence would
settle the claim, not whether the claim is true.

\begin{table}[t]
\centering
\small
\setlength{\tabcolsep}{6pt}
\renewcommand{\arraystretch}{1.2}
\begin{tabularx}{\textwidth}{@{}L L p{3.0cm}@{}}
\toprule
\textbf{Surface operation} & \textbf{Question posed} & \textbf{Typed check} \\
\midrule
\code{judge("assertion")} & dense-frame semantic verification of the complete assertion & general semantic \\
\code{exists("object")} & whether the object is visible at any decoded time & existence \\
\code{eq(count("phrase"), N)} & whether measured cardinality equals $N$ & count \\
\code{before(window(A), window(B))} & whether event $A$ precedes event $B$ & temporal order \\
\code{during(window(A), window(B))} & whether $A$ is temporally contained in $B$ & temporal containment \\
\code{rel("A||B", relation)} & a fixed-vocabulary spatial relation & spatial \\
\code{traj("phrase", relation, target)} & a fixed-vocabulary motion predicate & trajectory \\
\code{sound("description")} & whether the audio track contains the event & audio occurrence \\
\code{text("STRING", carrier=...)} & whether the named surface renders the string & rendered text \\
\bottomrule
\end{tabularx}
\caption{The planner's surface vocabulary. \code{count} and \code{window} are
structural terms. The deterministic compiler expands a surface operation into
its required execution closure. Thus \code{before} adds occurrence gates and
two timing actions; \code{rel} adds co-visibility and mask actions and, for
\emph{behind} or \emph{in front of}, depth actions. These dependencies need not
be written as extra surface operations, but they must be present in the compiled
graph. A sound operation reads the audio track; no audio track yields
\unknownv{}, not a negative measurement.}
\label{tab:planops}
\end{table}

The assembled surface plan is validated as a whole, since its binding constraints are cross-references between claims, checks, and actions (\cref{eq:plan-validity}). A failing group may be re-asked with its own parser or type complaints and a format restatement, but never with a suggested answer. After a fixed number of attempts, an unusable claim falls back to direct semantic verification; this fallback is recorded and charged as a model call, not a specialist measurement. A failed naming pass routes all claims through the same charged fallback, so plan failure never makes a prompt appear cheaper by silently dropping claims.

\medskip\noindent\textbf{Typed representation and static validation.}\quad
\label{sec:plan-lang}The compiler turns the surface program into a typed directed graph
\begin{equation}
G_{\Pi}=(\mathcal{A},\mathcal{Q},\mathcal{E},b),
\qquad
a_j=(o_j,\alpha_j,\omega_j,r_j),
\qquad
b:\mathcal{Q}\rightarrow\{1,\ldots,N_p\}.
\label{eq:plan-dag}
\end{equation}
Here $\mathcal{A}$ contains evidence-producing actions, including general
semantic passes and specialist calls; $\mathcal{Q}$ contains typed checks;
$\mathcal{E}$ points from a prerequisite to its consumer; and $b(q)=i$ binds
check $q$ to claim $c_i$. An action records its operation $o_j$, arguments
$\alpha_j$, declared scope policy $\omega_j$, and result type $r_j$.

Before any action runs, the graph must satisfy
\begin{equation}
\operatorname{Valid}(G_{\Pi})
=
\operatorname{Parse}(G_{\Pi})
\land \operatorname{Refs}(G_{\Pi})
\land \operatorname{Typed}(G_{\Pi})
\land \operatorname{Acyclic}(G_{\Pi})
\land
\left[\forall i,\ \exists q\in\mathcal{Q}:b(q)=i\right].
\label{eq:plan-validity}
\end{equation}
$\operatorname{Parse}$ checks that the surface program parses;
$\operatorname{Refs}$ checks identifiers and dependency closure;
$\operatorname{Typed}$ checks operator arity, fixed relation vocabularies, and
whether every check receives the result types it requires;
$\operatorname{Acyclic}$ checks that the dependency graph has no cycle. For example, a
directed trajectory must carry a target, and a depth-ordering check must consume
both masks and depth. An invalid whole line falls back to the general verifier;
an invalid term can be replaced by a recorded generalist sub-check. In both
cases the substitution is charged and reported rather than guessed.
\Cref{fig:plangraph} in the appendix shows a compiled plan of this form, with the dependency
closure that \eqref{eq:plan-validity} requires made explicit.

\subsection{Timeline-conditioned multi-tool execution}
\label{sec:chaining}
Execution follows a topological order. Each action's compiled scope function $g_j$ maps prerequisite records to the full frame index set or a declared local subset, then calls the appropriate backend:
\begin{equation}
\Omega_j
  =g_j\!\left(\{R_k:(a_k,a_j)\in\mathcal{E}\},
              \mathcal{T}_{\mathcal V}\right),
\qquad
R_j
  =F_{o_j}\!\left(\mathcal{V}|_{\Omega_j},\alpha_j\right).
\label{eq:scoped-execution}
\end{equation}
Here $F_{o_j}$ is the frozen backend implementing operation $o_j$ on the scoped
video $\mathcal{V}|_{\Omega_j}$ with typed arguments $\alpha_j$; the scope function is
part of the validated program. A usable measured window is padded, given a minimum
duration, and converted to the explicit indices whose timestamps fall inside it; a
missing, invalid, or overly broad window falls back to the whole clip---a conservative
fallback that avoids reading failure to localize as evidence that the requested event or
relation is absent. The backend behind each capability, with its returned object and
exact frozen model, size, and license, is listed in \cref{tab:specialists}. Three
sub-checks remain learned, resisting full symbolization---subtype identity, dense
action-event recall, and depth-ordered spatial interpretation on localized
evidence---and their outputs carry a learned base-predicate tag distinguishing them from
deterministic consequences of a measurement, so no holistic scorer appears in the verdict
path. Two actions share one execution only when their complete realized signatures
match---clip identity, operation $o_j$, typed arguments $\alpha_j$, references $r_j$, and
realized scope $\Omega_j$---and the executor calls each equivalence class once, following
exact common-subexpression reuse~\cite{sellis1988mqo,roy2000mqo}; because the realized
scope is part of the signature, whole-clip and localized requests never merge merely
because their natural-language queries are similar.

\begin{figure}[tp]
  \centering
  \includegraphics[width=\linewidth]{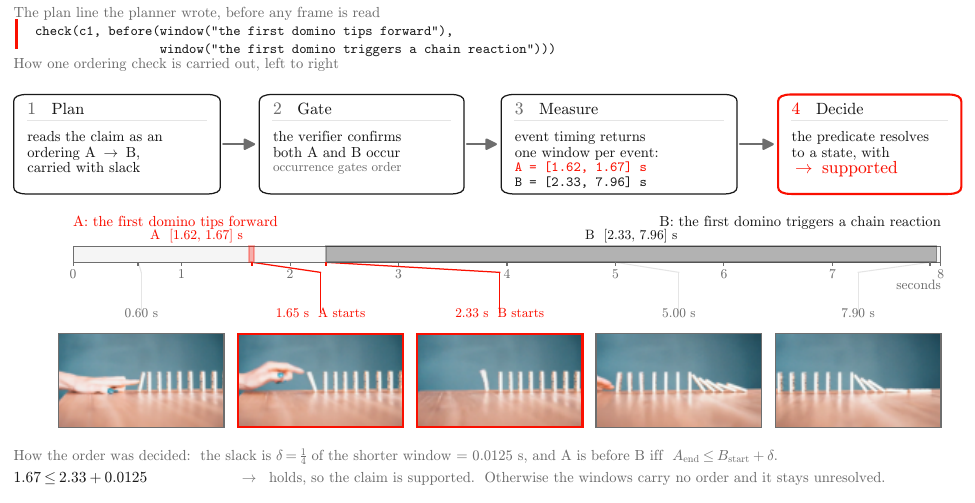}
  \caption{A real temporal trace. The planner writes the line before reading
  any frame. The verifier first confirms both events, the timing actions then
  return $W_A=[1.62,1.67]$\,s and $W_B=[2.33,7.96]$\,s, and
  \eqref{eq:temporal-before} resolves their order. Every number and displayed
  frame comes from the recorded run.}
  \label{fig:planchain}
\end{figure}

Timing is the one capability needing a guard before its windows are used: a returned window is not evidence the queried event occurred, since a forced localizer may return its best candidate even for an absent event. Visible events are thus confirmed first by the dense-frame semantic verifier, audible events on the audio track, and only supported occurrences let their intervals enter a temporal predicate. Writing the sub-check states over the three-valued codomain $\mathbb{V}=\{\mathsf{S},\mathsf{C},\mathsf{U}\}$ (\supported{}, \contradicted{}, \unknownv{}; \cref{sec:semantics}), a relation $R\in\{\mathrm{before},\mathrm{during}\}$ over occurrence states $o_A,o_B$ and measured windows $W_A,W_B$ resolves to \contradicted{} if either operand is \contradicted{}, to \unknownv{} if either is \unknownv{} or a window is unavailable, and to the order predicate $\rho_R(W_A,W_B)$ only when both events are supported---so timing may compose an order but never establish occurrence. That predicate $\rho_R$ compares interval endpoints under a boundary slack of one quarter of the shorter window (\cref{eq:temporal-before} in \cref{sec:app-temporal}): an order is supported or contradicted only when one window ends before the other begins by more than the slack, otherwise $\mathsf{U}$; the quarter-window rule prevents both directions from holding. Containment (\emph{during}) uses the same slack. These predicates are a tolerance-scaled subset of Allen's interval algebra~\cite{allen1983intervals}. \Cref{fig:planchain} traces one such ordering through a recorded run, from the plan line written before any frame was read to the two measured windows that settle it.

\subsection{Physical measurements and evidence records}
\label{sec:physics}
Physical claims are settled by eleven operations sharing one grounding step: SAM~3 segments the named subject and tracks its identity across frames, and every operation measures over those tracks. Each maps to a class of human complaint in the corpus.

The largest class is displacement: the dominant complaint is that something the prompt sets in motion does not move or moves the wrong way (e.g.\ a static sun), with ballistic arcs comparatively rare. Three conventions make it measurable: it is reported in multiples of the subject's own radius, so no scene calibration is needed; direction is taken from the furthest excursion, not the net, since a rise then fall cancels; and camera motion is removed by dense whole-frame optical flow, needing no stationary landmark. Three further operations cover failures with no net travel: contact measures the closest approach between two tracked masks (finer than box overlap); shape change measures surface motion relative to the background flow (a scroll rolling in place, which displacement reports as zero); and path shape distinguishes a circle from a figure of eight. The rest cover time and existence: permanence, whether a subject that should vanish persists; hop counting, whether one that should leave the ground does; damping, a spinning object that never slows; onset ordering, which of two events comes first (or never happens); clip timing for duration and playback rate; and viewpoint motion, separating a moving camera from a moving subject.

All eleven return a number with its evidence---frames read, tracker used, threshold applied---leaving the decision to \cref{sec:semantics}. Each may also abstain: a subject under one percent of the frame, or a track too short to measure, yields an explicit abstention, so an unmeasurable claim never resolves as supported.

\medskip\noindent\textbf{Evidence records and provenance.}\quad
\label{sec:contract}Each action is serialized as a claim-bound record
\begin{equation}
R_{j,i}
=
\bigl(\operatorname{id}(\mathcal V),p,s_i,o_j,\Omega_j,
       \sigma_j,\mu_j,h_j,\kappa_j\bigr),
\qquad
\sigma_j\in
\{\mathtt{measured},\mathtt{abstained},\mathtt{errored}\}.
\label{eq:evidence-record}
\end{equation}
The payload $\mu_j$ contains either a typed measurement or an explicitly tagged
learned base state; $h_j$ contains the native-output and artifact hashes; and
$\kappa_j$ contains calls and accelerator time, with optional wall time, price,
and token counts. Shared execution may bind the same record to several claims,
but its cost is charged once by execution identifier.

The three statuses are mutually exclusive: \code{measured} carries a payload and no abstention or error; \code{abstained} carries a reason and no payload; \code{errored} carries an infrastructure message. A valid negative measurement---a complete usable track lacking the requested motion---stays \code{measured} and may contradict a claim, whereas failure to track from blur, occlusion, low quality, or an instrument refusal is \code{abstained}, not a negative observation. The counting path can blur this: the mode of an intermittently grounded per-frame series can report a confident zero when grounding merely failed on most frames, indistinguishable from a true absence---hence the auxiliary fields (\code{stable}, range, frames read) are load-bearing, and \cref{sec:limitations} flags the consequence for count-derived figures. A terminal \code{errored} record is retried; if still terminal, we mark the clip $\bot_{\mathrm{infra}}$ and exclude it from the evaluation set rather than relabeling it \unknownv{}.

Specialist schemas reject judgment-shaped fields, nonfinite numbers, duplicate keys, and noncanonical payloads; every record carries the full prompt and verbatim span $s_i$, with heavy artifacts content-addressed by path and byte hash. This guarantees a specialist cannot emit the final prompt-satisfaction label and that accepted evidence is attributable and replayable---though it cannot unbias a learned measurement nor detect a well-formed but wrong payload.

\subsection{Verdict aggregation and diagnostics}
\label{sec:semantics}
A deterministic roll-up composes per-sub-check states into per-claim states and then the clip
decision (\cref{eq:claim-clip-rollup}); two learned refinements then act on top of it, and the run
emits a localized contradiction packet (\cref{eq:refinement-packet}) as its diagnostic interface.
For a completed trace, every sub-check resolves in
$\mathbb{V}=\{\mathsf{S},\mathsf{C},\mathsf{U}\}$, standing for \supported{},
\contradicted{}, and \unknownv{}. A typed resolver $\psi_q$ maps the measurements
$\mu_j$ of its prerequisite actions $\operatorname{Pa}(q)$ in $G_{\Pi}$ to a
sub-check state $z_q$: when all required actions are \texttt{measured} it applies
$\psi_q$; if any abstained or returned no usable evidence the state is
$\mathsf{U}$; and a terminal infrastructure error gives $\bot_{\mathrm{infra}}$.
The resolver is deterministic for count, timing,
track, transcription, and other explicit predicates, and learned only for the
semantic sub-checks named above, where it uses an asymmetric gate: it may
contradict only when a positively contradicting feature is visible, so that
unresolved evidence, unparseable output, or a missing modality yields
$\mathsf{U}$ rather than turning an instrument failure into an accusation about
the video.
Negation is applied to the positive sub-check state, never by asking a tool to
measure a negative quantity: $\neg_3\mathsf{S}=\mathsf{C}$,
$\neg_3\mathsf{C}=\mathsf{S}$, and $\neg_3\mathsf{U}=\mathsf{U}$.
Let $K_i=\{q\in\mathcal Q:b(q)=i\}$ be the validated checks for claim $c_i$,
after applying this negation where required. Claim and clip composition are
\begin{equation}
y_i=
\begin{cases}
\mathsf{C}, & \exists q\in K_i:z_q=\mathsf{C},\\
\mathsf{S}, & \forall q\in K_i:z_q=\mathsf{S},\\
\mathsf{U}, & \text{otherwise},
\end{cases}
\qquad
Y=
\begin{cases}
\bot_{\mathrm{infra}}, & \text{the trace terminates with an infrastructure error},\\
\textsc{Implausible}, & \exists i:y_i=\mathsf{C},\\
\textsc{Plausible}, & N_p\geq1\ \land\ \forall i:y_i=\mathsf{S},\\
\textsc{Abstain}, & \text{otherwise}.
\end{cases}
\label{eq:claim-clip-rollup}
\end{equation}
One contradiction makes the clip implausible, and \textsc{Plausible} asserts only over the claims the plan executed.

Those two refinements are as follows. First, a stronger reviewing model re-examines claims the video-aware verifier passed, and its verdict stands\label{sec:review}: overruling a pass resolves the claim to \contradicted{} rather than \unknownv{}, replacing $y_i$ where the sub-checks would otherwise roll up to $\mathsf{S}$. Both verdicts are recorded, so a row can be re-scored on the first verifier's alone. The direction is asymmetric by design: the reviewer examines only passed claims, never a specialist measurement, which carries evidence a second opinion cannot inspect. Second, a recognition engine reads frames directly to settle text claims: a generative model asked what a sign says repairs what it sees---rendering a malformed \code{Stly your} as the well-formed \code{Style your} the prompt requested, concealing the defect under test---whereas a recognition head returns the malformed string, which a deterministic comparison rejects. It reads every frame rather than a sampled fraction, since a caption present for part of a clip can fall between samples, and each record states the frames read.

Beyond the clip state $Y$, a run emits a localized contradiction packet\label{sec:loop}
\begin{equation}
\mathcal{F}(p,\mathcal V)
=
\left\{
\bigl(c_i,\{R_{j,i}\}_j,\Omega_i,B_i\bigr)
:\ y_i=\mathsf{C}
\right\},
\label{eq:refinement-packet}
\end{equation}
where $\Omega_i$ is the union of available temporal support and $B_i$ contains available box
tracks or masks. The packet is assembled from the trace that produced the verdict, so each
contradicted claim carries both the evidence records behind the conclusion and the region of
the clip in which the failure occurs: \emph{what} failed, \emph{on what evidence}, and
\emph{where}. That space--time localization is what \cref{sec:framework} maps onto a local edit
of the generator control (\cref{sec:framework:control}, \eqref{eq:control}).
Closed-loop refinement is not run in this report: \eqref{eq:refinement-packet} defines the
interface only, the critic and simulation-conditioned generator are evaluated on disjoint data,
and no \VeriPhy{} packet has been used for preference optimization, denoising guidance, candidate
selection, or localized regeneration---so we claim no clip has been made more physically correct
by the critic.

\section{Experiments}\label{sec:experiments}

This section evaluates the system in the order the pipeline builds it. \Cref{sec:dataset} sets out
the shared setup---the human-annotated flaw benchmark and the recall metrics and matching protocol
every result is scored under. \Cref{sec:gen-experiments} takes the \emph{physics-guided generator}:
whether the geometric control governs the motion, when the trajectory commits, and how far the scene
library reaches. \Cref{sec:eval} takes the \emph{\VeriPhy{} critic}: its flaw recall on the
evaluation core, against the annotated reference and published evaluators, and its agreement with
human ratings. \Cref{sec:icl-exp} takes the \emph{context-training self-evolution} of the
critic---whether lessons distilled from recorded misses raise recall without touching a
weight---and \cref{sec:refine} closes the critic's outward channel, feeding a verdict back as a
prompt rewrite that regenerates the clip. The evaluations use distinct held-out settings that should
not be conflated; \cref{sec:eval-scope} sets them side by side.

\subsection{Experimental setup}\label{sec:dataset}
This setup---models, benchmark, and matching protocol---is shared by both systems; each system's own configuration accompanies its results.

\paragraph{Models and specialists.} The critic runs one served vision-language model for both its text-only planner and its video-aware semantic verifier---Qwen3-VL-30B-A3B-Instruct~\cite{qwen3vl2025}, in BF16 without quantization---together with a frozen stack of low-level specialists, one per capability: SAM~3~\cite{sam3_2025}, TAPNext++~\cite{tapnext2025}, a depth model, PaddleOCR, and FlexSED~\cite{flexsed2025}, with a closed model as the fallback verifier. \Cref{tab:specialists} lists, for each capability, the backend that runs it and the object it returns, together with the exact checkpoint, size, and license; every component is frozen at the listed revision for all reported results.

\begin{table}[t]
\centering\small
\setlength{\tabcolsep}{5pt}
\renewcommand{\arraystretch}{1.25}
\begin{tabularx}{\textwidth}{@{}p{6.35cm} X X@{}}
\toprule
\textbf{Capability (model)} & \textbf{Running backend} & \textbf{Returned object} \\
\midrule
Planner (Qwen3-VL-30B-A3B, 31\,B)~\cite{qwen3vl2025}
  & prompt compiled pre-frame & claims and a typed plan \\
Semantic predicate (Qwen3-VL)
  & dense timestamped pass & learned base state, rationale \\
Event timing (Qwen3-VL)
  & over timestamped frames & interval, absence, or refusal \\
Fallback verify (gpt-5.6-sol)
  & on a claim the dense pass passed & three-valued state, both kept \\
Grounding, masks (SAM~3, 0.9\,B)~\cite{sam3_2025}
  & SAM~3 Video, persistent IDs & per-frame masks, tracked IDs \\
Object count (SAM~3)
  & cardinality of SAM~3 IDs & count, per-frame series \\
Physical measurement (TAPNext++)~\cite{tapnext2025}
  & 11 typed operations over tracks & typed measurement, evidence \\
Depth ordering (SenseNova-7B-MoT)
  & depth model on localized evidence & behind / in-front ordering \\
Camera cancellation (from tracks)
  & dense optical flow subtracted & subject-relative displacement \\
Rendered text (PaddleOCR PP-OCRv5)
  & OCR head + EasyOCR fallback & literal glyph transcription \\
Audio (FlexSED)~\cite{flexsed2025}
  & sound-event detection & sound intervals on the track \\
\bottomrule
\end{tabularx}
\caption{Execution backends and the frozen specialist behind each capability, for the
evaluated run: each row names a capability with the model or checkpoint that runs it (with
its parameter count where published), the backend it runs, and the object it returns.
SAM~3 supplies grounding, masks, and counting, while TAPNext++ supplies the tracks the
eleven typed operations (\cref{sec:physics}) are taken over; the two learned components
(semantic predicate and depth ordering) and the fallback verifier are the only backends
that may resolve a claim, and no holistic video-quality model produces the clip decision.
Every component is frozen at the listed revision for all reported results, and
\texttt{gpt-5.6-sol} is a closed model (SenseNova SOL) served through an internal proxy.
All released components are under permissive licenses (Apache-2.0, MIT, or Meta's SAM
license) except the depth-ordering model, which is CC-BY-NC-4.0 (non-commercial).}
\label{tab:specialists}
\end{table}

\subsubsection{Benchmark}\label{sec:bench-spec}
The benchmark is a corpus of $1{,}500$ AI-generated clips: a person watched each against its prompt and recorded every place the video failed to deliver what the text asked. The generating model is not recorded, no clip is attributed to a specific system, and the human-written fields are ground truth throughout. Each flaw is a localized record with four always-present fields---the \textbf{violated fragment}, a free-text \textbf{rationale}, a \textbf{severity}, and the annotator's \textbf{confidence}---and two optional, a \textbf{time span} and a \textbf{box track}. Two grouping tags---a \emph{modality} tag (seen, heard, or both) and a ten-item \emph{type} tag---are machine-assigned by majority of three zero-temperature model votes and never treated as ground truth. \Cref{sec:app-benchmark} gives per-field coverage, the per-type breakdown, and timing distributions.

Three recorded stages (\cref{fig:benchcomp} in \cref{sec:app-benchmark}) narrow the $1{,}500$ clips and $2{,}582$ flaws to the evaluation set: \emph{clip selection} keeps retrievable, dynamic clips; \emph{flaw selection} keeps only ``seen''- or ``both''-tagged flaws a pixel-only critic could find, leaving $1{,}107$. A deterministic seeded procedure, stratified by severity, flaw count, and duration, draws \emph{the frozen core}---$150$ clips and $306$ flaws (a $29$-clip continuity stratum exempt from replacement); a separate pool of $711$ clips is held disjoint from both frozen sets for learning. The core is an optimization set, so results on it measure fit to that core. One clip produced no measurement, so recall is reported over $149$ clips and their $304$ flaw entries, never the full $306$---the two excluded rather than counted as missed, making the reported rate very slightly optimistic.

Unlike clip-level resources\label{sec:bench-vs}---Physion-Eval~\cite{physioneval2026}, VideoPhy-2~\cite{videophy2_2025}, and per-dimension suites such as VBench~\cite{vbench2024} and T2V-CompBench~\cite{t2vcompbench2024}---our corpus ties each failure to the exact prompt words with an optional time span and box track, so an allegation can be matched to a particular annotated failure rather than only correlated with a clip-level score. This localization does not improve reliability: our corpus is single-annotator and contains only flawed clips, whereas VideoPhy-2 provides redundant judgments and majority resolution.
\subsubsection{Reference standard and matching protocol}\label{sec:eval-protocol}
The evaluation core is the reference standard: its $149$ clips carry $304$ annotated flaws---the denominator for all core recall analyses. Because every clip carries at least one flaw, recall is directly measurable but precision is not (a detector flagging everything scores perfect recall), so we report recall alongside the rate at which accusations land on a real flaw. Recall counts flaw entries, not distinct spans, since matching claims each entry at most once. Matching\label{sec:matching}, run after critic inference, cannot affect the critic's outputs. The comparison unit is a sentence pair (annotator vs.\ critic), judged by meaning as a \emph{whole} match (same defect), a \emph{partial} match (critic identifies at least one aspect of a compound annotated defect), or a non-match. Whole and partial pairings are candidate edges for maximum bipartite matching, each finding and flaw used at most once; every flaw is then reported as whole, partial, or missed. The matching judge, distinct from all evaluated components, votes at temperature zero with a retry-and-refallback rule and aborts rather than reporting a number if more than $2\%$ of flaws yield no usable outcome.
\subsection{The physics-guided generator}\label{sec:gen-experiments}

Three questions organize the generator experiments: whether the geometric control overrides
prompt-implied motion (\cref{sec:gen-prompt}), when a conditioned trajectory becomes insensitive to
control withdrawal (\cref{sec:gen-release}), and how the pipeline behaves beyond single-body probes
(\cref{sec:gen-breadth}). The first and third use the deployed generator; the release analysis is a
separate mechanism probe on the second backbone, its withdrawal operation defined in
\cref{sec:gen-withdrawal}.

\subsubsection{Setup: from simulation to a controllable clip}\label{sec:gen-setup}

The deployed backbone is Wan~2.2-VACE-Fun-14B (\code{WanVACEPipeline}, fixed negative prompt, no
appearance reference, deterministic sampling from the recorded seed). Scenes are simulated in MuJoCo
and rendered to $81$ frames at $16$\,fps; as the simulated duration varies, this render clock
stretches playback by $1.8\times$--$2.3\times$, so trajectory shape and event order come from the
simulation but absolute speed does not equal simulator time. Generation is staged---controls
rendered first, a 1.3B checkpoint screening each scene once, visibly degenerate outputs not queued
for the 14B model (a pipeline rule, not a generator property). Surviving scenes are generated in
bf16 at $848\times480$ with UniPC at $50$ steps and guidance $5.0$, conditioned on the control video
(globally normalized MuJoCo depth, or a silhouette for predeclared flat-scene templates), an
all-white mask, and the prompt; the conditioning scale is $s=1.0$ deployed, examined separately at
$\{0.5,0.8,1.0\}$.

\subsubsection{Control fidelity under prompt variation}\label{sec:gen-prompt}

We fix the geometric control---one projectile-depth video---and vary only the motion clause across three prompts sharing subject, scene, camera, material and style: a strong horizontal traverse, a neutral one, or one adversarial to the control (drops nearly straight down, almost no sideways motion). Each runs under three regimes---no control, control over the first $10$ of $50$ steps, control throughout---at three seeds (partial), two (unconditioned), one (full). We track horizontal travel (fraction of frame width) in the finished clip; metric and decision rules were registered before generation.

\begin{figure}[t]
  \centering
  \includegraphics[width=\linewidth]{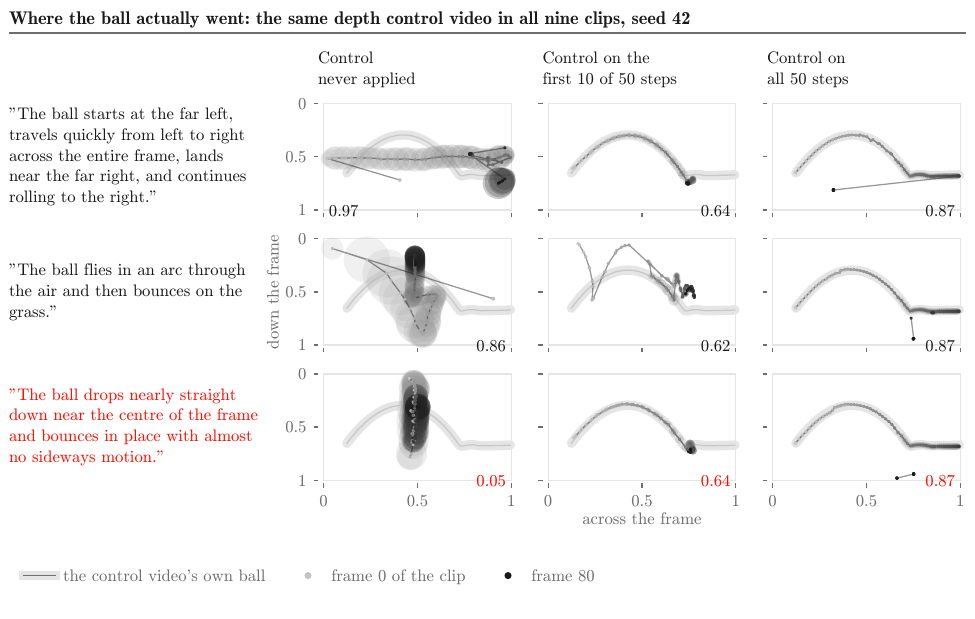}
  \caption{Tracked ball path in nine clips at one seed, against the same depth control throughout.
  Each box is a single $848\times480$ frame, $x$ left to right and $y$ downward; each dot is one
  frame, sized by the tracked region's area and shaded from first frame to last. The pale band is
  the control video's ball, in frame for $63$ of its $81$ frames; the number in each box is that
  clip's sideways travel. The three prompts differ only in the closing sentence beside each row;
  the red row asks for the motion the control does not perform. Without the control the path
  follows the clause and the framing shifts with it; with the control throughout, all three
  clauses reproduce its arc; with the control over the first ten steps the path holds the arc and
  departs at the landing.}
  \label{fig:prompttraj}
\end{figure}

The two inputs cross over (\cref{fig:prompttraj}). Without a control the prompt sets the motion: the clause-attributable range is $0.875$ against $0.252$ between seeds. The unconditioned clips also differ in framing---median tracked ball area $8{,}846$ pixels against $211$ under the control---so travel-fraction is not comparable across regimes, and within-regime comparisons carry the argument. With the control throughout, the clause-attributable range falls to $0.001$; in between, with the control over the first fifth of sampling then withdrawn, the two sources swap: clause range $0.060$ against a seed range of $0.427$.

The trajectories locate the shortfall (\cref{fig:prompttraj}). A clip conditioned over the first ten steps tracks the control's arc through the flight and departs only at the landing, stopping near $0.77$ of the frame while the control's ball rolls on to $0.99$: the lost travel is the post-contact roll, not a different flight---the same phase \cref{sec:gen-release} finds most sensitive to early withdrawal. Of the three clauses at this seed, two follow the control from the first frames and the neutral one does not, staying a mean $0.243$ of the frame from the control's path over the first ten frames. The control-set trajectory thus persists through the text-only steps.
\subsubsection{Trajectory commitment under control withdrawal}\label{sec:gen-release}

Moving the withdrawal point locates the noise level at which the simulated trajectory becomes fixed. The withdrawal operation, its second backbone, $\sigma_{\mathrm{rel}}$, and the two variants clearing or keeping the solver's cached derivatives are defined in \cref{sec:gen-withdrawal} (\cref{eq:flow-sigma}). Scenes are one moving rigid body with a closed-form trajectory (one of three yields only five post-contact frames, so its comparisons rest on two), one schedule setting per seed. Since noise falls during sampling, a \emph{high} $\sigma_{\mathrm{rel}}$ withdraws \emph{early}, leaving later steps unconditioned; a clip still following the simulation afterward was fixed before removal, and the earliest such withdrawal marks the commitment point. Three runs probe seed variability ($16$ seeds, one point), transition shape (six points, one seed), and measurement dependence (a denser sweep re-scored with an independent tracker).

\paragraph{Metrics.} Each run pairs against a $\sigma_{\mathrm{rel}}=0$ run sharing simulation, control, prompt, sampler, and seed, both tracked frame by frame (by colour, except the third run's segmentation tracker). \emph{Retention}, the per-axis Pearson correlation between generated and control paths (\cref{sec:app-retention}), is scored separately over \emph{airborne} and \emph{after-contact} windows split at the ground-contact frame $t_c$; near $1$ means the object moved as simulated, and the split localises disagreements a whole-clip value hides. Two prefixed failure counts accompany it: \emph{whole-clip disagreement} (full-clip horizontal $<0.7$) and \emph{late-window disagreement} (whole-clip horizontal $<0.7$ while airborne horizontal $>0.9$).

\paragraph{Airborne and after-contact agreement under an early release.} The first run fixes release at $\sigma = 0.952$ over two scenes, $16$ seeds each, both variants, against the never-released baseline: $96$ clips, $16$ per cell. Releasing at $\sigma = 0.952$ leaves the thrown ball's airborne flight intact---per-seed horizontal agreement drops $0.003$ versus never releasing ($95\%$ CI $0.001$ to $0.004$, $n=16$ paired seeds)---while whole-clip horizontal falls $0.13$ ($95\%$ CI $-0.02$ to $0.28$) and vertical $0.18$ ($95\%$ CI $0.07$ to $0.29$). The disturbance is confined to post-contact motion, where horizontal runs \emph{opposite} to the simulation on $9$ of $16$ seeds at $\sigma = 0.952$ versus $0$ of $16$ when never released. Whole-clip disagreements number $3$ of $16$ against $1$ of $16$, late-window $2$ of $16$ against $0$ of $16$.

\begin{figure}[t]
  \centering
  \includegraphics[width=\linewidth]{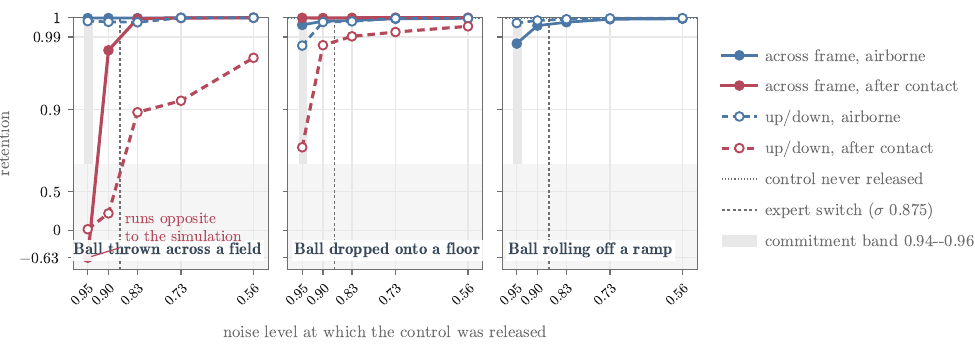}
  \caption{The release-point sweep: one clip per point, three scenes (including a
  ball rolling off a ramp), scored separately over the airborne window and the
  window after the simulated object first reaches the ground. The horizontal axis
  is the noise level at which the control was switched off, high noise on the left.
  Colour marks the window (blue airborne, rose after contact) and line style the axis
  (solid across-frame, dashed up/down); retention is the correlation between the generated object's tracked path
  and the simulation's, defined in the text. The thrown ball's after-contact
  agreement falls below the threshold in both directions at the earliest withdrawal,
  and the vertical component is the later of the two to recover. Guides mark the
  commitment band and the backbone's expert switch; the grey band marks the region
  below the $0.7$ threshold fixed before scoring.}
  \label{fig:injsweep}
\end{figure}

Sweeping six points on one seed across three scenes traces retention against noise level (\cref{fig:injsweep}). For the thrown ball, $\sigma=0.952$ gives after-contact horizontal $-0.63$ (opposite to the simulation) with airborne still $1.00$; $\sigma=0.903$ or later restores after-contact to $0.98{+}$ through the latest release, $\sigma=0.555$. A dropped ball stays above threshold everywhere: after-contact horizontal is $1.00$ throughout, and vertical---the informative axis---is lowest at $0.78$ ($\sigma=0.952$), rising to $0.99{+}$ from $\sigma=0.903$. A ball off a ramp stays at $0.99{+}$ in both axes. A third run (three thrown-ball points, three seeds, two guidance settings, $18$ clips, segmentation tracker) reproduces the direction: airborne horizontal $1.00$ (mean over $6$ clips at $\sigma=0.952$, $95\%$ CI $0.998$--$1.000$) while after-contact horizontal averages $0.69$ over an interval spanning zero ($-0.04$ to $1.43$, $n=6$), recovering to $0.99$ and $1.00$ at $\sigma=0.903$ and $0.833$; at the latest point the tracker finds no object on $3$ of $6$ clips (low-contrast ball on grass), excluded, widening that interval.

Across runs the trajectory stops responding to control removal in a narrow band, $\sigma \approx 0.94$--$0.96$. Under \eqref{eq:flow-sigma} at $S=40$, $\kappa=5.0$ this sits in the high-noise portion of sampling, consistent with the critical-window account~\cite{phasetrans2024,dynregimes2024,critwindows2024} and above the backbone's expert switch at timestep $0.875\,N$~\cite{wan22dual2026}, so motion is fixed before the handoff; we report noise level not step index because changing $S$ or $\kappa$ moves the step but not the level~\cite{sdedit2022,p2train2022}. The two components do not fix together: on the thrown ball, $\sigma=0.9522$ drops after-contact agreement below threshold in both coordinates ($-0.632$ horizontal, $+0.016$ vertical); one point later ($\sigma=0.9025$) horizontal recovers to $+0.981$ while vertical is still $+0.260$, not reaching $+0.894$ until $\sigma=0.8331$; the dropped ball shows the same ordering, so a schedule set by horizontal alone would be too early. (Whole-clip the ordering reverses---vertical $+0.986$ against horizontal $+0.421$---as the large flight excursion dominates the correlation and hides the rebound.)

The failure is specific: flight is reproduced, post-contact behaviour is not. Contacts are a documented failure mode of video generators~\cite{videophy2024}, traced to a weak contact prior; where the prior is weak the control must stay present to override it, so applied throughout it reproduces the contact on every seed and the reversal appears only under early withdrawal. Clearing the cached derivatives changes the whole-clip result by at most $0.04$ horizontally ($95\%$ CI $-0.04$ to $0.13$, $n=16$), the largest per-seed difference on any window or direction.

\FloatBarrier
\subsubsection{Scene-library breadth and failure boundaries}\label{sec:gen-breadth}

The same authoring, simulation and conditioning path scales beyond the single-body probes, producing $1{,}314$ scenes across $66$ \emph{kinds} (a kind is a template, a scene one parameter setting, simulated and generated once); $591$ of $652$ simulations passed a pre-render inspection of simulator output, not generated clips, judged to show the intended behaviour clearly enough to generate from. \Cref{fig:scenebreadth} shows twelve kinds as generated frames at body counts up to $34$, and \cref{fig:scenepairs} three as depth control and generated clip at matched timestamps.

\begin{figure}[tb]
  \centering
  \includegraphics[width=\linewidth]{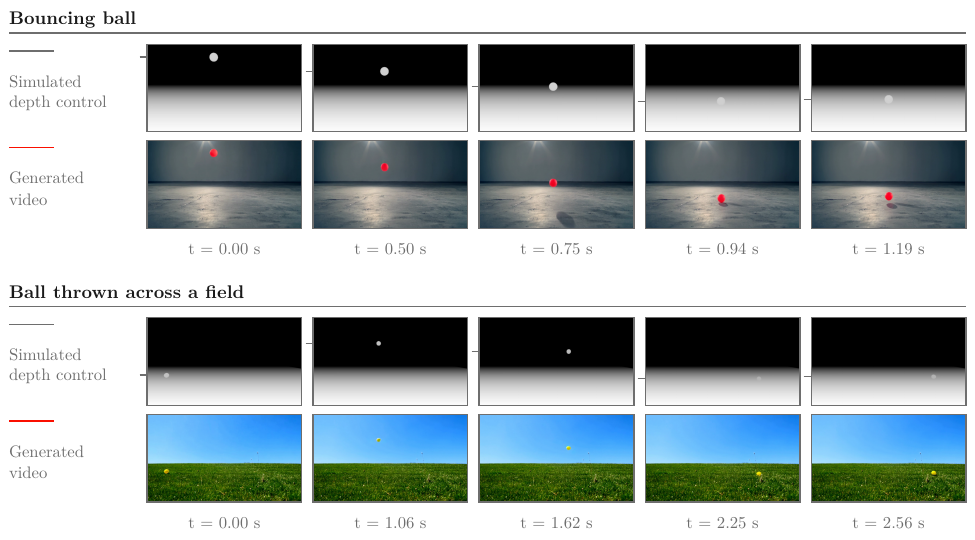}
  \caption{The same mechanism across two scenes with very different requested
  appearance---an indoor studio and an outdoor field---each shown as its depth
  control above and the resulting generated clip below. Timestamps are per scene,
  chosen from the turning points of its tracked vertical
  path, so each column falls within a moving interval. Appearance changes completely with
  the prompt while the motion stays the simulation's: for the thrown ball the
  generated and control paths agree to a correlation of $1.0000$ horizontally
  and $0.9998$ vertically, with a median disagreement of $0.7$ pixels across and
  $1.1$ pixels down, over the $63$ frames in which the ball is measurable in
  both before it leaves the frame.}
  \label{fig:gencases}
\end{figure}

\begin{figure}[tb]
  \centering
  \includegraphics[width=\linewidth]{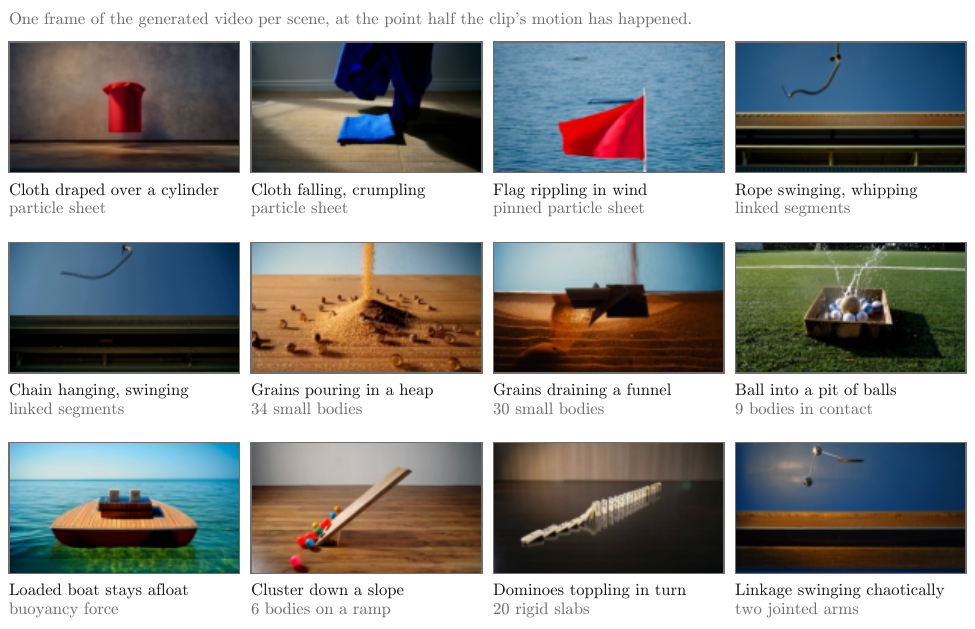}
  \caption{Twelve of the $66$ scene kinds, one generated frame each. Each tile is
  selected by rule: frame-to-frame change is accumulated over the clip and
  the tile is the frame by which half the total has occurred, which falls during the scene's principal motion rather than its start. The second caption line names the physics
  primitive, and every body count in it is the scene's own parameter. Each clip's simulation passed the
  inspection described in the text, and the tiles were additionally inspected manually.}
  \label{fig:scenebreadth}
\end{figure}

\begin{figure}[tb]
  \centering
  \includegraphics[width=\linewidth]{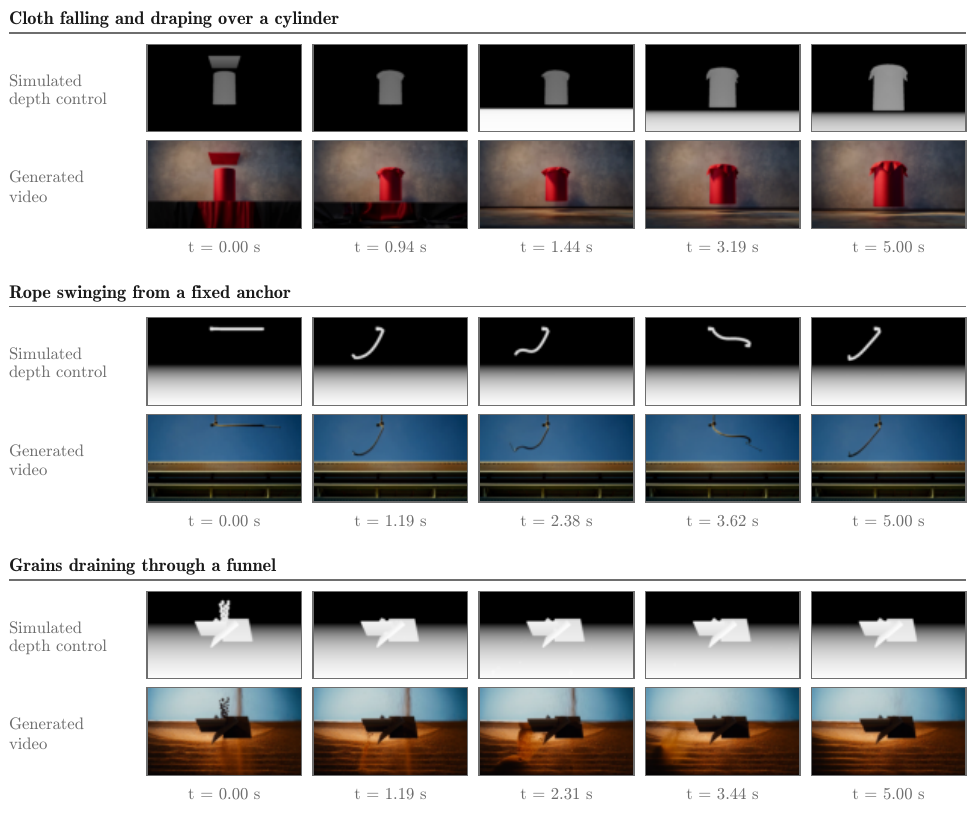}
  \caption{Three deformable and granular scenes, each with simulated depth
  control above and generated clip below, at matched timestamps. Columns are placed at equal steps of the clip's
  accumulated change, so none falls in a temporally static interval. Pairs are shown only for scenes
  whose depth control remains distinguishable from its background in every column;
  an object that settles onto the floor takes the same grey as the floor and leaves
  the control blank.}
  \label{fig:scenepairs}
\end{figure}

Fidelity does not extend uniformly. Assessed by manual inspection with the colour tracker of \cref{sec:gen-release}, single-object scenes reproduce the simulated trajectory throughout (\cref{fig:gencases}), but a multi-object interaction such as a billiards break does not---neither the deployed backbone nor the alternates produced a collision in which every body moved as specified---and its commitment point is unmeasured, the colour tracker being unable to follow a scatter as one path. The same rigid-body engine produces every scene, deformable and granular kinds built from its particle-and-link primitives (springs, linked segments, contacting bodies, buoyancy), so the library excludes fluid, smoke and finite-element soft-body solves.

\subsection{The \VeriPhy{} critic}\label{sec:eval}
The benchmark and metrics are in \cref{sec:dataset}; this subsection reports the critic's recall on the evaluation core, broken down by clip property and specialist trace. How the settings here read together is deferred to \cref{sec:eval-scope}.

\paragraph{Implementation and cost.}
The served Qwen3-VL model (\cref{sec:dataset}) runs with tensor parallelism across two GPUs. Planner and verifier are separate calls: the planner sees the prompt and claim state but no frames, the verifier sees densely decoded, timestamped frames and one declared predicate. Every measurement uses the logged dense-input path with the frozen target, BF16 inference, and pinned tool revisions. Since a detector, a tracker, and a full vision-language judgement differ in cost, cost is reported as model and tool calls per clip rather than wall-clock time, and uncertainty by resampling whole clips, since two flaws in one clip are not independent.

\paragraph{Planner validity.}\label{sec:planner-gaps}
Inspected before end-to-end evaluation without video or tool access, the prompt-only planner produced a valid, compilable plan for all $233$ claims of $22$ annotated clips (one or two calls per clip); $27$ claims ($11.6\%$) carried a specialist action surviving validation, against $2.3\%$ under an earlier tabular format. On diagnostic sets requiring quantitative measurement the routing rate was $19/30$ across counting, spatial, and ordering claims and $10/10$ for synthetic counting prompts---tests of routing and syntax, not verdict accuracy. The supported vocabulary lacks an ``on top of'' predicate and rejects negated measurements; such claims fall back to the generalist verifier.
\subsubsection{Flaw recall on the evaluation core}
\label{sec:headline}
\paragraph{Headline recall.}
Over the $149$ core clips and their $304$ human-annotated flaws, \VeriPhy{} finds \textbf{228
($75.0\%$)} under \cref{sec:matching}---$191$ whole, $37$ partial---and finds every annotated flaw
on $91$ of the $149$ clips. Two components drive matches: the fallback verifier gpt-5.6-sol overrode
a dense-verifier pass to contradiction on $83$ of the $228$, and the on-screen-text recognizer
transcribes malformed glyphs literally rather than normalizing them, so string comparison catches
garbled text.

\paragraph{Versus published evaluators.}\label{sec:external}
We compare \VeriPhy{} against defect-localizing evaluators---question-decomposition and scene-graph methods (per-question answers linkable to claims) and free-text evaluators (unstructured flaw lists)---and test whether CLIPScore tracks aggregate clip properties. Modular video QA is context-only, not evaluated; clip-level agreement with human ratings is deferred to \cref{tab:vs2}.

\begin{wrapfigure}{r}{0.46\textwidth}
  \centering
  \vspace{-10pt}
  \includegraphics[width=0.44\textwidth]{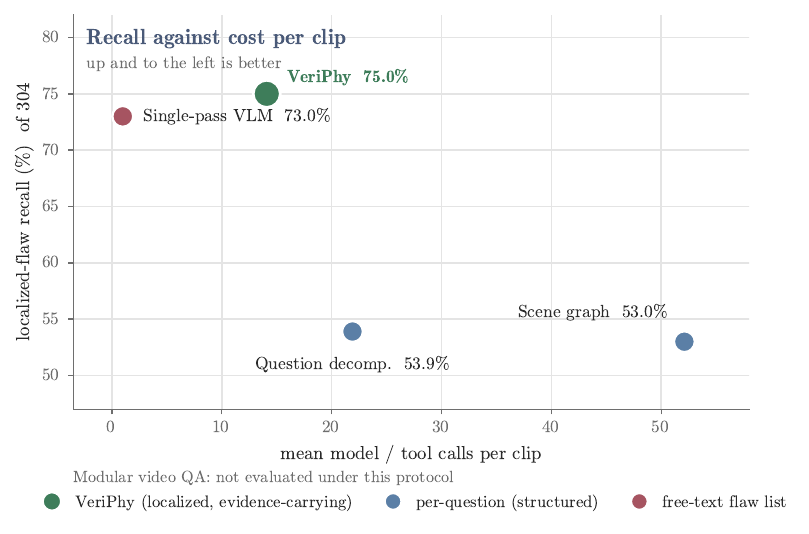}
  \caption{Localized flaw recall vs.\ cost per evaluator, over the same $149$
  clips and $304$ flaws (\cref{sec:matching}). Horizontal axis: mean model and tool calls
  per clip; vertical: recall. Up-and-left is better. Full comparison: \cref{tab:external}
  in \cref{sec:app-external}.}
  \label{fig:external}
  \vspace{-8pt}
\end{wrapfigure}

The same served model---single-pass over the prompt and every frame, no planning, tools, or per-claim verification---matches $222$ of the $304$ flaws at one call per clip; \cref{fig:external} reports it separately, since free-text lists are prompt-sensitive with no per-verdict evidence.

Question decomposition matches \textbf{164 of the $304$ flaws ($53.9\%$)}---$125$ whole, $39$ partial---versus \VeriPhy{}'s $228$, on the same $149$ densely-read clips and model (identical claim lists on $148$). It asks one closed question per claim and accepts the answer, whereas \VeriPhy{} plans measurements and may run several checks per claim. Higher recall is not a higher per-finding match rate: $29\%$ of its $557$ findings match a human flaw versus $26\%$ of \VeriPhy{}'s $871$. About a third of its questions only check object presence (\emph{Is there a vending machine?}), answerable from one frame; twelve clips get no finding. Clip by clip it finds fewer than \VeriPhy{} on $70$, equal on $63$, more on $16$.

As a clip-level diagnostic, CLIPScore correlates across the $149$ clips with severity ($r=+0.035$, $p=0.67$) and flaw count ($r=-0.024$, $p=0.77$); with $86$ clips carrying multiple flaws ($241$ of $304$), a single scalar cannot localize them.
\paragraph{Recall by category and clip property.}\label{sec:bycat}

\begin{table}[t]
\centering\small
\renewcommand{\arraystretch}{1.2}
\begin{tabularx}{\textwidth}{@{}L c c c c c@{}}
\toprule
\textbf{Annotated flaw category} & \textbf{Annotated} & \textbf{Whole}
 & \textbf{Partial} & \textbf{Found} & \textbf{Recall} \\
\midrule
action or event occurs & 137 & 100 & 14 & 114 & 83\% \\
how many & 37 & 10 & 8 & 18 & 49\% \\
object identity & 30 & 14 & 6 & 20 & 67\% \\
on-screen text & 25 & 21 & 0 & 21 & 84\% \\
spatial relation & 18 & 10 & 1 & 11 & 61\% \\
order and timing & 17 & 10 & 3 & 13 & 76\% \\
attribute & 16 & 10 & 2 & 12 & 75\% \\
camera and style & 10 & 7 & 2 & 9 & 90\% \\
other & 7 & 5 & 1 & 6 & 86\% \\
physical motion & 7 & 4 & 0 & 4 & 57\% \\
\midrule
\textbf{All} & \textbf{304} & 191 & 37 & \textbf{228} & \textbf{75.0\%} \\
\bottomrule
\end{tabularx}
\caption{Flaw recall on the evaluation core, by annotated flaw category: $228$ of $304$
human-annotated flaws over $149$ clips. Whole and partial are the match outcomes defined in
\cref{sec:matching}; Found is their sum. Categories are the annotators' labels, assigned
before any critic ran.}
\label{tab:headline}
\end{table}

By the annotators' own categories (\cref{tab:headline}), recall runs from $90\%$ to $49\%$. Camera
and style ($90\%$) and on-screen text ($84\%$) rank highest, both measured directly against an
explicit target; counting is lowest at $49\%$ because its path reduces a per-frame series to its mode
(\cref{sec:contract}), so an intermittently detected subject reads as absent. Physical motion holds
seven flaws, four found; most concern deformation---cloth folding wrongly, a body bending where it
should not---which lies outside the trajectory measurements, and its recall is read descriptively
given the size. Grouping the $304$ flaws instead by severity, duration, flaw count, and difficulty
stratum keeps subgroup recall within about ten points of the overall rate everywhere---severity~$5$
lowest at $18$ of $28$---so neither severity nor stratum clearly predicts whether the critic finds a
defect (\cref{tab:robust} in \cref{sec:app-robust}).

\subsubsection{Specialist measurement traces}
\label{sec:cases}
\begin{figure}[tp]
  \centering
  \includegraphics[width=0.95\linewidth]{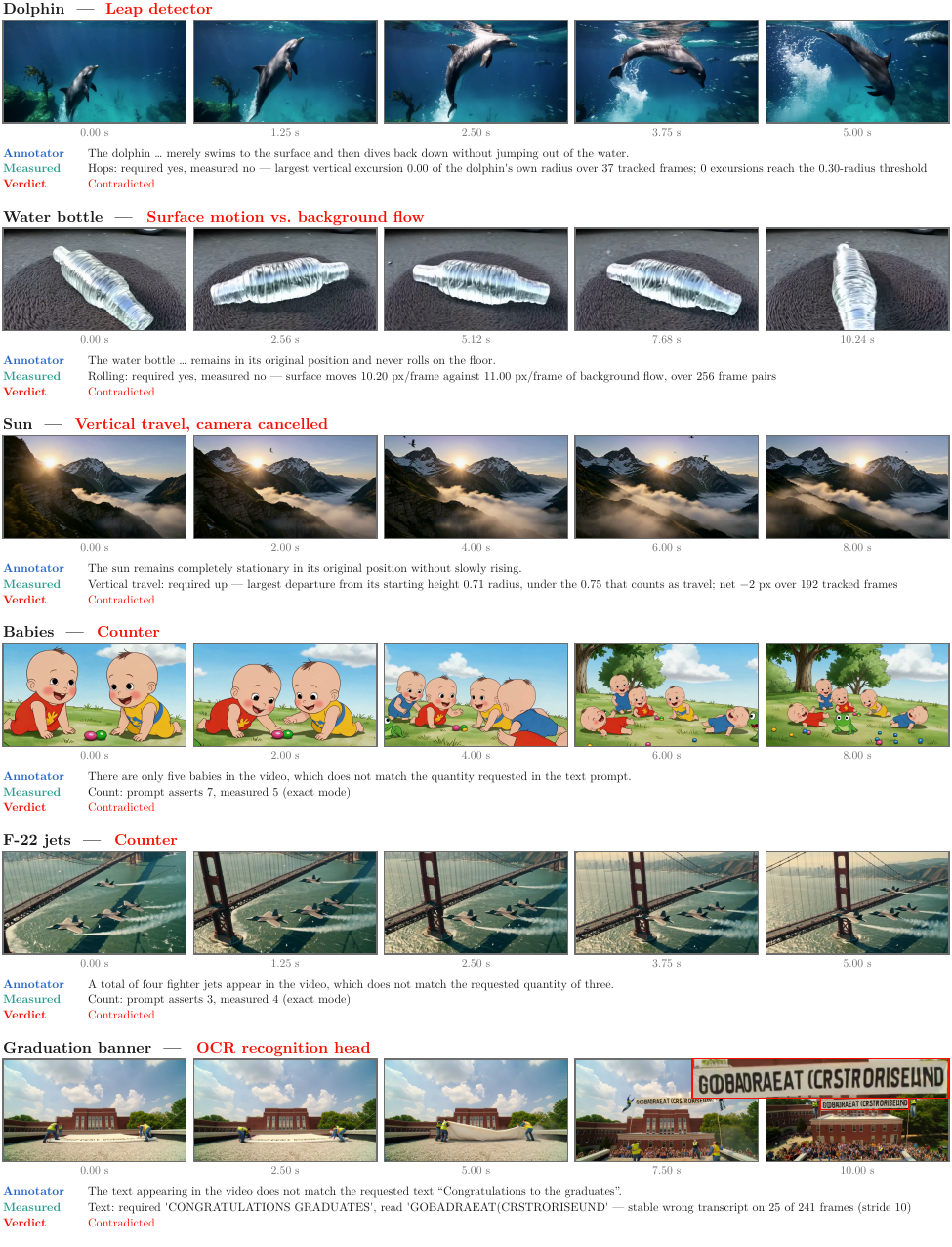}
  \caption{Six flaws identified through specialist measurement, one per row. Each row gives the
  specialist that answered, a five-frame filmstrip sampled uniformly across the clip, the human
  annotator's sentence verbatim, the specialist's own required-versus-measured pair with
  its evidence, and the resulting claim state. Displacements are normalized by the
  subject's radius, so no scene calibration is required. In the text case, the boxed
  banner region of the last frame is magnified in place, showing the render's own glyphs.
  Measurement text is copied from the run's records.}
  \label{fig:criticcases}
\end{figure}

\Cref{fig:criticcases} shows six matched findings from an earlier physics-routing pass over the evaluation core; they are not from the run of \cref{tab:headline} and do not contribute to its recall. Five rest on numeric measurements, the text case on a recognition-head transcript compared as a string. The selection rule was fixed in advance---clips whose every annotated flaw was linked to a finding, and only where the raw per-check record attributed that finding to a specialist measurement and stored the values---excluding ten candidates for mismatched linkage, a verifier-produced finding, or, once, an invalid rationale (a ``vertical 9:16'' aspect ratio misread as a $9.0$-versus-$16.0$-second duration).

\paragraph{What each measurement discriminates.} Each catches what an impression would not
(\cref{fig:criticcases}). The \emph{dolphin} leap is contradicted by a near-zero vertical excursion
a motion-only test would have passed; the \emph{bottle}'s surface and background move alike, so its
rolling is camera motion; the \emph{sun} is scored by peak departure (a rise then fall cancels) and
falls short amid comparable camera motion; \emph{counting} compares asserted against measured counts
exactly; and the \emph{text} head returns garbled glyphs that fail string equality, which a
generative model would silently normalize to the requested phrase.

\paragraph{Independent measurements and coverage.} A second dolphin sentence asserts the animal never
disappears, so two specialists evaluated the prompt's requirement that it disappear: the
vertical-travel check required downward motion but measured $2.07$ radii \emph{upward} ($-317$\,px
over $37$ frames, negative denoting up), and the persistence check required it to vanish but found it
on $121$ of $121$ frames with no gap of three or more; both independently contradict the
requirement. Each measurement also records its temporal coverage: in the banner case the recognition
head read $25$ of the clip's $241$ frames at stride ten, and a transcript stable across those $25$
evidences a persistent banner, whereas a briefly-appearing caption could fall between samples.

\paragraph{Specialist abstention.} In the run of \cref{tab:headline}, physical specialists were
called on $227$ checks, abstaining on $166$ and contradicting on $61$, each abstention carrying its
reason. The most common is no candidate grounding passing the semantic guard---shown one frame with
the candidate boxes and asked only whether they are the named subject: for seagulls scattering
behind fighter jets, the ``seagulls'' boxes held the jets and were discarded, yielding \unknownv{}
and separating a failure to find the subject from a violation by it.
\subsubsection{Agreement with human clip-level ratings}
\label{sec:vs2}
Since localized recall and clip-level scalars are not directly comparable, we compare \VeriPhy{} with VideoScore2~\cite{videoscore2_2025} on human clip-level semantic-adherence ratings, both on the $1$--$5$ scale and scored with VideoScore2's published script. The sample is a prespecified fixed-seed draw of $200$ VideoPhy-2~\cite{videophy2_2025} test clips ($55$ \emph{hard}-flagged) to which the scoring function was not fitted; agreement uses the four \cref{tab:vs2} columns, dropping per metric any unparsed prediction.

\begin{table}[t]
\centering\small
\renewcommand{\arraystretch}{1.2}
\begin{tabularx}{\textwidth}{@{}L c c c c c c@{}}
\toprule
 & \multicolumn{4}{c}{\textbf{All clips ($n=200$)}} & \multicolumn{2}{c}{\textbf{Hard subset ($n=55$)}} \\
\cmidrule(lr){2-5}\cmidrule(lr){6-7}
\textbf{Judge} & \textbf{Exact} & \textbf{Within 1} & \textbf{PLCC} & \textbf{SRCC}
 & \textbf{PLCC} & \textbf{SRCC} \\
\midrule
VideoScore2~\cite{videoscore2_2025} & \textbf{33.0\%} & \textbf{83.0\%} & 0.353 & 0.327 & 0.239 & 0.151 \\
\VeriPhy{} & 31.0\% & 76.0\% & \textbf{0.464} & \textbf{0.465} & \textbf{0.480} & \textbf{0.460} \\
\bottomrule
\end{tabularx}
\caption{Semantic-adherence agreement with human ratings on the $200$-clip VideoPhy-2
sample described in the text, scored with VideoScore2's evaluation script. The
physical-commonsense dimension is excluded because the critic does not emit a score on
that scale.}
\label{tab:vs2}
\end{table}

The critic trails on exact and within-one agreement but leads on PLCC and SRCC (\cref{tab:vs2}). A paired bootstrap gives SRCC gain $+0.138$ ($95\%$ CI $[+0.005,+0.277]$), PLCC gain $+0.110$ ($[-0.018,+0.242]$); on the $55$ hard clips, $+0.242$ ($[+0.018,+0.511]$) and $+0.309$ ($[+0.050,+0.583]$). This fits per-claim evidence degrading more gracefully than a single learned scalar on hard clips, though that subset is small and intervals wide. Caveats: the single $200$-clip draw describes that sample, not the benchmark; and the judges are not cost-matched (decomposition plus tool calls versus one forward pass), so this says nothing about efficiency.
\subsubsection{Held-out evaluation on Physion-Eval}\label{sec:leaderboard}
Physion-Eval (\cref{sec:bench-vs}) holds out clips from this system's development: $9{,}569$
generated clips from five generators, each with a human report of physical glitches ($8{,}745$
with at least one). Two release limits bound what is measurable: the matched real-world reference
videos for the benchmark's $J$ statistic are withheld for copyright, and glitch-free clips are not
verified negatives. The release thus supports neither $J$ nor false-alarm rates for any
evaluator---only flag rates on release clips and matched-report rates on reported-glitch clips.

\VeriPhy{} runs over a fixed randomized stream; figures are interim, covering the first
$601$ clips ($543$ with a reported glitch, $58$ without). The in-context lesson state
(\cref{sec:icl-exp}) was active and updated as the stream progressed, so the run is prequential over
an evolving configuration, with no component seeing a label before scoring its clip.

Three measures separate prompt-grounded findings from physics-grounded reports (\cref{tab:leaderboard}).
The \emph{matched-report clip rate} is the share of reported-glitch clips with a finding matching the
report under \cref{sec:matching}. The \emph{evidence-conditioned flag rate} is the share the served
model calls physically unrealistic from the check records alone---question, verdict, reason, and
numerical evidence---without frames or the reported defect. The \emph{anomaly-phrased finding rate}
is the share whose finding explicitly alleges a physical anomaly rather than an unmet prompt
requirement.

\begin{figure}[t]
  \centering
  \includegraphics[width=\linewidth]{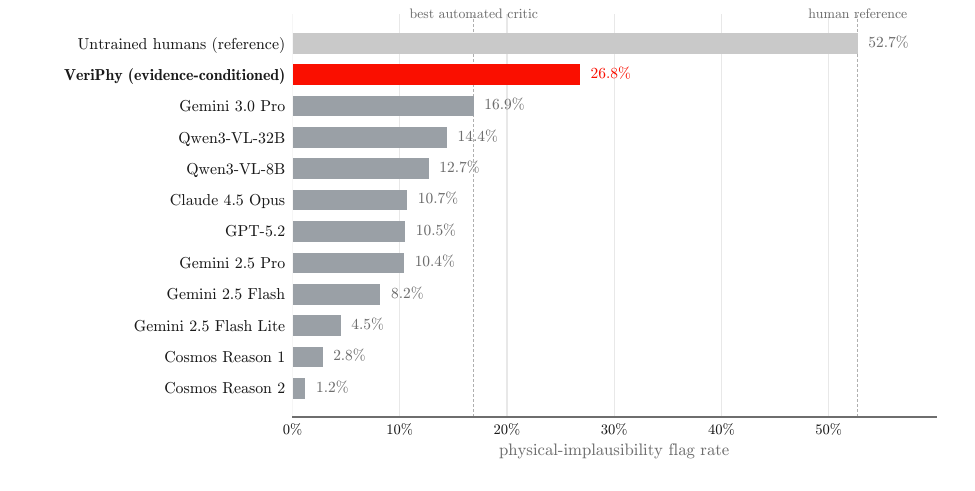}
  \caption{Physion-Eval physical-implausibility flag rate for every evaluator of
  \cref{tab:leaderboard}, sorted high to low, with the untrained-human reference (light) and
  this work's evidence-conditioned rate (red). This work's rate is measured on the first
  $601$ release clips; the benchmark's critic rates are transcribed from its Figure~4 over
  its full evaluation set, so the comparison is descriptive rather than common-protocol.}
  \label{fig:leaderboard}
\end{figure}

\Cref{fig:leaderboard} places this work's evidence-conditioned flag rate against the
benchmark's published flag rate for every critic it evaluated, with the untrained-human
reference for scale; the full table, adding this work's other two rates and the
benchmark's $J$ statistics, is \cref{tab:leaderboard} in \cref{sec:app-leaderboard}. Because this
work's rows and the benchmark's rows use different samples and system configurations, their
juxtaposition is descriptive rather than a common-protocol comparison.

The evidence-conditioned flag rate is nearly twice the anomaly-phrased finding rate,
indicating that the accumulated evidence supports physical-implausibility
judgements beyond what individual findings' wording states.
\subsection{Context-training self-evolution of the critic}
\label{sec:icl-exp}
The inward channel distills lessons from misses into the critic's readable state, improving without touching a weight; the formalism---distillation map \eqref{eq:distill}, acceptance gate \eqref{eq:elitism}, edited state $S=(\mathcal{O},M,K)$---is in \cref{sec:framework}. We measure flaw recall.

On a plan's miss of a human-annotated flaw, $\operatorname{Distill}$ \eqref{eq:distill} reads the trace and miss and writes a lesson naming a claim property and the action it licenses (e.g.\ turning a singular-entity or exact-quantity requirement into a \emph{separate cardinality check verified across beginning, middle and end}); curation groups candidates into $33$ \emph{families} of equivalent advice, one lesson each, appended to the planner's rubric. Only $M$ changes ($S' = (\mathcal{O}, M', K)$); a family's name labels the defect its lesson concerns.

Implementation departs from the formalism twice. First, all $33$ lessons go on every clip: the selective form (one relevance call per lesson--claim pair) costs $132$ model calls for a four-claim prompt and $495$ for a fifteen-claim one, whereas all-at-once costs no calls and adds $9$\,kB to a planner prompt already $17.8$\,kB long. Second, the acceptance gate \eqref{eq:elitism} is dropped: $U_{\mathrm{val}}$ scores only rules already seen in a run, not one from a fresh miss---diagnostically, on $80$ earlier candidates over a ten-clip set it admitted none ($39$ failed its minimum-evidence criterion, $41$ never triggered). The evaluated intervention is the curated $33$ lessons, without the gate.

\begin{figure}[t]
  \centering
  \begin{minipage}[c]{0.47\textwidth}
    \centering
    \includegraphics[width=\linewidth]{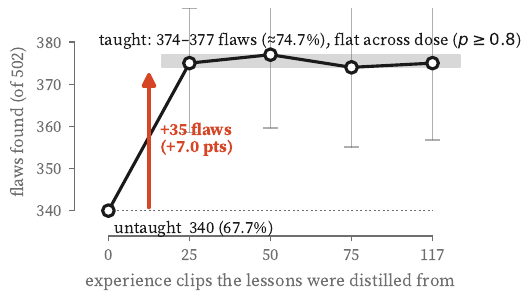}
  \end{minipage}\hfill
  \begin{minipage}[c]{0.49\textwidth}
    \caption{Flaws found against the number of experience clips the lessons were distilled from,
    over the same $502$ held-out flaws in every condition. The dotted line marks the untaught
    condition, $340$ of $502$ ($67.7\%$).
    Whiskers are $95\%$ intervals on the change from untaught, in flaws, computed from the flaws that
    switch status, since the conditions score the same flaws. Every taught condition exceeds the
    untaught one, while pairwise comparisons among the taught conditions return $p \geq 0.8$.}
    \label{fig:iclaggregate}
  \end{minipage}
\end{figure}

\paragraph{Protocol and result.} Lessons distilled from experience clips are tested on held-out ones: excluding every clip sharing an identifier or restating an experience-clip prompt leaves $306$ clips with $502$ flaws, so no lesson is evaluated on the prompt it was written from. Every condition scores those flaws under \cref{sec:matching}'s protocol (a flaw is \emph{found} when the critic covers it wholly or in part), untaught loading an empty lesson file through the same path; five nested conditions---no lessons, then $8$, $13$, $19$ and $33$ lessons from $25$, $50$, $75$ and $117$ clips, each a prefix of the next---form a dose curve. With all $33$ the planner finds $\mathbf{375}$ of $502$ flaws ($\mathbf{74.7\%}$) against $340$ untaught ($67.7\%$): $62$ switch missed-to-found and $27$ back, net $+35$ at $p = 3\!\times\!10^{-4}$ (paired sign test). More experience does not extend it (\cref{fig:iclaggregate}): the four taught conditions find $375$, $377$, $374$ and $375$, each $+34$ to $+37$ over untaught at $p < 10^{-3}$ but no two differing by more than $3$ ($p \geq 0.8$ pairwise), and the unfiltered set ($363$ clips, $626$ flaws) matches. Later lessons reshuffle which flaws are found, not how many---$36$ move each way between the $8$- and $33$-lesson conditions---so the gain is present already with lessons from $25$ clips.

\paragraph{Effect by flaw category.} Both conditions score the same flaws, so we report the paired change and its interval from flaws that switched status (\cref{tab:iclcategory}). Between untaught and $33$-lesson conditions, two categories' $95\%$ interval excludes zero: \emph{action or event occurs}, largest at $211$ flaws, gains $17$ ($8.1$ percentage points, $[+2, +12]$); \emph{on-screen text} gains $8$ of $73$ ($11.0$ points, $[+2, +15]$). All other intervals contain zero, and reading smaller movements as rates overstates them: $+7$ points for camera and style is one flaw of $14$, $+16$ for counting is three of $19$. Flaws rise in both severity grades holding the bulk of the set---$19$ of $288$ at severity~$3$ ($p = 0.01$) and $13$ of $163$ at severity~$4$ ($p = 0.03$)---the remaining $51$ at grades $1$, $2$ and $5$.

\begin{table}[t]
\centering\small
\renewcommand{\arraystretch}{1.2}
\begin{tabularx}{\textwidth}{@{}L c c c c c c c c c@{}}
\toprule
 & & \multicolumn{5}{c}{\textbf{Flaws found, with $\ell$ lessons}} & & & \\
\cmidrule(lr){3-7}
\textbf{Flaw category} & \textbf{Flaws} & $\ell{=}0$ & $8$ & $13$ & $19$ & $33$ & \textbf{Change} & \textbf{95\% interval (pts)} & \textbf{Families} \\
\midrule
\textbf{action or event occurs} & \textbf{211} & \textbf{146} & \textbf{169} & \textbf{169} & \textbf{166} & \textbf{163} & \textbf{+17 (+8)} & \textbf{$[+2, +12]$} & \textbf{20} \\
\textbf{on-screen text} & \textbf{73} & \textbf{53} & \textbf{63} & \textbf{56} & \textbf{62} & \textbf{61} & \textbf{+8 (+11)} & \textbf{$[+2, +15]$} & \textbf{1} \\
attribute & 46 & 29 & 30 & 32 & 31 & 33 & +4 (+9) & $[-2, +12]$ & 2 \\
order and timing & 22 & 11 & 13 & 13 & 13 & 14 & +3 (+14) & $[-9, +27]$ & 3 \\
how many & 19 & 9 & 9 & 11 & 11 & 12 & +3 (+16) & $[-10, +31]$ & 2 \\
object identity & 52 & 39 & 41 & 39 & 42 & 41 & +2 (+4) & $[-3, +7]$ & 4 \\
camera and style & 14 & 11 & 11 & 12 & 11 & 12 & +1 (+7) & $[-4, +7]$ & 0 \\
spatial relation & 45 & 28 & 29 & 30 & 28 & 28 & +0 (+0) & $[-10, +10]$ & 1 \\
physical motion & 15 & 9 & 8 & 12 & 9 & 9 & +0 (+0) & $[-11, +11]$ & 0 \\
\midrule
\textbf{All flaws} & \textbf{502} & \textbf{340} & \textbf{375} & \textbf{377} & \textbf{374} & \textbf{375} & \textbf{+35 (+7)} & \textbf{$[+3, +10]$} & \textbf{33} \\
\bottomrule
\end{tabularx}
\caption{Effect of the distilled lessons on each flaw category, over the same $502$ held-out flaws in every condition. A flaw is \emph{found} when the critic's findings cover it wholly or in part, as in \cref{tab:headline}. The five middle columns count that category's flaws found with no lessons and with $8$, $13$, $19$ and $33$ lessons, distilled from $25$, $50$, $75$ and $117$ experience clips; \textbf{Change} is the untaught-to-$33$ difference in flaws, with percentage points in brackets, since several categories carry few enough flaws that a rate alone would overstate the movement. Because the conditions score the same flaws the comparison is paired: the interval on the change is computed from the flaws whose found status switched, and is given in percentage points of that category's flaws. \textbf{Bold} marks the rows whose interval excludes zero. \textbf{Families} counts, over all $33$ lessons, those whose family name refers to that category, matching on the words in the name. The other category, containing $5$ flaws in all, is omitted from the category rows; the \emph{All flaws} row includes those flaws.}
\label{tab:iclcategory}
\end{table}

\paragraph{Where the lessons point.} Gains are not aligned with lesson names. All eight families from the first $25$ clips concern actions, none mentioning text, counting or spatial relations, yet at $8$ lessons \emph{on-screen text} posts the largest percentage-point gain of any category ($10$ of $73$) before any lesson mentions text, then loses two after one does; the later $25$ families do not move their nominal targets. This fits the first lessons supplying a general instruction---separate a compound requirement into checkable parts---applying well beyond the categories they were written from, which the plan measurements below assess.

\paragraph{Effect of the lessons on the plan.} Lessons compile the prompt into more claims, not more checks per claim: from untaught to $33$ lessons, checks per claim rise only $1.13\!\to\!1.18$ while typed physical claims rise $2{,}905\!\to\!5{,}106$ over $306$ clips ($9.5\!\to\!16.7$ per clip)---as the decomposition reading predicts. This locates the flattening: claims per clip keep rising across all five conditions ($9.5$, $15.1$, $16.0$, $16.1$, $16.7$), so later lessons keep changing the plan while flaws found stall. On-screen text shows this at clip level: text \emph{checks} rise $81\!\to\!136$ (already $114$ at $8$ lessons), yet of $56$ text-flaw clips the number getting any text check does not ($41$ untaught vs.\ $40$ taught)---lessons add checks without widening the set of clips checked.
\subsection{Writing a verdict back into generation}\label{sec:refine}
The critic's outward channel---a verdict with its evidence, fed back as a prompt rewrite that
regenerates the clip---is claimed but not yet measured. We close that loop on a held-out generator
and ask a deliberately narrow question: does it help on the metric that triggered it, and does an
\emph{independent} rater register the same change?

The loop composes the two maps of \eqref{eq:gen-crit-maps} through a rewrite operator: from a base
clip $\hat{\mathcal{V}}\sim\operatorname{Gen}(p,C)$ the critic returns
$\operatorname{Crit}(\hat{\mathcal{V}},p)=(v,\pi,\mathcal{F})$---a verdict
$v\in\{\textsf{plausible},\textsf{implausible},\textsf{abstain}\}$, a plausibility score
$\pi=p_{\text{plausible}}$, and the violated obligations $\mathcal{F}=\{(e_j,a_j)\}_j$ with their
evidence. A text-only rewrite $\rho$ folds those failures into the prompt and
the \emph{same} generator resamples under the same control and seed,
\begin{equation}
  p' = \rho\bigl(p,\mathcal{F}\bigr)\ \ \text{when}\ \ v=\textsf{implausible}\ \ (\text{else}\ p'=p),
  \qquad
  \hat{\mathcal{V}}' \sim \operatorname{Gen}\bigl(p',C\bigr),
  \label{eq:refine-loop}
\end{equation}
so the channel acts only through the prompt---diffusion weights, control $C$, mask, and seed
untouched---changing appearance while fixing the motion path. We run one round, reading the change
in $\pi$ and $v$ against an independent rater applied to both.

\paragraph{Protocol.} We run the full \emph{evaluate\,$\to$\,rewrite\,$\to$\,regenerate} loop over
Cosmos3-Nano, a generator unused anywhere in this system's development, on its $600$ VideoPhy-2 test
clips ($589$ complete the loop with a paired score, so the before/after columns rest on $600$ and
$589$). Each baseline and its regeneration are scored two ways: by \VeriPhy{}'s own metric
($p_{\text{plausible}}$ and the verdict of \cref{sec:eval}), the quantity the rewrite optimizes; and
by VideoPhy-2's official AutoRater~\cite{videophy2_2025}, an \emph{independent} rater returning
semantic adherence (SA), physical commonsense (PC), and their joint agreement (both $\ge 4$). The
AutoRater is the control---not the loop's objective and sharing no component with the critic.

\begin{figure}[t]
  \centering
  \includegraphics[width=\linewidth]{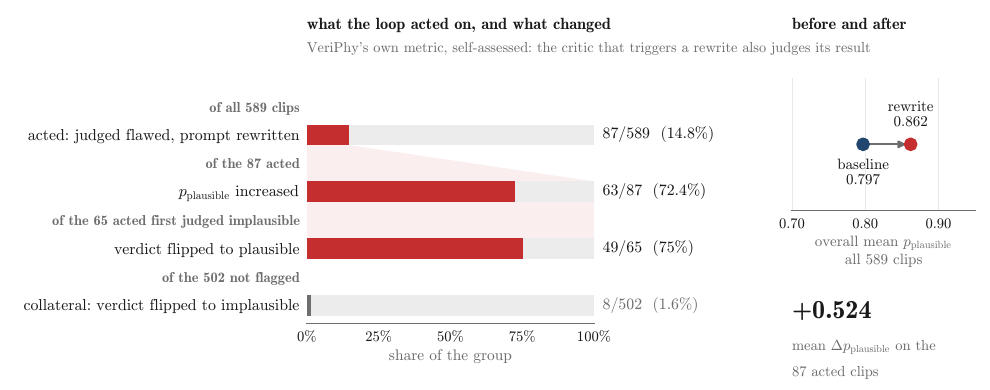}
  \caption{The same loop scored by \VeriPhy{}'s own physical metric. Every quantity
  here is self-assessed: the critic that triggers a rewrite is also the judge of its
  result, so these quantities show the loop is internally consistent---targeted (it acts
  on $87$ of $589$ clips, $14.8\%$), effective on its targets ($63/87$ raise
  $p_{\text{plausible}}$, $49$ of the $65$ acted clips first judged implausible flip to
  plausible, mean $\Delta p_{\text{plausible}}$ $+0.524$), and low-collateral ($8/502$,
  $1.6\%$)---and move the overall mean $p_{\text{plausible}}$ from $0.797$ to $0.862$.
  Each bar is drawn against its own denominator. They are not an independent
  confirmation and should be read against \cref{tab:refine-autorater}, where an
  independent rater registers no corresponding change, including on the physical
  dimension; every value is recomputed from the per-clip records in
  \cref{tab:refine-verify}.}
  \label{fig:refine-selfmetric}
\end{figure}

\paragraph{On its own metric the loop is effective.} By \VeriPhy{}'s own measurement
(\cref{fig:refine-selfmetric}) the loop is targeted with low collateral: it acts on the $14.8\%$ of
clips it judges flawed, raises $p_{\text{plausible}}$ on $72.4\%$ of those and flips $49$ of $65$
implausible verdicts to plausible, regressing only $1.6\%$ of unflagged clips.

\begin{table}[t]
\centering\small
\renewcommand{\arraystretch}{1.2}
\begin{tabularx}{\textwidth}{@{}L cc cc@{}}
\toprule
 & \multicolumn{2}{c}{\textbf{All ($n=589$)}} & \multicolumn{2}{c}{\textbf{Hard subset}} \\
\cmidrule(lr){2-3}\cmidrule(lr){4-5}
\textbf{VideoPhy-2 AutoRater metric} & \textbf{Baseline} & \textbf{Rewrite}
 & \textbf{Baseline} & \textbf{Rewrite} \\
\midrule
Mean semantic adherence (SA)      & 3.12   & 3.13   & 2.75  & 2.76  \\
Mean physical commonsense (PC)    & 3.59   & 3.58   & 3.36  & 3.32  \\
SA $\ge 4$                        & 27.8\% & 28.0\% & 8.9\% & 7.9\% \\
PC $\ge 4$                        & 55.0\% & 54.5\% & 37.2\% & 36.2\% \\
Joint (SA $\ge 4$ \emph{and} PC $\ge 4$) & 22.8\% & 23.1\% & 3.9\% & 3.4\% \\
\bottomrule
\end{tabularx}
\caption{The refinement loop scored by an \emph{independent} rater---VideoPhy-2's
own AutoRater~\cite{videophy2_2025}---on the same $589$ Cosmos3-Nano clips before and
after the loop. No cell moves beyond the rater's noise: joint agreement shifts by
$+0.3$ points over all clips and $-0.5$ on the hard subset, and both underlying
dimensions are flat to slightly lower, including the physical dimension the loop is
meant to improve (mean PC $3.59\!\to\!3.58$ all, $3.36\!\to\!3.32$ hard). On the hard
subset every entry is unchanged or slightly lower. The rewrite therefore produces no
change this independent rater can register, in either the semantic or the physical
dimension. Hard-flagged clips follow VideoPhy-2's designation.}
\label{tab:refine-autorater}
\end{table}

\begin{figure}[t]
  \centering
  \includegraphics[width=\linewidth]{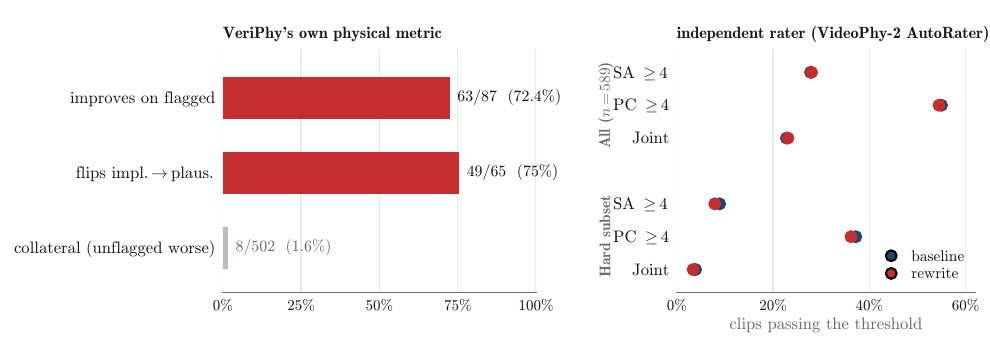}
  \caption{The refinement loop read two ways, on the same $589$ Cosmos3-Nano clips.
  \textbf{Left:} by \VeriPhy{}'s own physical metric---the quantity the rewrite
  optimizes---the loop is targeted and low-collateral: of the clips it flagged, $63/87$
  improve and $49/65$ flip from implausible to plausible, against $8/502$ unflagged
  clips that regress. \textbf{Right:} the same clips scored by VideoPhy-2's independent
  AutoRater show no movement---each dumbbell from baseline (navy) to rewrite (red) collapses
  to a point at every threshold, on both the full set and the hard subset, in the semantic
  \emph{and} the physical dimension. Every number is transcribed from
  \cref{tab:refine-autorater} and the self-assessed counts of \cref{fig:refine-selfmetric}. The improvement is thus
  visible on the critic's own metric and not on the one rater independent of it. Clips in
  \cref{fig:refinefilm}.}
  \label{fig:refinedata}
\end{figure}

\begin{figure*}[tbp]
  \centering
  \includegraphics[width=\linewidth]{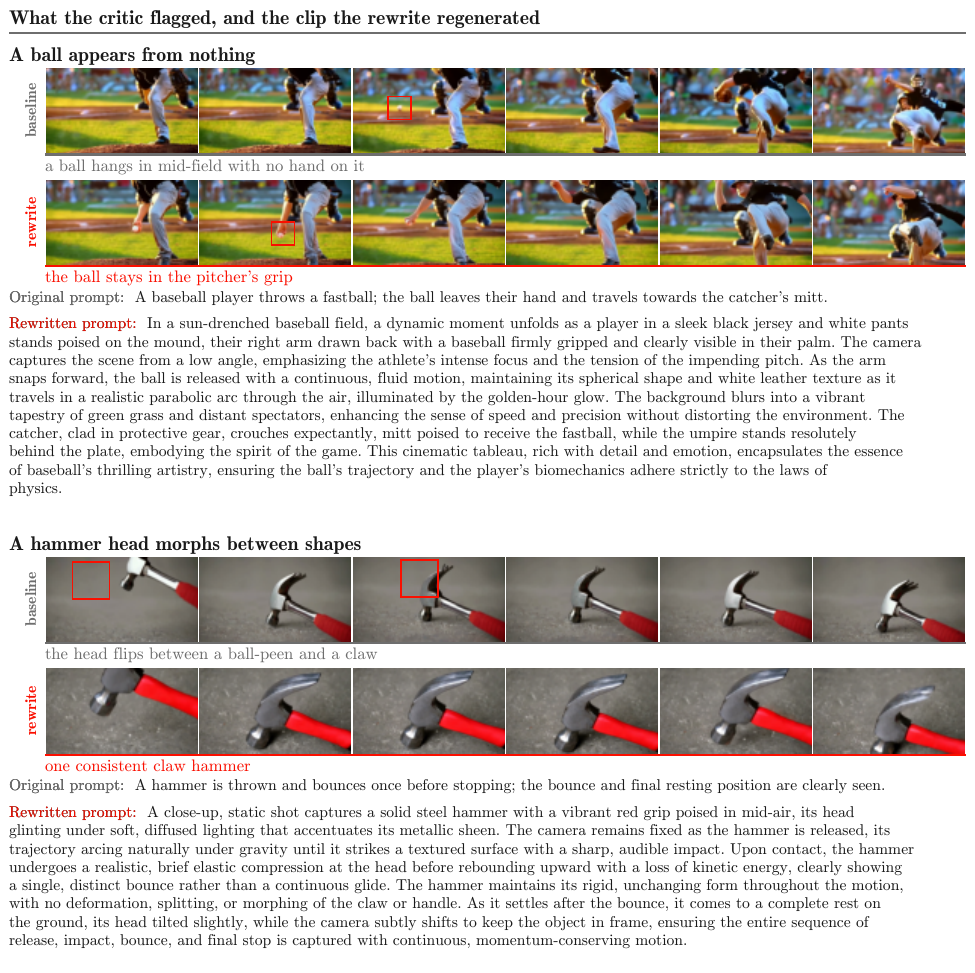}
  \caption{What the critic flagged, and the clip the rewrite regenerated---clips
  \VeriPhy{} judged implausible and resampled through \eqref{eq:refine-loop}. Each case
  is two real filmstrips: the \emph{baseline} clip (grey) and the clip regenerated from
  the rewritten prompt (\emph{rewrite}, red), six frames early to late; red boxes mark
  the cited physical violation and its repair, and the full regenerated prompt is printed
  verbatim beneath. Cases were chosen by an independent frame-by-frame review under an
  honesty constraint---the before/after must be the \emph{same} scene, not a rewrite that
  swaps the scene or introduces a new artifact---which most flagged clips fail; these are
  cases where the fix is visible and the comparison fair. Two further cases are in
  \cref{fig:refinefilm2}; the quantitative reading is \cref{fig:refinedata}.}
  \label{fig:refinefilm}
\end{figure*}

\begin{figure*}[tbp]
  \centering
  \includegraphics[width=\linewidth, trim=0 430 0 0, clip]{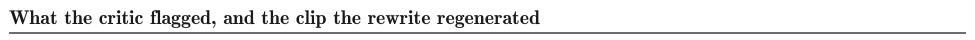}  \par\vspace{2pt}  \includegraphics[width=\linewidth, trim=0 0 0 238, clip]{fig_refine_film2.pdf}  \par\vspace{4pt}  \includegraphics[width=\linewidth, trim=0 0 0 20, clip]{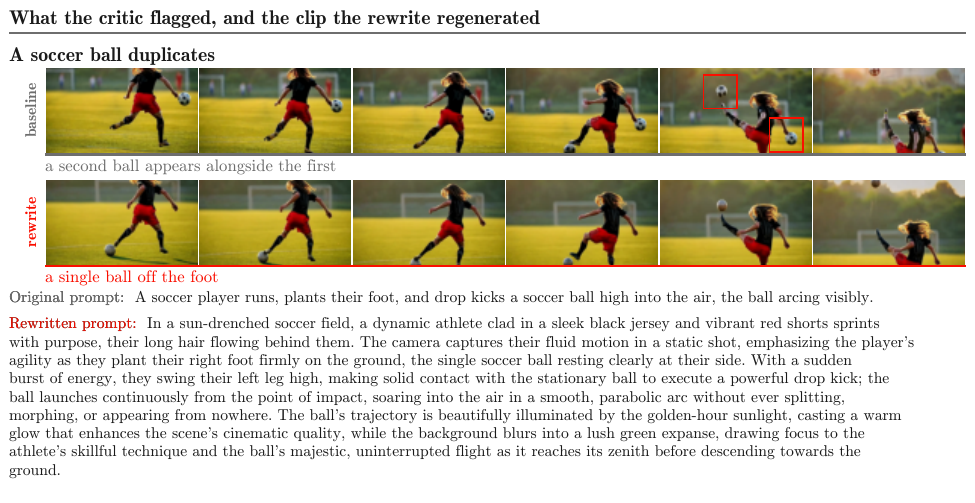}
  \caption{Critic-guided rewrite, continued, in the layout of \cref{fig:refinefilm}: a
  frisbee that warps into a red ball mid-flight, held to a single rigid object by the
  rewrite, and a soccer ball that duplicates during the kick---two balls in the baseline
  (boxed), one in the rewrite. Red boxes mark the flagged defect and its repair; the full
  rewritten prompt is printed verbatim beneath each.}
  \label{fig:refinefilm2}
\end{figure*}

\paragraph{An independent rater registers no change.} The same $589$ clips scored by VideoPhy-2's
AutoRater show no movement in either dimension (\cref{tab:refine-autorater}; \cref{fig:refinedata}
places the two readings side by side, clips in \cref{fig:refinefilm}): joint agreement, SA, and---critically---the \emph{physical} PC dimension all stay flat to lower, within rater noise. Stated
together, not collapsed: the gain appears only on the metric the critic itself assigns, while the
one rater independent of the loop registers no change on either axis. The design cannot separate two
readings---that rewrites improve a physical quality the critic can measure but the AutoRater cannot,
or that they raise the critic's own score without improving the clip---because the critic both
triggers and judges the effect it reports. (External position only: Cosmos3-Nano's AutoRater joint
score is unchanged by the rewrite, $22.8\!\to\!23.1\%$ (all) and $3.9\!\to\!3.4\%$ (hard), near
published human-rated leaderboard entries; but human and AutoRater joint scores share no common
protocol, so this is descriptive only.) Every reported figure was recomputed independently from the
run's per-clip files and reproduces exactly (\cref{tab:refine-verify} in \cref{sec:app-refine-verify}).

\paragraph{Scope.} Confirming the rewrites improve physical fidelity as an independent judge would
score it needs a rater both independent of the critic and sensitive to the targeted physical
property---which the AutoRater's flat PC suggests it is not on these clips. All $589$ per-clip
comparisons and both scoring streams ship with the run; a curated set of before/after clips,
with the critique and rewritten prompt for each, is on the project page.
\subsection{Scope and interpretation}\label{sec:eval-scope}
Each setting must be read on its own terms. The generator analyses (\cref{sec:gen-experiments}) are
pre-registered probes on small closed-form scenes, characterizing the deployed pipeline rather than
its generality. Core recall uses recorded annotations on development data, where an unrecorded defect
cannot be scored, and its question-decomposition baseline was specified after inspecting critic
outputs. The context-training result (\cref{sec:icl-exp}) is held out from the distillation pool but
still drawn from the development corpus. Only the interim, prequential Physion-Eval run
(\cref{sec:leaderboard}) is held out from development entirely---with an evolving lesson state, a
different annotation vocabulary, and no frozen-core protocol. The refinement loop
(\cref{sec:refine}) is closed on a held-out generator but confirmed only on the critic's own metric;
the one independent rater shows no change in either dimension.

\section{Conclusion}\label{sec:conclusion}
A quality score can rank two clips; it cannot identify which physical obligation was
violated, or when. Work on physical faithfulness in generated video divides into three
separate pieces: a way to \emph{steer} what gets generated, a way to \emph{measure}
whether obligations are met, and a source of \emph{ground truth} to score those
measurements against. Closing the gap takes all three at once. This report builds one
artifact for each and connects them into a single loop: a simulation-driven
controllable generator (MuJoCo\,$\rightarrow$\,Wan~2.2-VACE) in which the solver fixes
the motion and the prompt supplies only appearance; a benchmark of human-annotated
defects localized to the offending prompt words, time span, and box track; and a
critic, \VeriPhy{}, that compiles the prompt into typed physical obligations
\emph{before} it reads any frame, then calls frozen experts only within the scope
those obligations declare.

That ordering is the central design decision. Because the plan is fixed before any
pixel is read, every measurement traces back to the obligation that asked for it, and
every verdict (\supported{}/\contradicted{}/\unknownv{}, surfaced as \emph{plausible},
\emph{implausible}, or \abstain{}) carries the evidence that forced it. On a
development core of $149$ clips carrying $304$ human-marked defects, \VeriPhy{}
accounts for $228$, against $164$ for a published question-decomposition evaluator
reading the same clips with the same served model. Prompting that backbone
monolithically reaches $222$, so recall alone does not separate the two. What
separates them is that a reader can open any one of \VeriPhy{}'s $228$ findings and
see the measurement that produced it, the time window it was taken over, and the rule
that turned it into a verdict. The same reader can overturn that verdict by disputing
the evidence. The monolithic prompt returns a score with no supporting evidence.

An auditable state is also improvable without retraining. Distilling lessons from
experience clips and loading them as readable context lifts recall from $340$ of $502$
held-out flaws to $\mathbf{375}$ ($67.7\%\rightarrow\mathbf{74.7\%}$, net $+35$ at
$p=3\!\times\!10^{-4}$), with no lesson evaluated on the prompt it was written from
(\cref{sec:icl-exp}). The gain saturates quickly: lessons distilled from $25$ clips
already find $375$, the count also reached with $117$ clips, and the four taught
conditions differ by no more than $3$ ($375$, $377$, $374$, $375$). Later lessons
change \emph{which} flaws are found rather than how many. A small, legible body of
experience, in the form of text a human can read, audit, and edit, does work otherwise
done by fine-tuning, and it accrues to a state that stays inspectable instead of being
absorbed into model parameters.

The outward channel is less developed, and its evidence is split. Feeding a verdict
and its violated obligations back as a prompt rewrite, then resampling under the same
control and seed, closes the critic-to-generator loop end to end on Cosmos3-Nano, a
generator held out from this system's development. On the metric the rewrite
optimizes, the loop behaves as designed: it fires on $87$ of $589$ clips ($14.8\%$),
improves $63$ of those $87$, flips $49$ of $65$ implausible verdicts to plausible, and
regresses $8$ of the $502$ it leaves alone ($1.6\%$). An independent rater sharing no
component with the critic registers no corresponding movement in either the physical
or the semantic dimension (\cref{sec:refine}). The disagreement is the result: it
localizes the open problem to the rewrite channel itself, which at present can edit
only words. Whether a text-only edit can move a generator's physics, or whether the
correction must instead enter through the geometric control channel this report
builds, is an open question this setup can now measure.

The next direction is one this system is already built for and does not yet use:
\emph{sound}. Audio resolves physical implausibility that vision leaves ambiguous. A
visible collision that is silent, or an impact heard before contact, breaks causality
in a way visual fluency cannot conceal, and it does so on a timeline that can be
measured rather than judged. \VeriPhy{} already carries a sound-event specialist, and
the architecture already places audio intervals and visual contact windows on one
shared event clock, which is what a synchrony claim requires: a contact window from
vision, an onset from audio, and a typed relation between them, with an \abstain{}
when either stream is unusable. This report does not exercise that capacity; every
measured claim here is visual. Extending physical verification across modalities, with
obligations that bind what is seen to what is heard and a benchmark that localizes
audiovisual faults as precisely as this one localizes visual ones, is what we build
next.

\Cref{sec:framework} casts these components as a single state-optimization framework,
and \cref{sec:roadmap} lays out the phased plan that carries them there.

\section{Limitations}\label{sec:limitations}

The primary limitation of this work is that the control signal comes from a simulator, so scene diversity is
bounded by what is simulated, and the mapping from simulation to prompt appearance is not
itself verified for physical consistency. The corpus used here is also recall-only (no clean clips), so precision is
not directly measurable; flaws are single-annotator with no inter-annotator
agreement; the scope/category tags are machine-derived; and the source generator
of the clips is unrecorded. Repeatedly designing against one frozen set risks
overfitting to it even without training on it.

Moreover, the spatial relation predicate and negative-event
verification are limited; binding remains a central failure mode, since the compiler
enforces symbolic identifier consistency but cannot guarantee that heterogeneous
tools localized the same physical instance; and the reasoner-mediated sub-checks
reduce but do not remove semantic opacity.

Four limits bound what
\cref{sec:selfevolve:measured} establishes. The acceptance gate
\eqref{eq:elitism} is not exercised, so the measurement covers distillation
without the admission discipline that is meant to keep a bad lesson out; a rule
written from a fresh miss has no proxy history, and closing that gap needs a
$U_{\mathrm{val}}$ defined on unrun rules. The taught planner reads $1.76$ times
as many claims out of a prompt as the untaught one, and flaw recall rewards
asking more, so the effect combines a better-targeted plan with a larger one; a
run holding the taught arm to the untaught arm's question budget would separate
them. The lessons are induced once from a fixed pool rather than accumulated online, so the curve of
\cref{fig:iclaggregate} measures the effect of the amount of experience distilled, not of
accumulating it during deployment; the shape beyond $117$
experience clips is unmeasured. Finally, the
$502$-flaw held-out set is a different population from the $304$-flaw evaluation
core used everywhere else in this paper, and the two rates should not be compared
directly. Every distilled prior is \emph{critic-derived} rather than ground truth,
and retrieval quality, staleness, and the propagation of a wrong prior remain
open. The operator set $\mathcal{O}$ and the external channel $K$ are untouched: growing
the operator library through multimodal
distillation~\cite{plugmem2026,vilomem2026,m3agent2025} and admitting external
knowledge under a candidate-and-prune
discipline~\cite{contexttraining2026} are future work.

\appendix
\section{Appendix}\label{sec:appendix}

\subsection{Temporal predicates}\label{sec:app-temporal}
The order and containment predicates of \cref{sec:chaining} compare interval endpoints under a slack
of one quarter of the shorter window; for windows $W_X=[s_X,e_X]$,
\begin{equation}
\begin{gathered}
\delta(A,B)=\tfrac14
  \min(e_A-s_A,e_B-s_B),\\
\rho_{\mathrm{before}}(W_A,W_B)
=
\begin{cases}
\mathsf{S}, & e_A\leq s_B+\delta(A,B),\\
\mathsf{C}, & e_B\leq s_A+\delta(A,B),\\
\mathsf{U}, & \text{otherwise}.
\end{cases}
\end{gathered}
\label{eq:temporal-before}
\end{equation}
The quarter-window rule keeps the first two cases mutually exclusive for positive durations;
containment uses the same $\delta$.

\subsection{Withdrawal noise-level schedule}\label{sec:app-schedule}
\Cref{sec:gen-withdrawal} reports the withdrawal point by noise level $\sigma_{\mathrm{rel}}$, not
step index $i^{*}$. For $S$ steps at flow shift $\kappa$ over $N$ training timesteps, the UniPC
schedule assigns
\begin{equation}
\tilde{s}_i = \frac{(S-i)\,(1 - 1/N)}{S},
\qquad
\sigma_i = \frac{\kappa\,\tilde{s}_i}{1 + (\kappa-1)\,\tilde{s}_i},
\qquad
\sigma_{\mathrm{rel}} = \sigma_{i^{*}} ,
\label{eq:flow-sigma}
\end{equation}
with $\tilde{s}_i$ the normalized flow-matching timestep at step $i$ and $\sigma_i$ its noise level.

\subsection{Trajectory-agreement metrics}\label{sec:app-retention}
\Cref{sec:gen-release} scores how closely a generated clip's tracked path follows the simulation's.
With $p^{\mathrm{gen}}_t,p^{\mathrm{ctrl}}_t$ the per-frame tracked positions ($t\in\{0,\ldots,T-1\}$,
undefined where the tracker finds nothing) and $V$ the frames where both are defined,
\emph{retention} over a window $W$ is the per-axis Pearson correlation over $W\cap V$ ($R_x(W)$,
$R_y(W)$), reported only where $|W\cap V|$ meets a minimum count; $R$ near $1$ means the object moved
as simulated.

The windows split at the first ground-contact frame, off the control's vertical path: with
$y^{\mathrm{ctrl}}$ downward, $t_{\mathrm{apex}}=\arg\min_{t\in V} y^{\mathrm{ctrl}}_t$, and
$y_{\min},y_{\max}$ over the located frames,
$t_c=\min\{t\geq t_{\mathrm{apex}}: y^{\mathrm{ctrl}}_t \geq y_{\max}-\tfrac{1}{10}(y_{\max}-y_{\min})\}$,
giving \emph{airborne} $\{t<t_c\}$ and \emph{after-contact} $\{t\geq t_c\}$. Two prefixed failure
modes accompany retention: a \emph{whole-clip disagreement} (full-clip horizontal $<0.7$) and a
\emph{late-window disagreement} (full-clip horizontal $<0.7$ while airborne horizontal $>0.9$).

\subsection{Critic dataflow and procedure}\label{sec:app-critic-alg}
\begin{figure}[H]
  \centering
  \makeatletter
\@ifundefined{tikz@library@calc@loaded}{  \PackageError{fig_critic_pipeline}{    The preamble must contain\MessageBreak
    \string\usetikzlibrary{arrows.meta,positioning,calc}  }{Add that line after \string\usepackage{tikz}.}}{}
\makeatother

\begingroup
\providecommand{\cfsub}[1]{{\scriptsize\color{veriphygray}#1}}

\begin{tikzpicture}[
    x=1cm, y=1cm,
    cfbox/.style   = {rounded corners=2pt, line width=0.7pt, align=center,
                      inner sep=5pt, font=\sffamily\footnotesize,
                      text=veriphyink},
    cflearn/.style = {cfbox, draw=veriphyink, fill=veriphygray!12},
    cfdet/.style   = {cfbox, draw=veriphyink, fill=white},
    cfgate/.style  = {cfbox, draw=veriphyred, fill=white},
    cfout/.style   = {cfbox, draw=veriphyink, fill=white},
    cfdata/.style  = {cfbox, draw=veriphygray, fill=white, inner sep=4pt,
                      dash pattern=on 2pt off 1.6pt},
    cfline/.style  = {draw=veriphyink, line width=0.7pt},
    cfarr/.style   = {cfline, -{Straight Barb[length=3.6pt,width=3.8pt]}},
    cfnote/.style  = {font=\sffamily\scriptsize, text=veriphygray,
                      inner sep=0pt, align=left},
    cfsw/.style    = {rounded corners=1.2pt, line width=0.7pt, inner sep=0pt,
                      minimum width=11pt, minimum height=7.5pt},
  ]

  \node[cflearn, text width=9.4cm, anchor=north] (reasoner) at (0,0)
    {\textbf{Text-only planner}\\[1.5pt]
     \cfsub{extracts prompt-grounded claims and writes a surface plan\\
     specifying the checks that would settle them ---\\
     before any video frame is read}};

  \node[cfdata, text width=1.25cm, anchor=east]
    (prompt) at ($(reasoner.west)+(-0.62,0.20)$) {\textbf{Prompt}};

  \node[cfgate, text width=13.6cm, anchor=north]
    (gate) at ($(0,0)!1!(reasoner.south)+(0,-0.60)$)
    {{\color{veriphyred}\textbf{Every plan is checked before any tool runs}}\\[1.5pt]
     \cfsub{each line is parsed and type-checked; a line that cannot be used
     falls back to a general judgment,\\ and that fallback is counted in the
     plan's cost rather than treated as free}};

  \node[cfdata, text width=1.25cm, anchor=north east]
    (video) at ($(gate.south east)+(-0.15,-0.18)$) {\textbf{Video}};
  \coordinate (fanA) at ($(gate.south)+(0,-1.15)$);
  \node[cflearn, text width=5.6cm, anchor=north west]
    (verifier) at ($(gate.south west)+(0,-1.55)$)
    {\textbf{Semantic verifier}\\[1.5pt]
     \cfsub{reads the frames densely and decides\\ whether a thing is there and
     whether\\ an event happens}};
  \node[cflearn, text width=7.0cm, anchor=north east]
    (specialists) at ($(gate.south east)+(0,-1.55)$)
    {\textbf{Specialist instruments}\\[1.5pt]
     \cfsub{SAM 3 grounds the subject and tracks it;\\ eleven typed operations measure over those
     tracks;\\ counting reads its instance identities}};

  \coordinate (mergeA) at ($(verifier.south)+(0,-0.50)$);
  \node[cfdet, text width=13.6cm, anchor=north west]
    (evidence) at ($(verifier.north west |- mergeA)+(0,-0.40)$)
    {\textbf{Typed execution records}\\[1.5pt]
     \cfsub{specialists return measurements --- masks, tracks, counts, displacements,
     contacts, and windows; learned\\ base-predicate states are explicitly tagged; every
     record is tied to the prompt fragment that requested it}};

  \node[cfdet, text width=13.6cm, anchor=north west]
    (compose) at ($(evidence.south west)+(0,-0.40)$)
    {\textbf{Deterministic composition}\\[1.5pt]
     \cfsub{fixed rules combine the parts of a claim, then the claims into the
     clip: anything contradicted makes\\ its parent contradicted, everything
     supported makes it supported, and anything else is left unresolved}};

  \node[cflearn, text width=13.6cm, anchor=north west]
    (fallback) at ($(compose.south west)+(0,-0.40)$)
    {\textbf{Fallback verification}\\[1.5pt]
     \cfsub{a stronger model re-examines any claim the dense pass left supported; where it
     disagrees, the claim\\ resolves as contradicted rather than as absent evidence, and both
     verdicts are kept in the record}};

  \coordinate (fanB) at ($(fallback.south)+(0,-0.40)$);
  \node[cfout, text width=5.6cm, anchor=north west]
    (verdict) at ($(fallback.south west)+(0,-0.80)$)
    {{\textbf{Verdict}}\\[1.5pt]
     \cfsub{each claim supported, contradicted or\\ unknown; the whole clip
     plausible,\\ implausible, or abstained on}};
  \node[cfdet, text width=7.0cm, anchor=north east]
    (trace) at ($(fallback.south east)+(0,-0.80)$)
    {\textbf{Audit trace}\\[1.5pt]
     \cfsub{the claims checked, the evidence each call\\ returned, which claims
     fell back to a general\\ judgment, and the rule that combined them}};

  \draw[cfarr] (prompt.east) -- (reasoner.west);
  \draw[cfarr] (reasoner.south) -- (reasoner.south |- gate.north);

  \draw[cfline] (gate.south) -- (fanA);
  \draw[cfarr] (video.south) |- (fanA);
  \draw[cfline] (verifier.north |- fanA) -- (specialists.north |- fanA);
  \draw[cfarr]  (verifier.north |- fanA) -- (verifier.north);
  \draw[cfarr]  (specialists.north |- fanA) -- (specialists.north);

  \draw[cfline] (verifier.south)    -- (verifier.south    |- mergeA);
  \draw[cfline] (specialists.south) -- (specialists.south |- mergeA);
  \draw[cfline] (verifier.south |- mergeA) -- (specialists.south |- mergeA);
  \draw[cfarr]  (evidence.north |- mergeA) -- (evidence.north);

  \draw[cfarr] (evidence.south) -- (evidence.south |- compose.north);

  \draw[cfarr] (compose.south) -- (compose.south |- fallback.north);
  \draw[cfline] (fallback.south) -- (fanB);
  \draw[cfline] (verdict.north |- fanB) -- (trace.north |- fanB);
  \draw[cfarr]  (verdict.north |- fanB) -- (verdict.north);
  \draw[cfarr]  (trace.north   |- fanB) -- (trace.north);

  \node[cfsw, draw=veriphyink, fill=veriphygray!12, anchor=north west]
    (swA) at ($(verdict.south west)+(0,-0.50)$) {};
  \node[cfnote, right=3pt of swA] (lbA) {learned};

  \node[cfsw, draw=veriphyink, fill=white, right=9pt of lbA] (swB) {};
  \node[cfnote, right=3pt of swB] (lbB) {deterministic};

  \node[cfsw, draw=veriphyred, fill=white, right=9pt of lbB] (swC) {};
  \node[cfnote, right=3pt of swC] (lbC) {deterministic, before any tool};

  \node[cfsw, draw=veriphygray, fill=white, dash pattern=on 2pt off 1.6pt,
        right=9pt of lbC] (swE) {};
  \node[cfnote, right=3pt of swE] (lbE) {input};

\end{tikzpicture}
\endgroup
  \caption{\VeriPhy{} dataflow. The text-only planner writes claims and a surface
  plan; the compiler expands and validates it before any video is read. SAM~3
  grounds the subject and supplies the tracks the typed operations measure over, so
  specialists return numbers while the video-aware verifier returns tagged learned
  states. Composition is deterministic; the one learned step after it, the fallback
  verifier, may resolve a claim the dense pass left supported. Shaded boxes are
  learned, outlined boxes deterministic.}
  \label{fig:criticpipe}
\end{figure}
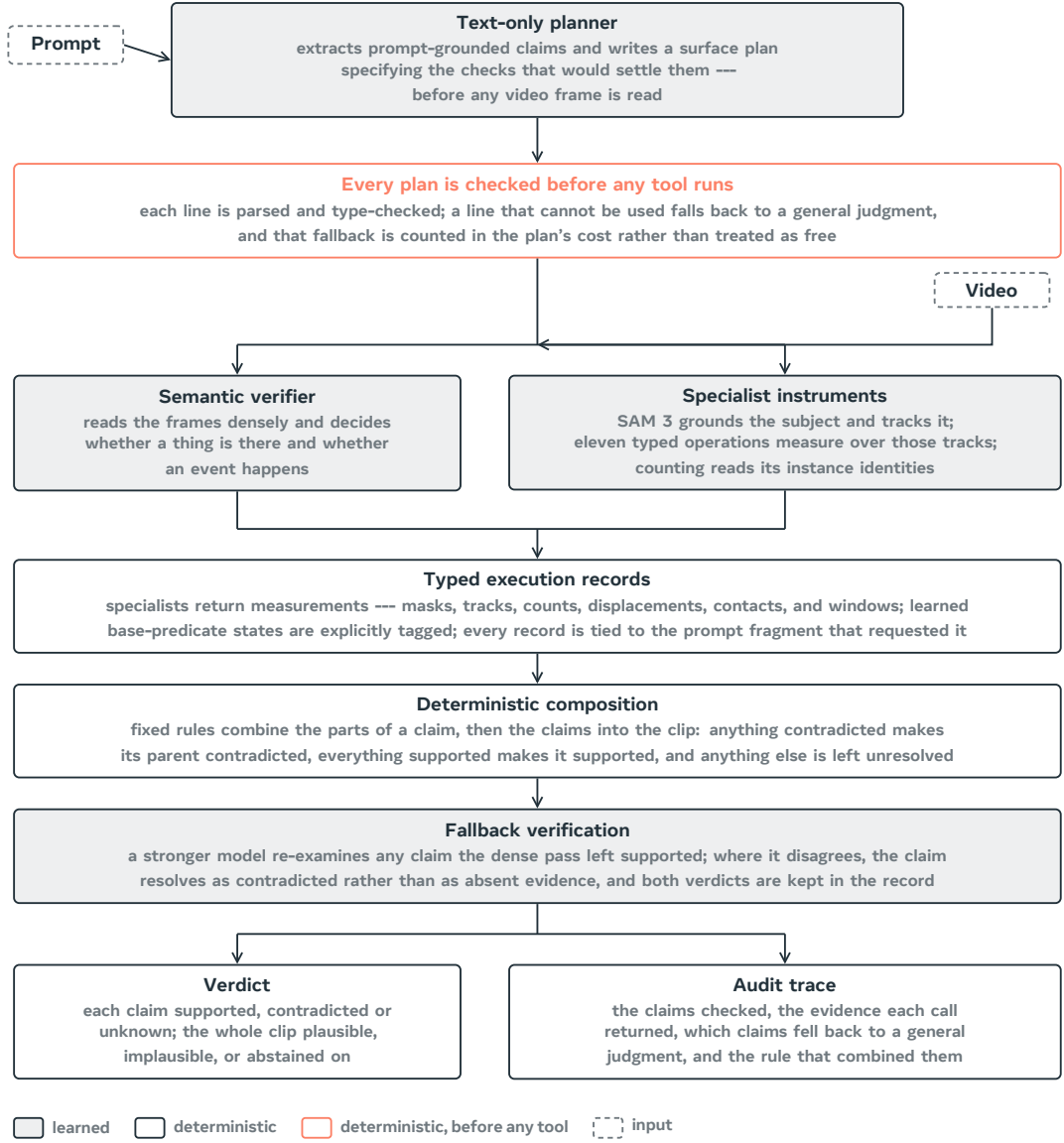

\Cref{fig:criticpipe} shows the dataflow of \cref{sec:critic} and \cref{alg:critic} the full one-clip
procedure, each step tagged learned or deterministic with its defining equation; both summarise
\cref{sec:critic} and add no new step.

\begin{algorithm}[H]
\caption{VeriPhy critic: bounded learned planning/verification embedded in deterministic compilation, scoping, and roll-up (executable summary of \cref{sec:critic}).}
\label{alg:critic}
\begin{algorithmic}[1]
\Require clip $\mathcal{V}=\{(I_t,\tau_t)\}_{t\in\mathcal{T}_{\mathcal V}}$, prompt $p$
\Ensure verdict $Y\in\{\textsc{Plausible},\textsc{Implausible},\textsc{Abstain}\}$ (or $\bot_{\mathrm{infra}}$)
\Procedure{Critic}{$\mathcal{V},p$}
  \State $(\mathcal{C},\widetilde{\Pi})\gets\Pi_\theta(p)$, \; $\mathcal{C}=\{c_i=(s_i,\phi_i)\}_{i=1}^{N_p}$, $s_i\sqsubseteq p$ \Comment{\eqref{eq:claims-and-plan}; text-only, video unseen \emph{(learned)}}
  \State $G_\Pi=(\mathcal{A},\mathcal{Q},\mathcal{E},b)\gets\textsc{Compile}(\widetilde{\Pi})$, \; $a_j=(o_j,\alpha_j,\omega_j,r_j)$ \Comment{typed DAG, before any frame is read; \eqref{eq:plan-dag} \emph{(det.)}}
  \If{$\neg\operatorname{Valid}(G_\Pi)$} \Comment{$\operatorname{Valid}=\operatorname{Parse}\land\operatorname{Refs}\land\operatorname{Typed}\land\operatorname{Acyclic}\land[\forall i\,\exists q\!:b(q){=}i]$; \eqref{eq:plan-validity} \emph{(det.)}}
     \State $G_\Pi\gets\textsc{Fallback}(G_\Pi)$; \; \textsc{Charge}\,\&\,\textsc{Report} \Comment{invalid line$\to$general verifier, invalid term$\to$recorded sub-check; never guessed}
  \EndIf
  \For{$a_j\in\mathcal{A}$ in topological order of $G_\Pi$} \Comment{video first touched here}
     \State $\Omega_j\gets g_j(\{R_k:(a_k,a_j)\in\mathcal{E}\},\mathcal{T}_{\mathcal V})$; \; $R_j\gets F_{o_j}(\mathcal{V}|_{\Omega_j},\alpha_j)$ \Comment{obs.\ gate scope \emph{(det.)}; frozen backend, SAM~3 grounds tracks; \eqref{eq:scoped-execution}}
  \EndFor
  \For{$q\in\mathcal{Q}$}
     \State $z_q\gets\psi_q(\{\mu_j:a_j\in\operatorname{Pa}(q)\})\in\mathbb{V}=\{\mathsf{S},\mathsf{C},\mathsf{U}\}$; \; negate via $\neg_3$ \Comment{$\psi_q$ det.\ for count/timing/track/transcription, learned asymmetric gate ($\mathsf{C}$ only on positive contradiction) for semantic; $\mathsf{U}$ if abstained, $\bot_{\mathrm{infra}}$ on infra error; \cref{sec:semantics}}
  \EndFor
  \For{$i\gets1$ to $N_p$}
     \State $K_i\gets\{q:b(q)=i\}$; \; $y_i\gets\mathsf{C}$ if $\exists q\!\in\!K_i\!:z_q{=}\mathsf{C}$, \; $\mathsf{S}$ if $\forall q\!\in\!K_i\!:z_q{=}\mathsf{S}$, \; else $\mathsf{U}$ \Comment{claim roll-up; \eqref{eq:claim-clip-rollup} \emph{(det.)}}
  \EndFor
  \ForAll{$i$ with $y_i=\mathsf{U}$} \Comment{single post-composition step \emph{(learned)}}
     \State $y_i\gets\textsc{FallbackVerify}(c_i,\{R_j\})$
  \EndFor
  \State $Y\gets\bot_{\mathrm{infra}}$ on infra error; \; \textsc{Implausible} if $\exists i\!:y_i{=}\mathsf{C}$; \; \textsc{Plausible} if $N_p{\ge}1\land\forall i\,y_i{=}\mathsf{S}$; \; else \textsc{Abstain} \Comment{clip roll-up; \eqref{eq:claim-clip-rollup} \emph{(det.)}}
  \State \Return $Y$
\EndProcedure
\end{algorithmic}
\end{algorithm}

\begin{figure}[tp]
  \centering
  {\ifdefined\tikzlibraryarrowsmetaloaded\else
  \expandafter\ifx\csname tikz@library@arrows.meta@loaded\endcsname\relax
    \PackageError{fig_plan_graph}{The preamble must contain
      \string\usetikzlibrary{arrows.meta,positioning,calc}}{}  \fi
\fi
\footnotesize
\begin{tikzpicture}[
  font=\footnotesize,
  >={Stealth[length=4pt,width=3pt]},
  node distance=0pt,
  pgclaim/.style   = {draw=veriphyink, fill=veriphygray!14, rounded corners=1.6pt,
                      line width=0.6pt, align=left, inner sep=4pt,
                      text width=3.32cm, minimum height=0.86cm},
  pgcheck/.style   = {draw=veriphyred, fill=white, rounded corners=1.6pt,
                      line width=0.6pt, align=center, inner sep=3.2pt,
                      text width=2.32cm},
  pgmeas/.style    = {draw=veriphyink, fill=white, rounded corners=1.6pt,
                      line width=0.6pt, align=left, inner sep=3.2pt,
                      text width=4.15cm},
  pggen/.style     = {pgmeas, fill=veriphygray!8, draw=veriphygray},
  pgedge/.style    = {draw=veriphygray, line width=0.55pt, ->},
  pglab/.style     = {font=\scriptsize, text=veriphygray, inner sep=1.2pt},
]

\node[draw=veriphygray, fill=veriphygray!6, line width=0.5pt, rounded corners=1.6pt,
      inner sep=5pt, text width=13.05cm, align=left, font=\ttfamily\scriptsize]
  (code) at (0,0)
  {check(c0, judge("a dolphin leaps out of the water"))\\
   check(c1, before(window("the dolphin leaps"), window("the dolphin dives back under")))\\
   check(c2, judge("two salmon emerge at the same moment") and eq(count("salmon"), 2))\\
   check(c3, judge("the salmon are behind the fishing boat") and rel("salmon||fishing boat", "behind"))};

\node[pgclaim, anchor=north west] (c0) at ($(code.south west)+(0.10,-0.62)$)
  {\textbf{c0}\\ the dolphin leaps};
\node[pgcheck, anchor=west] (p0) at ($(c0.east)+(1.05,0)$) {event occurs};
\node[pggen, anchor=west]  (m0) at ($(p0.east)+(1.05,0)$)
  {general verifier};
\draw[pgedge] (c0) -- (p0);
\draw[pgedge] (p0) -- (m0);

\node[pgclaim, anchor=north west] (c1) at ($(c0.south west)+(0,-0.55)$)
  {\textbf{c1}\\ it dives back \emph{after} the leap};
\node[pgcheck, anchor=west] (p1) at ($(c1.east)+(1.05,0)$) {order in time\\ (before, with slack)};
\node[pgmeas, anchor=south west] (m1a) at ($(p1.east)+(1.05,0.06)$)
  {event timing (leap)};
\node[pgmeas, anchor=north west] (m1b) at ($(p1.east)+(1.05,-0.06)$)
  {event timing (dive)};
\draw[pgedge] (c1) -- (p1);
\draw[pgedge] (p1.east) to[out=25,in=180] (m1a.west);
\draw[pgedge] (p1.east) to[out=-25,in=180] (m1b.west);

\node[pgclaim, anchor=north west] (c2) at ($(c1.south west)+(0,-1.15)$)
  {\textbf{c2}\\ two salmon emerge together};
\node[pgcheck, anchor=south west] (p2a) at ($(c2.east)+(1.05,0.42)$) {event occurs};
\node[pgcheck, anchor=north west] (p2b) at ($(c2.east)+(1.05,-0.42)$) {count is 2};
\node[pggen, anchor=west] (m2a) at ($(p2a.east)+(1.05,0)$)
  {general verifier};
\node[pgmeas, anchor=west] (m2b) at ($(p2b.east)+(1.05,0)$)
  {counter};
\draw[pgedge] (c2.east) to[out=25,in=180] node[pglab,pos=0.55,above]{and} (p2a.west);
\draw[pgedge] (c2.east) to[out=-25,in=180] node[pglab,pos=0.55,below]{and} (p2b.west);
\draw[pgedge] (p2a) -- (m2a);
\draw[pgedge] (p2b) -- (m2b);

\node[pgclaim, anchor=north west] (c3) at ($(c2.south west)+(0,-1.45)$)
  {\textbf{c3}\\ the salmon are behind the boat};
\node[pgcheck, anchor=south west] (p3a) at ($(c3.east)+(1.05,0.42)$) {event occurs};
\node[pgcheck, anchor=north west] (p3b) at ($(c3.east)+(1.05,-0.42)$) {behind\\ (depth order)};
\node[pggen, anchor=west] (m3a) at ($(p3a.east)+(1.05,0)$)
  {general verifier};
\node[pgmeas, anchor=west] (m3b) at ($(p3b.east)+(1.05,0)$)
  {outlines and depth};
\draw[pgedge] (c3.east) to[out=25,in=180] node[pglab,pos=0.55,above]{and} (p3a.west);
\draw[pgedge] (c3.east) to[out=-25,in=180] node[pglab,pos=0.55,below]{and} (p3b.west);
\draw[pgedge] (p3a) -- (m3a);
\draw[pgedge] (p3b) -- (m3b);

\node[pglab, anchor=south] at ($(c0.north)+(0,0.10)$) {claims from the prompt};
\node[pglab, anchor=south] at ($(p0.north)+(0,0.10)$) {typed checks};
\node[pglab, anchor=south] at ($(m0.north)+(0,0.10)$) {measurements};

\node[pgclaim, anchor=north west, text width=0.30cm, minimum height=0.30cm,
      inner sep=0pt] (kA) at ($(c3.south west)+(0,-1.05)$) {};
\node[pglab, anchor=west] (kAl) at ($(kA.east)+(0.10,0)$) {a claim};
\node[pgcheck, anchor=west, text width=0.30cm, minimum height=0.30cm,
      inner sep=0pt] (kB) at ($(kAl.east)+(0.55,0)$) {};
\node[pglab, anchor=west] (kBl) at ($(kB.east)+(0.10,0)$) {a typed check};
\node[pggen, anchor=west, text width=0.30cm, minimum height=0.30cm,
      inner sep=0pt] (kC) at ($(kBl.east)+(0.55,0)$) {};
\node[pglab, anchor=west] (kCl) at ($(kC.east)+(0.10,0)$) {answered by the general verifier};
\node[pgmeas, anchor=west, text width=0.30cm, minimum height=0.30cm,
      inner sep=0pt] (kD) at ($(kCl.east)+(0.55,0)$) {};
\node[pglab, anchor=west] at ($(kD.east)+(0.10,0)$) {measured by a specialist};

\end{tikzpicture}}
  \caption{A surface plan compiled into claims, typed checks, and evidence-producing
  actions (from \cref{sec:plan-lang}): a numeral adds a count check beside a semantic
  assertion; an ordering consumes two event windows after occurrence gating; a depth
  relation consumes localization, masks, and depth from the compiler's dependency
  closure. Illustrative of the planner's vocabulary, not an added learned verdict.}
  \label{fig:plangraph}
\end{figure}
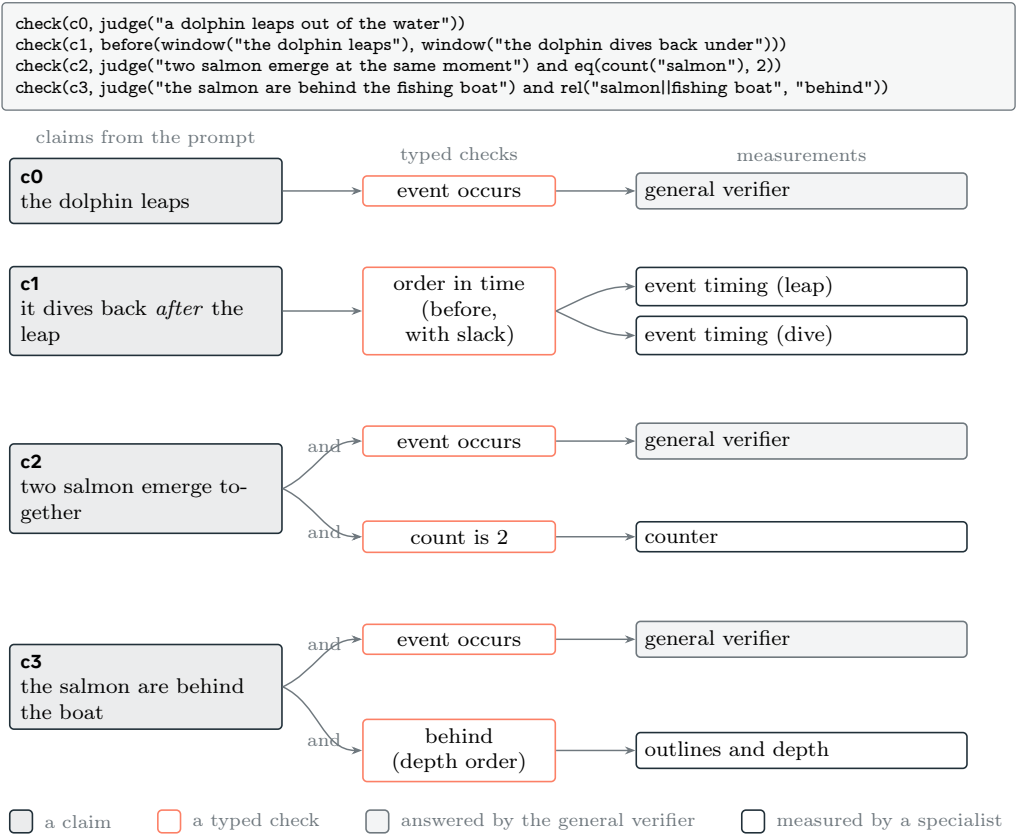

\subsection{Benchmark details}\label{sec:app-benchmark}
\begin{figure}[t]
  \centering
  \includegraphics[width=\linewidth]{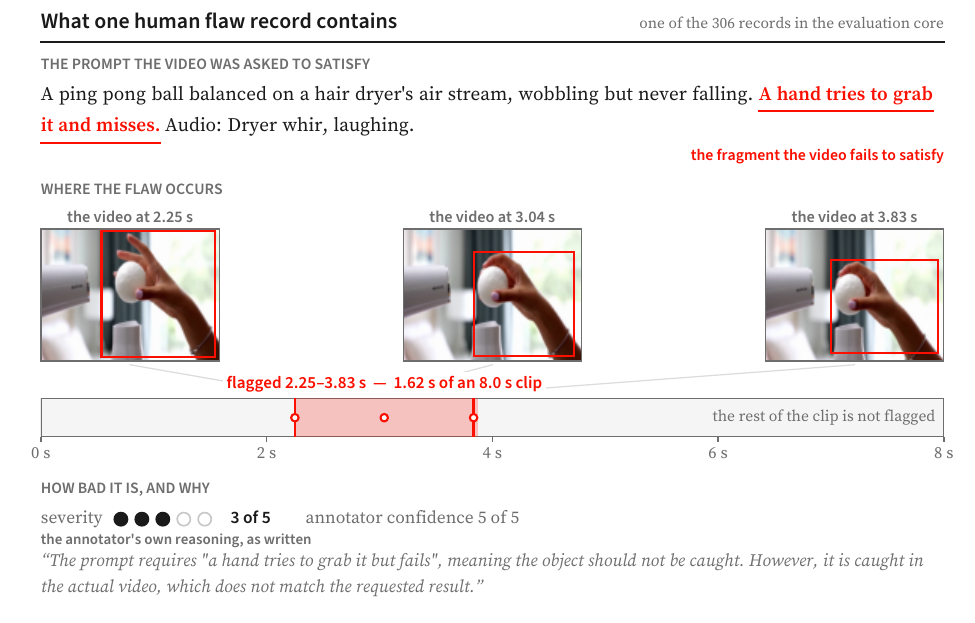}
  \caption{One human flaw record in full: the quoted prompt fragment, the interval
  of visible failure ($1.62$\,s of an $8$\,s clip), boxes on the objects at the
  interval's start/middle/end, and a severity, confidence, and rationale. No critic
  output appears.}
  \label{fig:benchanatomy}
\end{figure}

\begin{figure}[t]
  \centering
  \includegraphics[width=\linewidth]{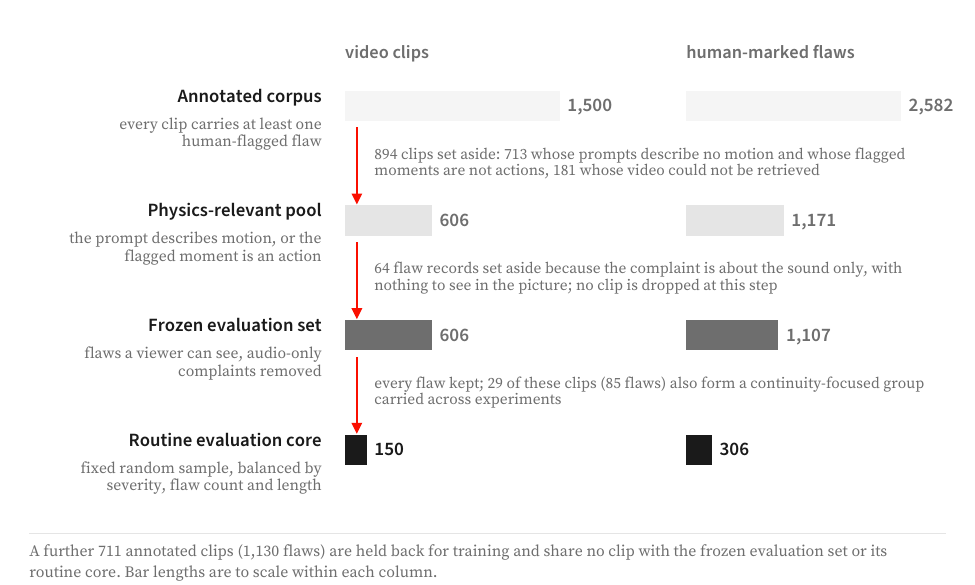}
  \caption{Deriving the evaluation set. Only the first step removes clips; the
  second removes audible-but-not-visible flaw \emph{records}, leaving the clip
  count unchanged. The core is a fixed random sample balanced by severity, flaw
  count, and length; a further $711$ annotated clips, disjoint from both frozen
  sets, are kept for training.}
  \label{fig:benchcomp}
\end{figure}

\begin{figure}[t]
  \centering
  \includegraphics[width=\linewidth]{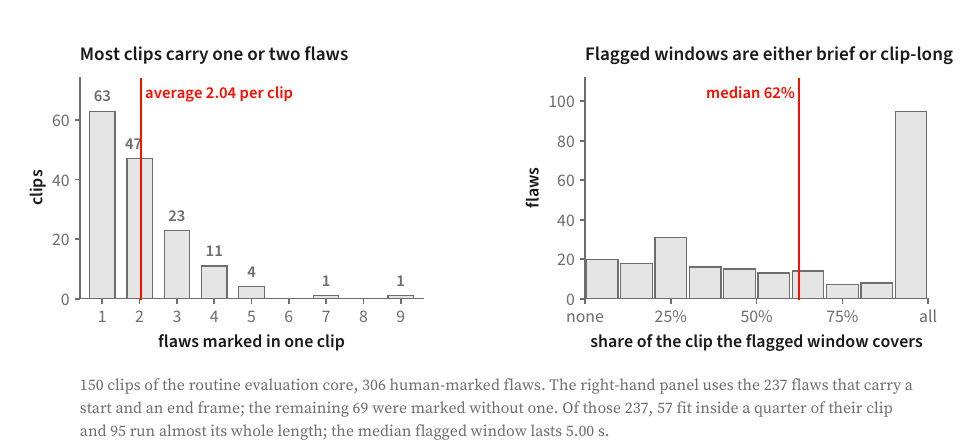}
  \caption{Two core properties that shape a detector. \textbf{Left:} most clips are
  flagged in only one or two places, so many complaints per clip are mostly ones no
  annotator marked. \textbf{Right:} flagged windows are bimodal---some failures
  confined to a small part of the clip, a comparable group spanning nearly all of
  it---so both a localized and a whole-clip check are needed.}
  \label{fig:benchdist}
\end{figure}

Field coverage is in \cref{tab:flawfields}, a full record in \cref{fig:benchanatomy}, the set's
derivation in \cref{fig:benchcomp}, and the core's distributions in \cref{fig:benchdist}.

By machine-assigned type, the core's $306$ flaws (median two per clip) split as actions absent or
wrong $139$; wrong counts $36$; wrong or missing objects $30$; garbled text $25$; wrong spatial
arrangement $19$; wrong ordering $17$; wrong attributes $16$; camera and style $10$; physical motion
$7$; residual $7$---counting, text, and arrangement together $80$, so this is not a physics benchmark
in the narrow sense. The $150$ clips run a median $8.0$\,s ($5.0$--$10.3$), mostly $24$\,fps
($118/150$), under prompts of median $42$ words (longest $188$), many naming a shot list, count,
ordering, on-screen string, and sound at once. Sixty-six carry more than one flaw kind. Of the $237$
time-spanned flaws the flagged window is bimodal---$57$ span a quarter of the clip or less, $95$ at
least nine-tenths (\cref{fig:benchdist})---so a detector needs both a localized and a whole-clip
check.

\begin{table}[t]
\centering\small
\setlength{\tabcolsep}{6pt}\renewcommand{\arraystretch}{1.18}
\begin{tabular}{@{}l l c@{}}
\toprule
\textbf{Field} & \textbf{Note} & \textbf{Coverage} \\
\midrule
violated fragment & median $8$ words & 306/306 \\
rationale & median $25$ words & 306/306 \\
severity & $1$--$5$, centred at $3$--$4$ & 306/306 \\
confidence & $1$--$5$, $239$ marked $4$--$5$ & 306/306 \\
time span & & 237/306 \\
box track & & 208/306 \\
\bottomrule
\end{tabular}
\caption{Coverage of a flaw record's six fields (defined in \cref{sec:bench-spec}) across the
$150$-clip core: the first four always present, timing and boxes only when the annotator
supplies them.}
\label{tab:flawfields}
\end{table}

\subsection{Published-evaluator comparison, in full}\label{sec:app-external}
\Cref{tab:external} is the full form of \cref{fig:external}, adding each evaluator's output and the
modular video-QA row (listed for context, not evaluated under this protocol, so no figure point).

\begin{table}[t]
\centering\small
\renewcommand{\arraystretch}{1.2}
\begin{tabularx}{\textwidth}{@{}L L c c c@{}}
\toprule
\textbf{Evaluator} & \textbf{Output it produces} & \textbf{Found} & \textbf{Recall}
 & \textbf{Mean model/tool calls per clip} \\
\midrule
\VeriPhy{} & localized flaw finding & \textbf{228} & \textbf{75.0\%} & 14.1 \\
\addlinespace
\multicolumn{5}{@{}l}{\textit{Other evaluators with structured localized outputs}} \\
\quad Question decomposition~\cite{tifa2023} & per-question answers & 164 & 53.9\% & 21.9 \\
\quad Davidsonian scene graph~\cite{dsg2024} & per-question answers, dependency-gated & 161 & 53.0\% & 52.1 \\
\addlinespace
\multicolumn{5}{@{}l}{\textit{Free-text localized output}} \\
\quad Single-pass VLM, same model~\cite{qwen3vl2025} & free-text flaw list & 222 & 73.0\% & 1.0 \\
\bottomrule
\end{tabularx}
\caption{Localized flaw recall over the same $149$ clips and the same $304$ flaws, judged by
the protocol of \cref{sec:matching}. The calls column reports the mean number of model and tool calls per clip and excludes
calls made by the matching protocol, which scores outputs rather than generating
evaluations. Question decomposition uses the same claim
list as \VeriPhy{} on $148$ of the $149$ clips. The free-text row uses the same served
model as \VeriPhy{}, prompted once with every frame; it is listed separately because it
produces an unstructured flaw list rather than structured per-question answers or
per-verdict evidence. A dash marks an evaluator not evaluated
under this protocol.}
\label{tab:external}
\end{table}

\subsection{Physion-Eval comparison, in full}\label{sec:app-leaderboard}
\Cref{tab:leaderboard} is the full form of \cref{fig:leaderboard}, adding this work's matched-report and
anomaly-phrased rates and the benchmark's $J$ statistics---defined against withheld real-world
reference videos, hence dashed for every evaluator on the release.

\begin{table}[t]
\centering\small
\renewcommand{\arraystretch}{1.2}
\begin{tabularx}{\textwidth}{@{}L c c c@{}}
\toprule
\textbf{Evaluator} & \textbf{$J$ exo} & \textbf{$J$ ego} & \textbf{Flag rate} \\
\midrule
Untrained humans (reference) & $30.7\%$ & $56.0\%$ & $52.7\%$ \\
\addlinespace
\multicolumn{4}{@{}l}{\textit{This work, judging from the critic's measured evidence}} \\
\quad Evidence-conditioned flag rate & --- & --- & $\mathbf{26.8\%}$ \\
\quad Matched-report clip rate & --- & --- & $18.4\%$ \\
\quad Anomaly-phrased finding rate & --- & --- & $15.5\%$ \\
\addlinespace
\multicolumn{4}{@{}l}{\textit{Critics evaluated by the benchmark, judging raw frames}} \\
\quad Gemini 3.0 Pro & $11.3\%$ & $5.5\%$ & $16.9\%$ \\
\quad Qwen3-VL-32B & $3.2\%$ & $1.0\%$ & $14.4\%$ \\
\quad Qwen3-VL-8B & $2.7\%$ & $0.8\%$ & $12.7\%$ \\
\quad Claude 4.5 Opus & $5.7\%$ & $2.3\%$ & $10.7\%$ \\
\quad GPT-5.2 & $4.0\%$ & $3.7\%$ & $10.5\%$ \\
\quad Gemini 2.5 Pro & $7.1\%$ & $1.1\%$ & $10.4\%$ \\
\quad Gemini 2.5 Flash & $4.4\%$ & $1.7\%$ & $8.2\%$ \\
\quad Gemini 2.5 Flash Lite & $2.0\%$ & $0.4\%$ & $4.5\%$ \\
\quad Cosmos Reason 1 & $0.1\%$ & $0.5\%$ & $2.8\%$ \\
\quad Cosmos Reason 2 & $0.9\%$ & $0.5\%$ & $1.2\%$ \\
\bottomrule
\end{tabularx}
\caption{Physion-Eval~\cite{physioneval2026} benchmark table. This work's rows are
measured on the first $601$ release clips of a fixed randomized stream ($543$ with a
reported glitch, $58$ without): the evidence-conditioned and anomaly-phrased rows report
the share of all $601$ clips flagged, and the matched-report row the share of the $543$
reported-glitch clips whose report a finding matches; the three measures are defined in
the text. The benchmark's rows are transcribed from its Figure~4 and are measured on its
full evaluation set: $J$ is its Youden statistic averaged over the five generators, and
the flag rate is the share of generated clips called physically unrealistic, pooled over
the two viewpoints. A dash in the $J$ columns marks a statistic defined against matched
real-world reference videos that the public release withholds, so it cannot be computed
for any evaluator run on the release.}
\label{tab:leaderboard}
\end{table}

\subsection{Recall by clip property}\label{sec:app-robust}
\Cref{tab:robust} groups the $304$ core flaws (run of \cref{tab:headline}) by severity, duration,
flaw count, and difficulty stratum. Subgroup recall stays within about ten points of the overall
rate everywhere, so neither severity nor stratum predicts whether a defect is found.

\begin{table}[t]
\centering\small
\renewcommand{\arraystretch}{1.2}
\begin{tabularx}{\textwidth}{@{}L c c c@{}}
\toprule
\textbf{Group} & \textbf{Annotated} & \textbf{Found} & \textbf{Recall} \\
\midrule
\multicolumn{4}{@{}l}{\textit{Annotator severity}} \\
\quad severity 1 & 3 & 3 & 100\% \\
\quad severity 2 & 28 & 22 & 79\% \\
\quad severity 3 & 146 & 107 & 73\% \\
\quad severity 4 & 99 & 78 & 79\% \\
\quad severity 5 & 28 & 18 & 64\% \\
\addlinespace
\multicolumn{4}{@{}l}{\textit{Clip duration}} \\
\quad 6 s or less & 76 & 60 & 79\% \\
\quad 6-8 s & 46 & 29 & 63\% \\
\quad over 8 s & 182 & 139 & 76\% \\
\addlinespace
\multicolumn{4}{@{}l}{\textit{Number of annotated flaws per clip}} \\
\quad 1 & 63 & 49 & 78\% \\
\quad 2 & 92 & 65 & 71\% \\
\quad 3-4 & 113 & 87 & 77\% \\
\quad 5 or more & 36 & 27 & 75\% \\
\addlinespace
\multicolumn{4}{@{}l}{\textit{Difficulty stratum}} \\
\quad hard & 83 & 60 & 72\% \\
\quad not hard & 221 & 168 & 76\% \\
\bottomrule
\end{tabularx}
\caption{Recall grouped by annotator severity, clip duration, number of annotated flaws in
the clip, and difficulty stratum, on the run of \cref{tab:headline}. Severity is the
annotator's own $1$--$5$ judgement of how badly the clip fails.}
\label{tab:robust}
\end{table}

\subsection{Refinement figures recomputed from the raw run}\label{sec:app-refine-verify}
\Cref{tab:refine-verify} recomputes every figure of \cref{sec:refine} from the per-clip records (all
reproduce exactly) and pins two definitions: the $8/502$ collateral figure and the differing baseline
and rewrite denominators.

\begin{table}[t]
\centering\small
\renewcommand{\arraystretch}{1.2}
\begin{tabularx}{\textwidth}{@{}L c c c@{}}
\toprule
\textbf{Quantity} & \textbf{Reported} & \textbf{Recomputed} & \\
\midrule
\multicolumn{4}{@{}l}{\textit{VeriPhy self-assessed metric, from the per-clip critic records ($589$ rows)}} \\
\quad acted (cited $\ge 1$ failure)              & $87/589$  & $87/589$  & \checkmark \\
\quad of acted, $p_{\text{plausible}}$ increased & $63/87$   & $63/87$   & \checkmark \\
\quad verdict implausible$\to$plausible          & $49/65$   & $49/65$   & \checkmark \\
\quad mean $\Delta p$ (acted)                    & $+0.524$  & $+0.524$  & \checkmark \\
\quad mean $p_{\text{plausible}}$ base$\to$rw    & $0.797\!\to\!0.862$ & $0.797\!\to\!0.862$ & \checkmark \\
\quad collateral (verdict flip plaus.$\to$implaus.) & $8/502$ & $8/502$   & \checkmark \\
\addlinespace
\multicolumn{4}{@{}l}{\textit{VideoPhy-2 AutoRater, rewrite side, per-clip scores ($589$ rows)}} \\
\quad joint \% (All / Hard)     & $23.1 / 3.4$  & $23.1 / 3.4$  & \checkmark \\
\quad mean SA (All / Hard)      & $3.13 / 2.76$ & $3.13 / 2.76$ & \checkmark \\
\quad mean PC (All / Hard)      & $3.58 / 3.32$ & $3.58 / 3.32$ & \checkmark \\
\quad SA\,$\ge 4$ \% (All / Hard) & $28.0 / 7.9$ & $28.0 / 7.9$ & \checkmark \\
\quad PC\,$\ge 4$ \% (All / Hard) & $54.5 / 36.2$ & $54.5 / 36.2$ & \checkmark \\
\bottomrule
\end{tabularx}
\caption{Every number in \cref{fig:refine-selfmetric,tab:refine-autorater} recomputed
independently from the run's per-clip raw files, against the reported value.
All reproduce exactly. Two definitions are pinned by the recompute: \emph{collateral}
is a verdict flip (plausible$\to$implausible) on a non-flagged clip---$8/502$; scored
instead as \emph{any} decrease in $p_{\text{plausible}}$ it would be $31/502$, so the
stricter reading is the one reported. The AutoRater \emph{baseline} column of
\cref{tab:refine-autorater} is not in this check: the released bundle synced only the
rewrite-side per-clip scores and a $12$-clip baseline subset, so the $600$-clip baseline
means are corroborated by the run log rather than recomputed here. Baseline $n=600$ and
rewrite $n=589$ (the $11$ absent clips did not complete regeneration or re-scoring), so
the two columns of \cref{tab:refine-autorater} rest on slightly different denominators.}
\label{tab:refine-verify}
\end{table}

\renewcommand{\bibfont}{\footnotesize}

\section{Framework: Learning as State Optimization}\label{sec:framework}

The preceding sections described three artifacts: a
physics-guided generator (\cref{sec:generation}), a $1{,}500$-clip human-authored
defect benchmark (\cref{sec:dataset}), and a physically grounded, tool-using
critic whose plan-based adjudication we measure in \cref{sec:eval}. This section states the
abstraction that ties them together and the form the framework is built toward. Our thesis is that a physical-reasoning
agent should keep everything it has learned in one
\emph{human-readable, typed, provenance-carrying state}, and should improve by
\emph{optimizing that state}. Every improvement
is then an edit to an object a person can read, version, and veto, and the arc of
the framework culminates in a system that evolves itself without ever touching a
weight.

\Cref{sec:experiments} reports experiments; the remainder of this section
sets out the framework and the roadmap. Because all learning is state optimization on a readable
$S$, that boundary falls inside the formalism itself, and \cref{tab:builtvsroadmap} marks it.

\subsection{From ReAct and agent memory to a physical agent with typed state}
\label{sec:framework:genealogy}

Agentic reasoning in the ReAct lineage interleaves \emph{reason}, \emph{act}, and
\emph{observe} steps, letting a model call external tools and condition on their
returns~\cite{react2023,reflexion2023,critic2024}. Program-of-thought variants
push this further: the chain of thought is emitted \emph{as a program} that
composes tool calls~\cite{rewoo2023,llmcompiler2024,vipergpt2023,visprog2023,toolformer2023,hugginggpt2023}. Orthogonally, a fast-growing agent-memory
literature asks where an agent's accumulated experience should live---episodic
transcripts, distilled semantic priors, or reusable
procedures~\cite{contexttraining2026,reasoningbank2026,explicitmem2026,plugmem2026,vilomem2026,amem2025,agentmemsurvey2024,m3agent2025,hindsight2026,aimeetsbrain2026}. \VeriPhy{} combines these two lines and changes the \emph{carrier} of memory: where these
agents store free-text notes or latent vectors, it stores structured physical-verification traces
with full provenance. Auditability follows from that choice, and it is the throughline of the
framework below.

\subsection{The unifying abstraction}
\label{sec:framework:abstraction}

\VeriPhy{} maintains a single, human-readable \emph{agent state}
\begin{equation}
  S \;=\; (\mathcal{O},\; M,\; K),
  \label{eq:state}
\end{equation}
and it improves by optimizing $S$. The reasoning model and the low-level physical
experts are frozen, so what the system accumulates is held in $S$, which is inspectable,
diffable, versionable, and revertible. ``Training'' and
``inference'' are one readable procedure applied in two directions: one
map reads $S$ to act, and channels write to $S$ (or to the generator) to improve.

\begin{description}
  \item[$\mathcal{O}$ --- operator library (procedural memory).] A set of reusable,
    already-compiled measurement operators. Each operator $o\in\mathcal{O}$ is a closure
    mapping a video and typed arguments to a physical measurement, and it carries
    an explicit \emph{cost node} $c(o)$ recording its spending (compute, expert
    calls, latency). $\mathcal{O}$ grows monotonically as recurring reasoning
    sub-procedures are packaged into named operators.
  \item[$M$ --- experience library (episodic + semantic memory).] Versioned,
    human-readable priors distilled from past traces: nearest-neighbor
    \emph{plans} for similar prompts and \emph{physics-failure priors} over
    common defect modes. $M$ keeps \emph{evidence} (what was measured) separate
    from \emph{belief} (what we now expect): evidence is retained as
    recorded, while belief is defeasible and is overridden by fresh measurement.
  \item[$K$ --- external knowledge channel.] Retrieved external or social
    knowledge (physical facts, domain conventions, human corrections) admitted
    only through a candidate-and-pruning discipline. $K$ is the deliberate
    \emph{open} port that keeps the internal improvement loop from closing on
    itself (\cref{sec:selfevolve}).
\end{description}

The rest of the framework consists of two kinds of maps over this state: a
single-episode \emph{read} operator that consumes $S$ and emits an auditable
trace (\cref{sec:framework:episode}), and \emph{write} channels that update the
generator and $S$ from that trace. The six design pillars of the paper are
recovered as special cases of read-vs-write over $S$: P1/P2 and the read side of
P4 are ``read $S$ to produce a trace''; P3 and P6 are ``write $S$''; P5 is
``write the generator''; and the supervisory side of P4 wraps the writes.

\subsection{The single-episode operator (read path): physicalized ReAct as a program over \texorpdfstring{$\mathcal{O}$}{O}}
\label{sec:framework:episode}

Given a prompt--video pair $(p, V)$, one episode is a deterministic pipeline that
only \emph{reads} $S$:
\begin{align}
  \text{retrieve:}    &\quad (\pi_p, m_p, k_p) = r(S, p), \label{eq:retrieve}\\
  \text{compile:}     &\quad (\mathcal{C}, G) = \operatorname{Reason}(p, V \mid \pi_p, m_p, k_p), \label{eq:compile}\\
  \text{act--observe:}&\quad R = \operatorname{Exec}_{\mathcal{O}}(G, V), \label{eq:exec}\\
  \text{adjudicate:}  &\quad (Y, \mathcal{F}) = \Phi(\mathcal{C}, R), \label{eq:adjudicate}\\
  \text{emit:}        &\quad T = (\mathcal{C}, G, R, Y, \mathcal{F}). \label{eq:trace}
\end{align}
where the retrieval $r$ reads from $S$ the nearest-neighbour plan $\pi_p$, the experience prior
$m_p$, and the external knowledge $k_p$ relevant to prompt $p$. These maps are the abstract form of
the critic of \cref{sec:critic}: \eqref{eq:compile}
is the claim-and-plan synthesis of \eqref{eq:claims-and-plan}, \eqref{eq:exec} the scoped
execution of \eqref{eq:scoped-execution}, and \eqref{eq:adjudicate} the roll-up and
packet of \eqref{eq:claim-clip-rollup} and \eqref{eq:refinement-packet}.
Here $\mathcal{C}$ is a set of \emph{typed physical obligations}---explicit,
machine-checkable statements of what the video must satisfy to honor the prompt
(object presence and count, contact and support, trajectory continuity, depth
ordering, on-screen text, audio events). $G$ is a \emph{program over $\mathcal{O}$}:
reasoning takes the form of a composition of operator calls
that the model writes and the runtime executes, in the tradition of
program-of-thought and modular visual
reasoning~\cite{visprog2023,vipergpt2023,rewoo2023,llmcompiler2024}. Executing $G$
drives the classic reason$\rightarrow$act$\rightarrow$observe loop of ReAct-style
agents~\cite{react2023}, but every \texttt{act} is a call into a \emph{frozen}
physical expert operator---promptable segmentation and tracking~\cite{sam3_2025},
counting over its instance identities, the typed physical measurements taken over those
tracks, monocular depth, OCR, and audio event
detection~\cite{flexsed2025}---registered in $\mathcal{O}$. This is the sense in which
\VeriPhy{} specializes ReAct to physical measurement: it inherits the control-flow harness from
mainstream tool-using agents~\cite{toolformer2023,hugginggpt2023} and specializes
it with explicit physical obligations and physical-measurement operators
\textbf{(P1)}. This general form is instantiated concretely by the critic of
\cref{sec:critic}, which produces the same three-valued verdict used there.

Each operator returns a measurement together with a \emph{gate} on its own scope:
when an operator's preconditions are not met (out-of-scope object, untrackable
occlusion, no audio track), it emits \abstain{}.
The adjudicator $\Phi$ in \eqref{eq:adjudicate} is a \emph{deterministic} roll-up
from evidence $R$ to a per-obligation verdict $Y\in\mathbb{V}^{|\mathcal{C}|}$ in the three-valued
codomain $\mathbb{V}=\{\mathsf{S},\mathsf{C},\mathsf{U}\}$ of \cref{sec:critic}---one entry per
obligation (\supported{}, \contradicted{}, or \unknownv{})---plus a \emph{contradiction packet} $\mathcal{F}$ that localizes, for each unmet obligation,
which measurement contradicts it and where. \unknownv{}, and \abstain{} for out-of-scope
obligations, are first-class outcomes of the codomain, so declining to certify is available to
the critic as a verdict in its own right. The
emitted trace $T=(\mathcal{C},G,R,Y,\mathcal{F})$ carries full provenance---every element of $Y$ is
back-traceable to the operator calls in $G$ and the measurements in $R$.

\paragraph{CoT-as-program and cost nodes \textnormal{(P2)}.} Because $G$ is a
program, recurring sub-procedures (e.g.\ ``verify a bounce $=$ track the object,
detect the contact frame, check the pre/post vertical-velocity sign flip'') can
be packaged, named, and re-registered as new operators in $\mathcal{O}$. Each operator
carries its cost node $c(o)$, so a program has a well-defined budget
$c(G)=\sum_{o\in G} c(o)$ and the reasoner can trade breadth of measurement
against spending. Over many episodes $\mathcal{O}$ scales up into a growing procedural
memory, and reasoning increasingly becomes \emph{composition of trusted
operators} rather than re-derivation from scratch. What is instantiated today is
a fixed operator set $\mathcal{O}_0$ and a fixed compiler; the \emph{scaling-up} of $\mathcal{O}$
via distillation is roadmap (\cref{sec:framework:builtvsroadmap}).

\subsection{The outward channel: verification-guided refinement \textnormal{(P5)}}
\label{sec:framework:control}

The read path produces, for a failed clip, a localized contradiction packet $\mathcal{F}$
(defined in \cref{sec:loop}). The outward
\emph{control channel} maps $\mathcal{F}$ into a \emph{local edit $\Delta_{\mathrm{ctrl}}$ of the
generator's control signal} and regenerates:
\begin{equation}
  V' \;=\; \operatorname{Gen}\!\big(p,\; \delta(\mathcal{F})\big),
  \qquad \delta:\ \mathcal{F} \;\longmapsto\; \Delta_{\mathrm{ctrl}},
  \label{eq:control}
\end{equation}
where $\operatorname{Gen}$ is the physics-guided pipeline of \cref{sec:generation}
(MuJoCo simulation~\cite{todorov2012mujoco} driving a Wan~2.2 video
model~\cite{wan2025} through VACE control conditioning~\cite{vace2025}), and
$\delta(\mathcal{F})$ is an edit to the simulation and control inputs addressing the violated obligation.
Since the control signal is a rendered field over space and time, a contradiction localized in
both maps onto a local edit of the conditioning stream. The critic here is independent and
frozen, so the grading signal does not originate in the generator being graded, unlike
self-evaluating refinement loops~\cite{mavis2025} and verification-guided
repair~\cite{neuse2025}. Both endpoints of \eqref{eq:control} are built
(\cref{sec:generation,sec:eval}); the map $\delta$ and the closed loop are roadmap.

\subsection{Auditability of the state and the trace}
\label{sec:framework:audit}

Auditability follows from the formalism. The learned object is the readable state $S$ and every
conclusion is a deterministic roll-up $\Phi$ over provenance-carrying measurements, so any verdict
$Y$ replays back to the operator calls in $G$ and the measurements in $R$, and any state change
$S\!\rightarrow\!S'$ is a human-readable diff of $\mathcal{O}$, $M$, or $K$. Weight-free self-evolution and
its human supervision both rest on that property: the reviewer inspects readable objects rather
than gradients.

Auditability determines what a verdict can be taken to mean. A \supported{} verdict is a critic judgment, where the
ground truth in this paper is human annotation. Determinism in the roll-up leaves evaluator drift
in the underlying operators and the compiler untouched. A prompt surviving the checks is
\emph{plausible} with respect to the obligations tested, and a static critic improving says
nothing about the generator.

\subsection{Weight-free context-training self-evolution}
\label{sec:selfevolve}

The framework's inward channel provides: a critic that improves itself without
updating a weight. Everything above---a readable state, a program over
operators, provenance traces, an independent verdict---exists so that the
system's own experience can be written into $S$ under a guard. This inward \emph{context-training} channel
\textbf{(P3)} is the last of the framework's components to be instantiated.

Given a trace $T$ and feedback $\phi$ (human labels, downstream outcomes, or a
later, more complete measurement), we distill a \emph{candidate} state edit
\begin{equation}
  (\Delta\mathcal{O}, \Delta M, \Delta K) = \operatorname{Distill}(T, \phi), \label{eq:distill}
\end{equation}
and accept it under a strict guard, the map $\operatorname{Accept}$ defined by \eqref{eq:elitism}
below. Both successful and failed episodes are distilled (hindsight over
traces~\cite{hindsight2026,reasoningbank2026}): a success may add a plan to $M$ or
an operator to $\mathcal{O}$; a failure may add a physics-failure prior or a scope restriction.
This form of self-evolution leaves weights unchanged, in the spirit of
context-distillation and experience-memory
agents~\cite{contexttraining2026,reasoningbank2026,reflexion2023,explicitmem2026,amem2025,agentmemsurvey2024,vilomem2026}, and it treats deployment-time adaptation
as \emph{learning as state optimization}. Acceptance is governed by
\begin{equation}
  S' =
  \begin{cases}
    S \oplus \Delta, & \text{if } U_{\mathrm{val}}(S\oplus\Delta) > U_{\mathrm{val}}(S),\\[2pt]
    S, & \text{otherwise (do-nothing elitism),}
  \end{cases}
  \label{eq:elitism}
\end{equation}
where $U_{\mathrm{val}}$ is a \emph{held-out proxy} score, $\Delta=(\Delta\mathcal{O},\Delta M,\Delta K)$
is a distilled candidate edit from \eqref{eq:distill}, and $S\oplus\Delta$ applies that edit
component-wise to the state, appending the operator, plan and knowledge deltas to
$\mathcal{O}$, $M$ and $K$ respectively. Candidates that do not strictly improve the proxy are
pruned, and when none improves it the state is left unchanged. That default makes the proxy
monotone across updates, bounding how far a distilled candidate can degrade $S$.

\paragraph{Admitting external knowledge \textnormal{(P6)}.}
A system whose only inputs are its own traces---CoT $\rightarrow$ feedback
$\rightarrow$ context $\rightarrow$ CoT---is a \emph{closed} loop that can amplify
its own biases: distilled priors condition future reasoning, whose traces distill
into more priors. The external knowledge channel $K$ is the opening provided for this.
\VeriPhy{} may issue retrieval or ask for external or social knowledge (physical
facts, human corrections, domain conventions) and admit it into $K$ under the same
candidate-and-pruning discipline as \eqref{eq:elitism}: a retrieved candidate is admitted only when it improves the held-out proxy, and $K$-sourced facts carry a provenance tag distinct from
distilled beliefs, so an admission can be reverted. External knowledge therefore enters under the
same elitism that governs $S$.

Every prior distilled into $M$ is \emph{critic-derived}: a summary of what \VeriPhy{}'s own
verification loop concluded rather than verified truth. The external channel $K$ is roadmap. The
inward channel is instantiated, and \cref{sec:icl-exp} measures it.

\subsection{Distilling lessons from recorded misses}
\label{sec:selfevolve:measured}
We instantiate \eqref{eq:distill} by distilling lessons from episodes in which a plan missed a
human-recorded flaw and appending them to the planner's rubric, leaving the operator library
$\mathcal{O}$ and external channel $K$ untouched so that $S'=(\mathcal{O},M',K)$. The measured
effect on flaw recall---lessons distilled from recorded misses raise held-out recall to
$375$ of $502$ against $340$ untaught, a gain already present from the smallest experience
pool---is reported with the other experiments in \cref{sec:icl-exp}.

\subsection{Human oversight}
\label{sec:framework:human}

A human is kept in the loop at a few specific gates: a
reviewer may audit a trace $T$, veto or edit a proposed state update in
\eqref{eq:elitism}, and approve whether a refined generation \eqref{eq:control} or
a new state $S'$ is deployed. Human feedback also enters $\operatorname{Distill}$ as a
first-class signal $\phi$. Because updates are readable diffs of $S$, oversight consists of
reading and approving them. The reviewing tooling and deployment protocol are roadmap; the
artifacts they act on --- traces, three-valued state, provenance --- exist today.

\subsection{Built and roadmap components}
\label{sec:framework:builtvsroadmap}

Because all learning is state optimization on a readable $S$, we can mark
built-vs-roadmap directly on the formalism, and each pillar is one component of the
same loop (\cref{tab:builtvsroadmap}). \textbf{Instantiated
today (Phase~0):} the frozen generator pipeline $\operatorname{Gen}$
(\cref{sec:generation}); the $1{,}500$-clip defect benchmark that exercises the
read path (\cref{sec:dataset}); the episode read operator
\eqref{eq:retrieve}--\eqref{eq:trace} with a fixed operator set $\mathcal{O}_0$, typed
obligations $\mathcal{C}$, three-valued deterministic adjudication $\Phi$, and provenance
traces $T$; the flaw-level measurement in \cref{sec:eval} against human labels
($228$ of $304$ for \VeriPhy{}, $164$ for question decomposition, recall-only, on the development core, which is
not held out); and the inward distillation of
$M$ via \eqref{eq:distill}, measured on a held-out split in
\cref{sec:icl-exp} ($375$ of $502$ against $340$ untaught).
\textbf{Roadmap (Phase~1+):} the
outward control map $\delta$ and the closed critic$\rightarrow$generator loop
\eqref{eq:control}; the scaling-up of $\mathcal{O}$; the acceptance gate \eqref{eq:elitism}
that admits distilled candidates on a measured proxy improvement; the external
knowledge channel $K$; and human-in-the-loop deployment at scale. Every roadmap item is a \emph{write}
path in the same formalism whose \emph{read} counterpart is already exercised. The executable
plan follows in \cref{sec:roadmap}.

\begin{table}[t]
\centering
\small
\begin{tabular}{@{}p{0.29\linewidth} p{0.38\linewidth} p{0.25\linewidth}@{}}
\toprule
\textbf{Pillar} & \textbf{Where it lives in the loop} & \textbf{Status} \\
\midrule
P1 Physical verification + operators & compile / act--observe over frozen experts~\cite{sam3_2025,flexsed2025} & Phase 0 (built) \\
P2 CoT-as-program + cost nodes & program $G$ over $\mathcal{O}$, budget $c(G)$ & Phase 0 partial \\
P4 Auditable trace + three-valued~$Y$ & trace $T=(C,G,R,Y,F)$, provenance & Phase 0 (artifacts) \\
P5 Verification-guided refinement & outward control channel $\delta,\,\operatorname{Gen}$ & roadmap \\
P2$^{+}$ Operator-library scale-up & procedural-memory growth in $\mathcal{O}$ & roadmap \\
P6 Breaking the loop & external channel $K$, candidate$+$prune & roadmap \\
P3 Context-training self-evolution & inward channel $\operatorname{Distill}$~\eqref{eq:distill} & Phase 0 (measured, gate not applied) \\
P3$^{+}$ Acceptance gate & elitism~\eqref{eq:elitism} over distilled candidates & roadmap \\
P4$^{+}$ Human oversight & supervision $+$ deployment gate & roadmap \\
\bottomrule
\end{tabular}
\caption{Each pillar is one component of the same loop. Only
Phase~0 rows denote implemented artifacts or measured analyses; solid arrows in \cref{fig:veriphy}
correspond to them, dashed arrows to the roadmap rows, which are unmeasured.}
\label{tab:builtvsroadmap}
\end{table}

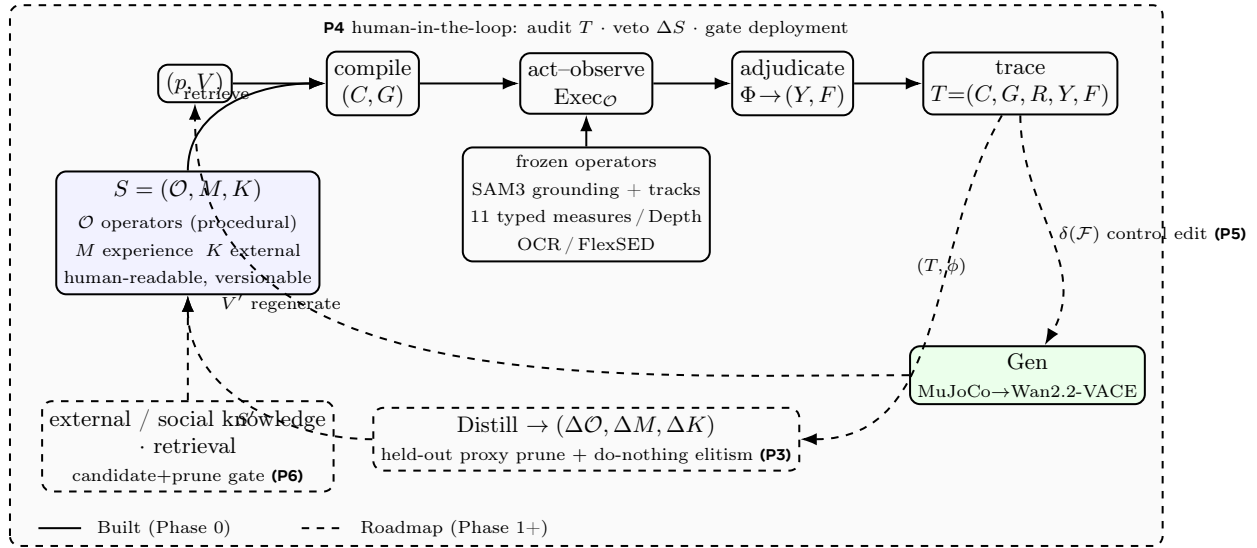
\begin{figure*}[t]
\centering
\resizebox{\linewidth}{!}{\begin{tikzpicture}[
  font=\small,
  built/.style={draw, thick, rounded corners, align=center, inner sep=3pt},
  road/.style ={draw, thick, dashed, rounded corners, align=center, inner sep=3pt},
  barrow/.style={-{Latex}, thick},
  rarrow/.style={-{Latex}, thick, dashed}]

  \draw[road, fill=black!2] (-7.5,-4.3) rectangle (8.7,3.3);
  \node[anchor=north] at (0.6,3.2)
    {\scriptsize \textbf{P4} human-in-the-loop: audit $T$ $\cdot$ veto $\Delta S$ $\cdot$ gate deployment};

  \node[built, fill=blue!5] (S) at (-5.0,0.1)
    {$S=(\mathcal{O},M,K)$\\[2pt]
     \scriptsize $\mathcal{O}$ operators (procedural)\\
     \scriptsize $M$ experience\; $K$ external\\
     \scriptsize human-readable, versionable};

  \node[built] (in)   at (-4.9,2.2) {$(p,V)$};
  \node[built] (comp) at (-2.4,2.2) {compile\\$(C,G)$};
  \node[built] (exec) at (0.6,2.2)  {act--observe\\$\operatorname{Exec}_{\mathcal{O}}$};
  \node[built] (adj)  at (3.5,2.2)  {adjudicate\\$\Phi\!\to\!(Y,F)$};
  \node[built] (T)    at (6.7,2.2)  {trace\\$T{=}(C,G,R,Y,F)$};
  \draw[barrow] (in)--(comp);
  \draw[barrow] (comp)--(exec);
  \draw[barrow] (exec)--(adj);
  \draw[barrow] (adj)--(T);
  \draw[barrow] (S) to[out=90,in=180] node[left,pos=0.65]{\scriptsize retrieve} (comp);

  \node[built] (ops) at (0.6,0.5)
    {\scriptsize frozen operators\\
     \scriptsize SAM3 grounding + tracks\\
     \scriptsize 11 typed measures\,/\,Depth\\
     \scriptsize OCR\,/\,FlexSED};
  \draw[barrow] (ops)--(exec);

  \node[built, fill=green!8] (gen) at (6.8,-1.9)
    {$\operatorname{Gen}$\\ \scriptsize MuJoCo$\to$Wan2.2-VACE};

  \draw[rarrow] (T) to[out=-90,in=60]
    node[right,pos=0.5]{\scriptsize $\delta(\mathcal{F})$ control edit \textbf{(P5)}} (gen);
  \draw[rarrow] (gen) to[out=180,in=-90]
    node[below,pos=0.72]{\scriptsize $V'$ regenerate} (in);

  \node[road] (dist) at (0.6,-2.8)
    {Distill $\to (\Delta\mathcal{O},\Delta M,\Delta K)$\\
     \scriptsize held-out proxy prune $+$ do-nothing elitism \textbf{(P3)}};
  \draw[rarrow] (T) to[out=-120,in=0] node[above,pos=0.4]{\scriptsize $(T,\phi)$} (dist);
  \draw[rarrow] (dist) to[out=180,in=-90] node[below,pos=0.5]{\scriptsize $S'$} (S);

  \node[road] (ext) at (-5.0,-2.9)
    {external / social knowledge\\ $\cdot$ retrieval\\
     \scriptsize candidate$+$prune gate \textbf{(P6)}};
  \draw[rarrow] (ext)--(S);

  \draw[thick] (-7.1,-4.05)--(-6.5,-4.05);
  \node[anchor=west] at (-6.4,-4.05) {\scriptsize Built (Phase 0)};
  \draw[thick,dashed] (-3.4,-4.05)--(-2.8,-4.05);
  \node[anchor=west] at (-2.7,-4.05) {\scriptsize Roadmap (Phase 1+)};
\end{tikzpicture}
}
\caption{State-optimization view of \VeriPhy{}. Everything the system learns
lives in one human-readable state $S=(\mathcal{O},M,K)$. A single \emph{read} operator
(top, solid = built) conditions on $S$, compiles $(p,V)$ into typed obligations
$\mathcal{C}$ and a program $G$ over $\mathcal{O}$, runs frozen physical operators, and emits a
provenance-carrying three-valued trace $T$. Two structurally dual \emph{write}
channels (dashed = roadmap), both driven by the same independent frozen critic,
close the loop: the outward control channel edits the generator's control signal
via $\delta(\mathcal{F})$ and regenerates (P5); the inward context-training channel
distills $T$ into candidate state edits accepted only under held-out proxy
pruning with do-nothing elitism (P3). The external channel $K$ breaks the
internal loop (P6). A human-in-the-loop layer wraps both channels (P4). Solid
arrows are instantiated in \cref{sec:generation,sec:dataset,sec:eval};
dashed arrows are the roadmap of \cref{sec:roadmap}.}
\label{fig:veriphy}
\end{figure*}

\section{An Executable Roadmap}
\label{sec:roadmap}

The framework of \cref{sec:framework} defines one learned object, the state
$S=(\mathcal{O},M,K)$, and three maps over it: a read operator (built) and two write
channels (roadmap). We phase the work accordingly. Phase~0 is the read path and
the two frozen endpoints already instantiated in this paper; Phase~1 moves the
evaluation onto a held-out split scored for precision as well as recall;
Phases~2--6 exercise
the write paths, each with a measurable success metric and each a strict
superset of what precedes it. No number below the Phase~0 line is a result;
Phases~1+ state \emph{targets and protocols}, not outcomes.

\paragraph{Phase~0 --- Built: the read path and frozen endpoints (\emph{done}).}
\begin{itemize}
  \item \textbf{Delivered.} The physics-guided generator $\operatorname{Gen}$
    (MuJoCo\,$\rightarrow$\,Wan2.2-VACE, \cref{sec:generation}); the $1500$-clip
    human-authored defect benchmark (\cref{sec:dataset}); the episode read
    operator with fixed $\mathcal{O}_0$, typed obligations $\mathcal{C}$, three-valued
    deterministic adjudication $\Phi$, and provenance traces $T$
    (\cref{sec:critic}); and the flaw-level measurement against human labels
    ($228$ of $304$ for \VeriPhy{} against $164$ for question decomposition, on
    the development core, non-held-out, \cref{sec:eval}).
  \item \textbf{Covers.} P1 (physical-verification read path), and the read side of
    P2 (programs over $\mathcal{O}_0$) and P4 (provenance traces).
  \item \textbf{Honest limits.} Recall-only, single non-held-out split; critic
    verdicts are judgments scored against human labels, not ground truth; a
    static evaluator result is not evidence any generator improved; the two
    write channels are not yet exercised.
\end{itemize}

\paragraph{Phase~1 --- Held-out, precision+recall evaluation with provenance completeness.}
\begin{itemize}
  \item \textbf{Goal.} Promote the critic result from dev-core / recall-only to a
    held-out, precision+recall protocol, and guarantee every verdict carries a
    complete provenance trace $T$.
  \item \textbf{Needs.} A frozen held-out split of the benchmark disjoint from
    dev-core; a provenance schema for $(C,G,R,Y,F)$.
  \item \textbf{Experiments.} Plan-based vs.\ per-claim adjudication on held-out
    clips with full confusion matrices; audit that every \contradicted{} verdict
    localizes to frame-level evidence via $F$.
  \item \textbf{Success metric.} Held-out plan-based accuracy $\geq$ per-claim
    with non-degrading precision; complete provenance coverage.
  \item \textbf{Risk.} Held-out gap reveals dev-core overfitting; the
    three-valued codomain inflates \unknownv{} to dodge hard cases --- mitigate
    by reporting the \unknownv{} rate alongside accuracy.
\end{itemize}

\paragraph{Phase~2 --- Operator library with cost accounting (P2).}
\begin{itemize}
  \item \textbf{Goal.} Turn fixed $\mathcal{O}_0$ into a growing, versioned $\mathcal{O}$ whose
    operators carry cost nodes $c(o)$ and whose programs have measurable budgets
    $c(G)$.
  \item \textbf{Needs.} An operator registry with provenance and versioning; a
    compiler that detects recurring $G$-sub-programs and promotes them into named
    operators; per-operator cost instrumentation wired into $\Phi$.
  \item \textbf{Experiments.} Track the coverage/cost frontier as $|\mathcal{O}|$ grows;
    ablate frozen $\mathcal{O}_0$ vs.\ promoted $\mathcal{O}$ at matched budget.
  \item \textbf{Success metric.} A Pareto improvement in obligation coverage
    per unit cost over the Phase~0 fixed pipeline at equal or lower $c(G)$.
  \item \textbf{Risk.} Library bloat and stale operators; promoted operators
    encode clip-specific shortcuts --- mitigate with usage-frequency thresholds
    and held-out re-validation before promotion.
\end{itemize}

\paragraph{Phase~3 --- Context-training self-evolution of $M$ (P3).}
\begin{itemize}
  \item \textbf{Goal.} Distill traces (success and failure) into human-readable,
    versionable priors in $M$ --- nearest-neighbor plans and physics-failure
    priors --- accepted under held-out proxy pruning with do-nothing elitism
    \eqref{eq:elitism}, keeping evidence separate from belief.
  \item \textbf{Needs.} A distillation function $\operatorname{Distill}$; a held-out
    validation split $U_{\mathrm{val}}$ (new data, disjoint from the Phase~0/1
    core); a diffable memory store.
  \item \textbf{Experiments.} Compare $\{$no memory, retrieval-only, distilled
    $M$ with elitism, distilled $M$ without elitism$\}$ on the held-out split; a
    pollution stress test injecting wrong priors and checking elitism rejects
    them.
  \item \textbf{Success metric.} Monotone non-decreasing held-out proxy across
    accepted updates, a positive gap of ``$M$ with elitism'' over both baselines,
    and zero acceptance of injected harmful priors.
  \item \textbf{Risk.} Belief leaking into evidence, or proxy overfitting;
    mitigate with the evidence/belief split and periodic held-out refresh.
\end{itemize}

\paragraph{Phase~4 --- Closing the control channel critic\,$\rightarrow$\,generator (P5).}
\begin{itemize}
  \item \textbf{Goal.} Implement $\delta:\mathcal{F}\mapsto\Delta_{\mathrm{ctrl}}$ of
    \eqref{eq:control}, converting a contradiction packet into a \emph{targeted}
    local edit of the MuJoCo/VACE control signal, and regenerate.
  \item \textbf{Needs.} A parameterization of Wan2.2-VACE control inputs by
    obligation type; a differentiable-or-heuristic edit policy $\delta$; a
    re-adjudication harness.
  \item \textbf{Experiments.} On clips with \contradicted{} verdicts, compare
    $\{$blind resampling, prompt-only edit, targeted $\delta(\mathcal{F})$ edit$\}$; report
    the fraction of previously violated obligations that flip to \supported{}
    after regeneration, \emph{human-verified}, and re-run the full obligation set
    $\mathcal{C}$ post-edit to report net defect change.
  \item \textbf{Success metric.} Targeted $\delta(\mathcal{F})$ regeneration repairs a
    significantly higher fraction of localized violations than blind resampling
    at equal generation budget, confirmed by human annotation (not by the critic
    alone), with no rise in unrelated-obligation contradictions.
  \item \textbf{Risk.} Goodharting the critic or whack-a-mole edits that fix one
    obligation while breaking another; mitigated by human verification and the
    net-defect audit.
\end{itemize}

\paragraph{Phase~5 --- Breaking the loop with external knowledge $K$ (P6).}
\begin{itemize}
  \item \textbf{Goal.} Open the internal CoT$\rightarrow$feedback$\rightarrow$context
    loop by admitting external/social knowledge and retrieval into $K$ under the
    same candidate-and-pruning discipline, with a do-nothing fallback.
  \item \textbf{Needs.} A retrieval/ingestion interface; a candidate gate scored
    by $U_{\mathrm{val}}$; provenance tagging that marks $K$-sourced facts
    distinctly from distilled beliefs.
  \item \textbf{Experiments.} On a bias-probe subset (defect modes
    under-represented in the internal traces), compare closed-loop
    ($K=\emptyset$) vs.\ open-loop ($K$ admitted with gating); a noise-injection
    test on retrieved sources.
  \item \textbf{Success metric.} Reduced error on the bias-probe subset with no
    regression on the general held-out split; zero admissions when no candidate
    improves the proxy (fallback correctness).
  \item \textbf{Risk.} Context pollution from noisy or off-distribution sources;
    mitigated by the same elitism gate as $S$, plus source provenance so bad
    admissions are revertible.
\end{itemize}

\paragraph{Phase~6 --- Human-in-the-loop deployment at scale (P4).}
\begin{itemize}
  \item \textbf{Goal.} Operate both channels under human supervision: reviewers
    audit traces $T$, veto/edit candidate state updates, and gate deployment of
    refined generations and new states $S'$.
  \item \textbf{Needs.} Trace-replay and state-diff review tooling; an approval
    workflow feeding $\operatorname{Distill}$ and $K$; audit logs over
    $S\!\rightarrow\!S'$ with rollback.
  \item \textbf{Experiments.} Held-out study measuring reviewer time per audited
    trace, inter-annotator agreement on state-update acceptance and on
    three-valued verdicts, and the catch-rate of injected bad updates that
    elitism alone would have missed.
  \item \textbf{Success metric.} Auditing cost per trace low enough to be
    practical, with human review reliably catching a measurable residual of
    harmful updates beyond the automatic gate.
  \item \textbf{Risk.} Reviewer fatigue and rubber-stamping (which would
    reintroduce an unaudited closed loop); mitigate by surfacing only
    elitism-borderline updates for human attention.
\end{itemize}

\noindent\emph{Ordering rationale.} Phase~1 secures the evaluation;
Phases~2--3 grow the two components of $S$ that the inward channel writes ($\mathcal{O}$,
$M$); Phase~4 exercises the outward channel; Phase~5 opens $K$; Phase~6 wraps
everything in human oversight. Each phase reuses the Phase~0 read path unchanged,
so at every step the system remains weight-free and auditable by construction.

\clearpage
\bibliographystyle{assets/plainnat}
\bibliography{references}

\end{document}